\documentclass{article}

\usepackage{microtype}
\usepackage{graphicx}
\usepackage{subfigure}
\usepackage{booktabs} 

\usepackage{hyperref}

\usepackage[accepted]{icml2023}

\usepackage{amsmath}
\usepackage{amssymb}
\usepackage{mathtools}
\usepackage{amsthm}
\usepackage{booktabs}
\usepackage{xspace}
\usepackage{tabularx}

\usepackage[utf8]{inputenc} 
\usepackage[T1]{fontenc}    

\usepackage{url}            
\usepackage{booktabs}       
\usepackage{amsfonts}       
\usepackage[subfigure]{tocloft}
\usepackage{nicefrac}       
\usepackage{microtype}      
\usepackage{xcolor}         

\usepackage{graphicx}      
\usepackage[toc,page,header]{appendix}
\usepackage{minitoc}
\usepackage{makecell}
\usepackage{adjustbox}
\usepackage{colortbl}
\definecolor{Gray}{gray}{0.92}
\usepackage{amsmath,bm,bbm}
\usepackage{amssymb}
\usepackage{mathtools}
\usepackage{amsthm}
\usepackage{multirow}
\usepackage{wrapfig}
\usepackage{enumitem}
\usepackage[capitalize,noabbrev]{cleveref}

\newcommand{\myref}[1]{Eq.\eqref{#1}}

\usepackage{mdframed}
\theoremstyle{plain}
\definecolor{theoremcolor}{rgb}{0.94, 0.94, 0.94}
\definecolor{examplecolor}{rgb}{1, 1, 1.0}
\mdfsetup{
    backgroundcolor=theoremcolor,
    linewidth=0pt,
}

\newcommand*{\circled}[1]{\lower.7ex\hbox{\tikz\draw (0pt, 0pt)%
    circle (.5em) node {\makebox[1em][c]{\small #1}};}}

\newcommand{\knn}[0]{AdaNPC\xspace}

\theoremstyle{plain}
\newmdtheoremenv[linewidth=0pt,innerleftmargin=0pt,innerrightmargin=0pt]{definition}{Definition}
\newmdtheoremenv[linewidth=0pt,innerleftmargin=4pt,innerrightmargin=4pt]{prop}{Proposition}
\newmdtheoremenv[linewidth=0pt,innerleftmargin=4pt,innerrightmargin=4pt]{assump}{Assumption}
\newmdtheoremenv[linewidth=0pt,innerleftmargin=0pt,innerrightmargin=0pt,backgroundcolor=examplecolor]{example}{Example}
\newmdtheoremenv{corollary}{Corollary}
\newmdtheoremenv[linewidth=0pt,innerleftmargin=4pt,innerrightmargin=4pt]{theorem}{Theorem}
\newmdtheoremenv[linewidth=0pt,innerleftmargin=4pt,innerrightmargin=4pt]{lemma}{Lemma}
\usepackage[textsize=tiny]{todonotes}
\usepackage[toc,page,header]{appendix}
\usepackage{minitoc}
\usepackage[capitalize]{cleveref}
\crefname{section}{Sec.}{Secs.}
\Crefname{section}{Section}{Sections}
\Crefname{table}{Table}{Tables}
\crefname{table}{Tab.}{Tabs.}

\icmltitlerunning{AdaNPC: Exploring Non-Parametric Classifier for Test-Time Adaptation}

\begin{document}

\twocolumn[
\icmltitle{AdaNPC: Exploring Non-Parametric Classifier for Test-Time Adaptation}

\icmlsetsymbol{equal}{*}

\begin{icmlauthorlist}
\icmlauthor{Yi-Fan Zhang}{intern,casia,mais}
\icmlauthor{Xue Wang}{ali}
\icmlauthor{Kexin Jin}{prin}
\icmlauthor{Kun Yuan}{pku}
\icmlauthor{Zhang Zhang }{casia,mais}
\icmlauthor{Liang Wang }{casia,mais}
\icmlauthor{Rong Jin }{twitter}
\icmlauthor{Tieniu Tan }{casia,mais,nanjing}
\end{icmlauthorlist}

\icmlaffiliation{intern}{Work done during an internship at Alibaba Group.}
\icmlaffiliation{mais}{MAIS, CRIPAC, CASIA.}
\icmlaffiliation{casia}{School of Artificial Intelligence, University of Chinese Academy of Sciences (UCAS), CRIPAC & NLPR.}
\icmlaffiliation{casia}{School of Artificial Intelligence, University of Chinese Academy of Sciences (UCAS)}
\icmlaffiliation{ali}{Machine Intelligence Technology, Alibaba Group. }
\icmlaffiliation{prin}{ Department of Mathematics at Princeton University.}
\icmlaffiliation{pku}{Center for Machine Learning Research, Peking University.}
\icmlaffiliation{twitter}{Work done at Alibaba Group, and now affiliated with Meta.}
\icmlaffiliation{nanjing}{Nanjing University.}

\icmlcorrespondingauthor{Yi-Fan Zhang}{yifanzhang.cs@gmail.com}

\icmlkeywords{Machine Learning, ICML}

\vskip 0.3in
]
\printAffiliationsAndNotice{} 

\vspace{-0.1cm}
\begin{abstract}
\vspace{-0.1cm}

Many recent machine learning tasks focus to develop models that can generalize to unseen distributions. Domain generalization (DG) has become one of the key topics in various fields. Several literatures show that DG can be arbitrarily hard without exploiting target domain information. To address this issue, test-time adaptive (TTA) methods are proposed. Existing TTA methods require offline target data or extra sophisticated optimization procedures during the inference stage. In this work, we adopt \textbf{N}on-\textbf{P}arametric \textbf{C}lassifier to perform the test-time \textbf{Ada}ptation (\knn). In particular, we construct a memory that contains the feature and label pairs from training domains. During inference, given a test instance, \knn first recalls $k$ closed samples from the memory to vote for the prediction, and then the test feature and predicted label are added to the memory. In this way, the sample distribution in the memory can be gradually changed from the training distribution towards the test distribution with very little extra computation cost. We theoretically justify the rationality behind the proposed method. Besides, we test our model on extensive numerical experiments. \knn significantly outperforms competitive baselines on various DG benchmarks. In particular, when the adaptation target is a series of domains, the adaptation accuracy of \knn is $50\%$ higher than advanced TTA methods. Code is available at \href{https://github.com/yfzhang114/AdaNPC}{https://github.com/yfzhang114/AdaNPC}. 

 \vspace{-0.2cm}
\end{abstract}
\vspace{-0.1cm}
\vspace{-0.3cm}
\section{Introduction}\label{sec:intro}
The classic machine learning models generally suffer from degraded performance when the training and test data are not from the same distribution. Many researchers consider developing out-of-distribution (OOD) generalization approaches (e.g., disentanglement~\cite{zhang2022towards}, causal invariance~\cite{arjovsky2020invariant,zhang2022exploring}, and adversarial training~\cite{ganin2016domain,li2018deep}.), in which models are trained on multiple source domains/datasets and can be directly deployed on \emph{unseen} target domains.

\begin{figure}[t]
   \subfigure[\knn memorizes features and labels of source domain instances. \textit{During inference}, each arrival target sample will be classified by a KNN classifier, where the nearest neighbors are searched in the memory. \textit{For test-time adaptation}, the target feature and prediction will be further stored in the memory bank.]{
    \centering
    \includegraphics[width=0.95\linewidth]{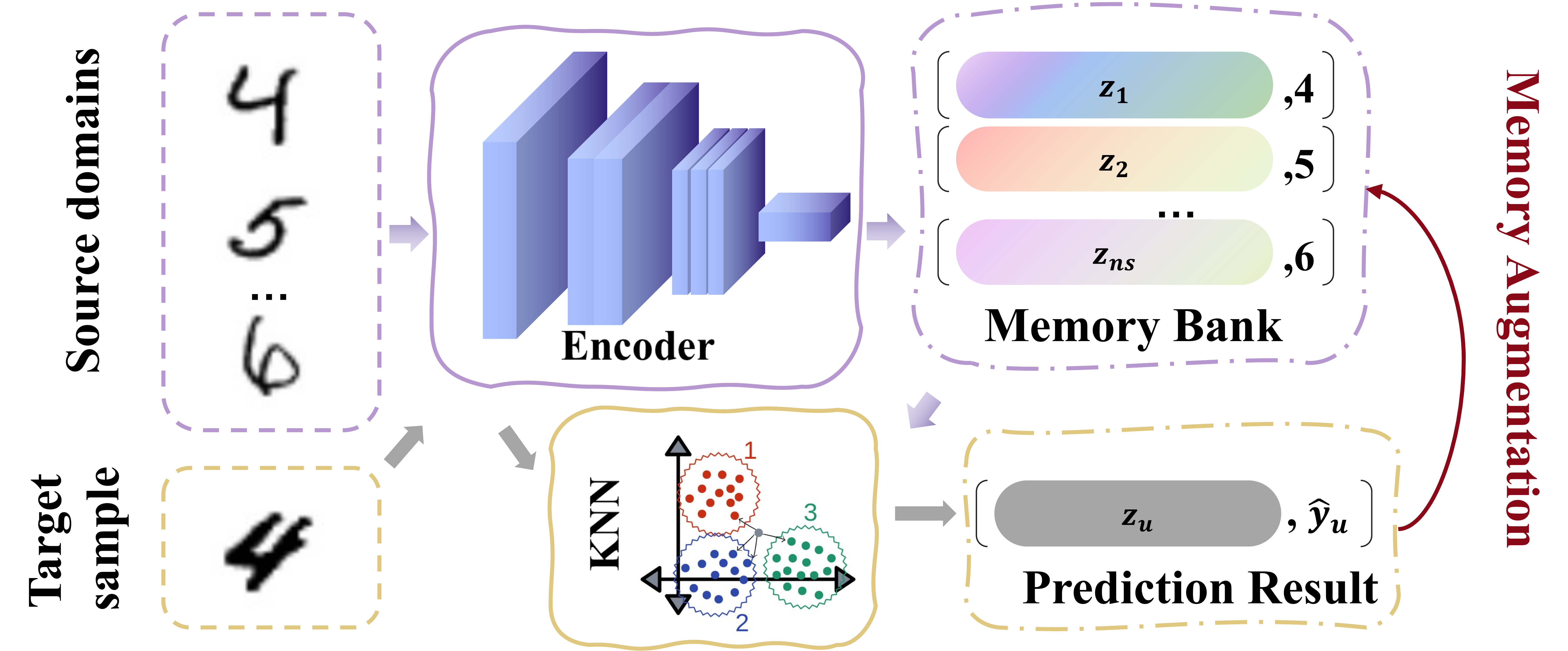}
    \vspace{-4mm}
    \label{fig:teaser}
    }
   \subfigure[Strong knowledge expandability of \knn. When our model is trained on a domain $d_0$ and adapted to target domains $d_1,...,d_5$ successively, advanced TTA methods only bring margin performance improvement, however, \knn and its variants boost the accuracy significantly.]{
    \includegraphics[width=0.9\linewidth]{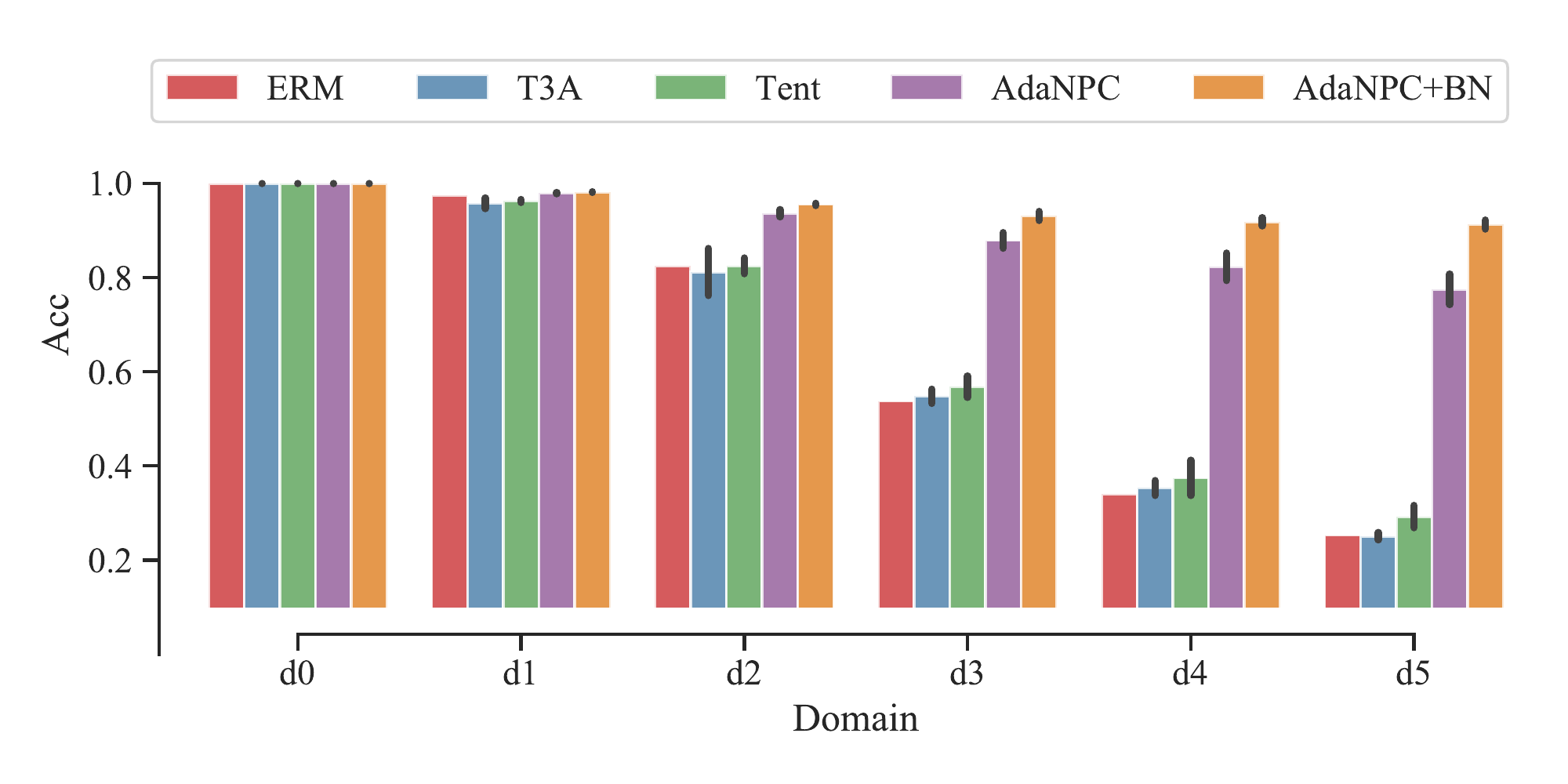}
    \vspace{-4mm}
    \label{fig:rmnist_continue}
    }
    \caption{ \textbf{An illustration example of \knn that proposes to utilize a non-parametric classifier for test-time adaptation.}} 
    \vspace{-2mm}
\end{figure}

In recent studies \cite{zhang2022domain,dubey2021adaptive}, people show that robustifying a model to any unknown distribution is almost impossible without utilizing target samples during inference. The test-time adaptive (TTA) methods are then proposed to utilize target samples with computationally practical constraints. However, current TTA methods suffer from several drawbacks. (1) \textbf{Overhead computation:} existing TTA methods either need batches of target data for gradient updates~\cite{sun2020test,wang2020tent,zhang2021test} and/or an 
additional model for fine-tuning~\cite{sun2020test,dubey2021adaptive,zhang2022domain}, which are prohibited when target sample arrives one by one in the \textit{online} manner. 
(2) \textbf{Domain forgetting:} existing TTA methods require making changes in the trained model. The model would gradually lose the prediction ability of the training domains, indicating that some knowledge loss is inevitable. This issue is especially significant when conducting inference for a series of domains. Let us consider a simple test on Rotated MNIST dataset~\cite{ghifary2015domain}, we perform test-time adaptation to $d_1,d_2,...,d_5$ serially using the latest TTA methods, T3A~\cite{iwasawa2021test} and Tent~\cite{wang2020tent}. In~\figurename~\ref{fig:rmnist_continue}, we observe that the generalization ability on $d_5$ of all existing methods is poor even after adaptation in the first four domains. We also summarize the generalizability in the source domain $d_0$ in~\figurename~\ref{fig:rmnist_forget} drops significantly. That is, current TTA methods cannot adapt to a series of online domains and easy to forget historical knowledge.



To this end, we propose a \textit{non-parametric adaptation} approach, debuted \knn. In particular, \knn trains the model with a $K$ nearest neighbor (KNN) based loss instead of the cross-entropy loss, which minimizes the influence of outliers / irrelevant samples on the potentially noisy training dataset. After training, \knn constructs a memory bank to maintain the trained feature and label pairs of the training dataset. When switching to the inference stage, the feature of a given test sample is first computed using a forward procedure, and then, based on similarity, the top $k$ closed samples in the memory bank are collected to generate a voting prediction. Finally, the new testing pair (feature and predicted label) is added to the memory bank. We illustrate the whole procedure in~\figurename~\ref{fig:teaser}. As dense vectors' searching can be efficiently implemented with logarithmic dependence in total sample size~\cite{johnson2019billion}, the computation cost of \knn in the inference stage is almost the same as a single forward pass and \textit{is significantly smaller than backward gradient updates.} On the other hand, our approach separates the feature extraction procedure and individual sample memorization. It facilitates us to maintain the information among source and target domains simultaneously. The main contributions of this paper are:
\begin{figure}
    \centering
    \includegraphics[width=0.4\textwidth]{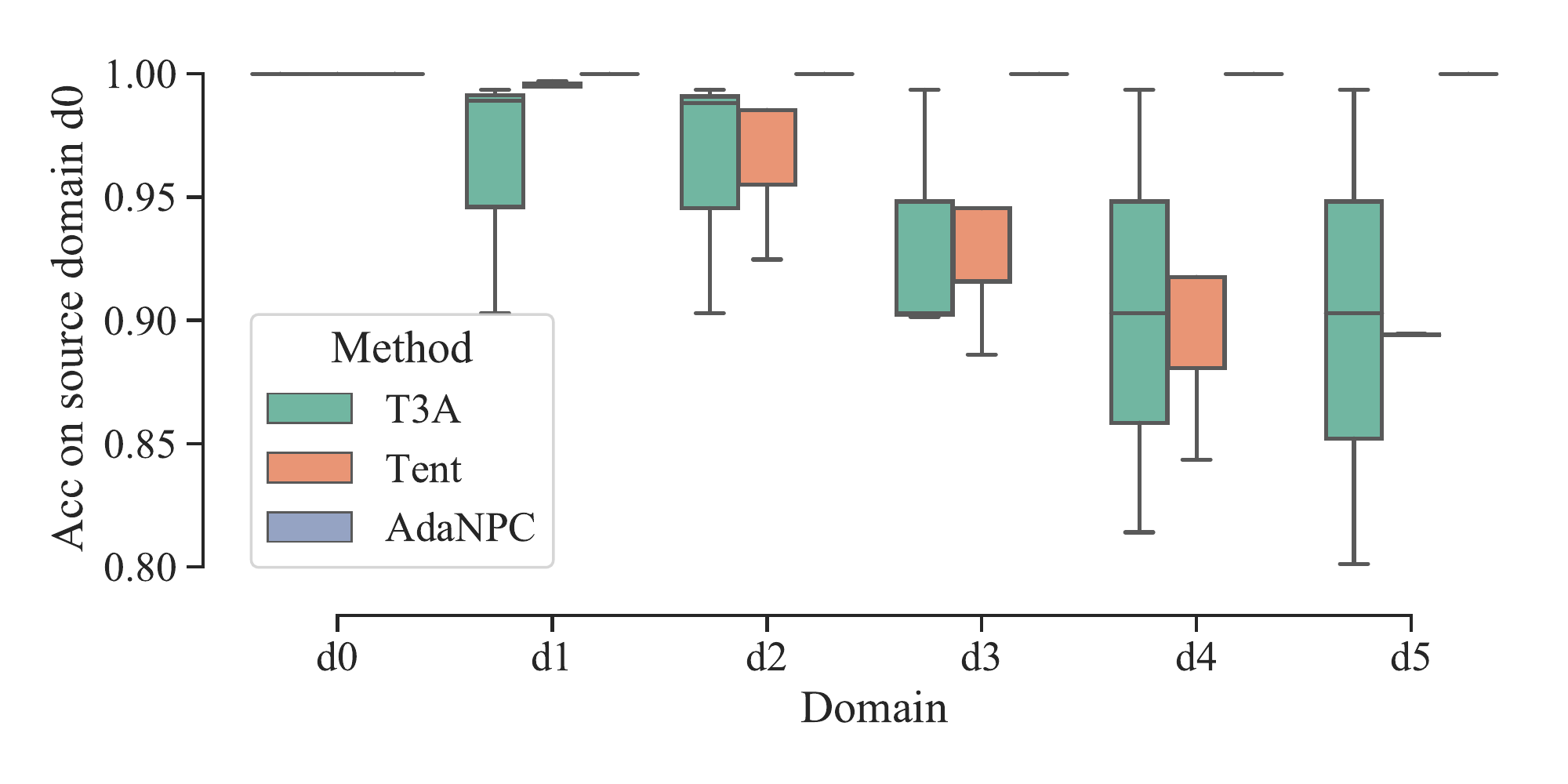}
    \vspace{-0.3cm}
    \caption{\textbf{Re-evaluation on the source domain $d_0$ after the model is adapted to unseen target domain $d_i$.} When adapted to more target domains, the performance of existing TTA methods on $d_0$ drops drastically while \knn always attains a high  accuracy.}
    \label{fig:rmnist_forget}
    \vspace{-0.3cm}
\end{figure}

1. We investigate a non-parametric paradigm to perform test-time adaptation by storing features and predicted pseudo-labels of target instances. The proposed \knn can be incorporated with any representation learning models.

2. We {theoretically} derive target domain error bounds under both the covariate shift setting and the posterior shift setting. Our theoretical results show that a non-parametric classifier can explicitly reduce the domain divergence and makes the target error bound tighter; Besides, \knn, which incorporates online target instances into the memory bank, will further reduce the target risk.

3. We perform extensive {experiments} on $5$ popular OOD benchmarks with $3$ different backbones, where the results show that \knn (1) achieves competitive generalization performance in both target and source domains; (2) beats most existing test-time adaptation methods by a large margin; (3) using non-parametric classifier attains faster convergence and performs well even without fine tuning, which is promising for large pre-trained models.
    
4. \knn has some unique benefits, such as \textit{interpretability:} by analyzing the neighborhood samples chosen by \knn, we can interpret how past knowledge is used for inference results, and \textit{knowledge expandability:} \knn stores all the data features that have been seen and avoid catastrophic forgetting.

\vspace{-0.1cm}
\section{Related Work}
\vspace{-0.1cm}
\label{sec:related}

\textbf{Test-Time adaptive methods} are recently proposed to utilize target samples. The taxonomy of Test-Time adaptive methods is summarized in Appendix Table~\ref{tab:test_methods}, where Test-Time Training methods design proxy tasks during tests such as self-consistence~\cite{zhang2021test}, rotation prediction~\cite{sun2020test} and need extra models; Test-Time Adaptation methods adjust model parameters based on unsupervised objectives such as entropy minimization~\cite{wang2020tent,zhang2021memo} or update a prototype for each class~\cite{iwasawa2021test}. Domain adaptive method~\cite{dubey2021adaptive} needs additional models to adapt to target domains. Both the Test-Time adaptation methods and the domain-adaptive methods need batches of target samples. Single sample generalization methods are recently proposed, which need to learn an adapt strategy from source domains, i.e.,~\cite{xiao2022learning} formulates the adaptation to the single test sample as a variational Bayesian inference problem and uses the meta-learning pipeline to learn the adaptative strategy; \cite{zhang2022domain} introduces specific classifiers for different domains and adapt the voting weight for test samples dynamically. All aforementioned approaches either require accessing the batch of data or need gradient computation to update the models, which is computationally inefficient especially in the online setting with steaming data. Our proposed \knn performs Test-Time adaptation by storing the test features and predicted labels and doesn't have those restrictions. See more related work in Appendix~\ref{sec:app_related}.


\vspace{-0.1cm}
\section{Methods}\label{sec:method}
\vspace{-0.1cm}

\textbf{Problem setting.} In domain generalization (DG), we consider a source domain dataset ${D}_S=\{(x_s^i,y_s^i)\}_{i=1}^{n_s}$ is collect from different environments or domains, where $(x_s^i,y_s^i)$ is sampled i.i.d. from the distribution $\mathbbm{D}_S$ and $n_s$ is the total sample size. The DG aims to train a predictor $\hat{f}$ on source domain dataset $D_S$ and to perform well on a testing unseen dataset $D_U=\{(x_u^i)\}_{i=1}^{n_u}$ that sampled from a distribution $\mathbbm{D}_{U}$, which is inaccessible during training. 
We formally define the classification error and its variants on $\mathbbm{D}_U$, whereas definitions for other domains are the same.
\begin{definition}
    (\textbf{Regression function}.)  In binary classification setting\footnote{Theories and examples in this work consider binary classification for easy understanding and can be easily extended to multi-class setting.}, given a distribution $\mathbbm{D}_{U}$,  the regression functions are defined to represent the conditional distributions.
\begin{equation}
\begin{aligned}
    \eta_{U}(x)= P_{x\sim \mathbbm{D}_U}(Y=1|X=x).
\end{aligned}
\end{equation}
(\textbf{Classification error.}) The error of hypothesis $\hat{f}\in\mathcal{H}:\mathcal{X}\rightarrow\{0,1\}$ under the distribution $\mathbbm{D}_U$ is defined as 
\begin{small}
\begin{equation}
\begin{aligned}
\epsilon_U(\hat{f})&=\mathbb{E}_{(x,y)\sim\mathbbm{D}_U}[|\hat{f}(x)-y|]=P_U(\{(x,y):\hat{f}(x)\neq y\}) 
\end{aligned}
    \label{equ:excess}
\end{equation}
\end{small}
where $P_U(A)$ denotes the probability of an event $A$ in $\mathbbm{D}_U$. 
\end{definition}
\begin{definition}
    (\textbf{excess error and its dual form}.) Given the Bayes classifier under $\mathbbm{D}_U$: $f_U^*(x)= \mathbb{I}\{\eta_U(x)\geq 1/2\}$, the excess error of $\hat{f}$ is defined as
\begin{equation}
\begin{aligned}
\mathcal{E}_U(\hat{f})&=\epsilon_U(\hat{f})-\epsilon_U(f_U^*)\\
&=2\mathbb{E}_{x\sim \mathbbm{D}_U}\left[\left|\eta_U(x)-\frac{1}{2}\right|\mathbb{I}\{\hat{f}(x)\neq f_U^*(x)\}\right] 
\end{aligned}
\end{equation}
\end{definition}
We then introduce the proposed method using the following pipelines: representation learning, making predictions, test-time memory augmentation, and some useful tricks.

\textbf{Learning representative features.} Let $h_\theta(\cdot)$ be the encoder parameterized by $\theta$. We denote $B_{\theta,D_S}(x):= \{a^{(i)}\}_{i=1}^{n_S}$ as the ordered index set in the source domain dataset $D_S$ for any $x$, formally as follows:
satisfying $\|h_{\theta}(x_{a^{(1)}})-h_{\theta}(x)\|_2\le \|h_{\theta}(x_{a^{(2)}})-h_{\theta}(x)\|_2\le \cdots \le \|h_{\theta}(x_{a^{(n_s)}})-h_{\theta}(x)\|_2$.
We denote $B_{k,\theta,D_S}\subseteq B_{\theta,D_S}$ as the subset only contains the first $k$ elements in $B_{\theta,D_S}$. 

We consider optimizing the KNN loss function $\mathcal{L}_{KNN}$:
\begin{small}
\begin{align}
\min_{\theta}\frac{-1}{n_s}\sum_{i} \log \frac{\sum_{j\in B_{k,\theta,D_S}(x_i)} \exp\left(w_{ij}/\tau \right)\mathbb{I}\{y_i=y_j\}}{\sum_{j\in B_{k,\theta,D_S}(x_i)} \exp\left(w_{ij}/\tau \right)}, 
\label{equ:knn_training}
\end{align}
\end{small}
\noindent where $w_{ij}= \frac{h_{\theta}(x_i)^{\top}h_{\theta}(x_j)}{\| h_{\theta}(x_i)\|_2 \|h_{\theta}(x_j)\|_2}$ is the consine similarity between $h_{\theta}(x_i)$ and $h_{\theta}(x_j)$ and $\tau>0$ is the temperature parameter to scale the influence of $w_{ij}$.

With this configuration, features with the same label are zoomed closer and different labeled features are pushed away. Compared to cross-entropy loss, the \myref{equ:knn_training} training paradigm leads to better representations, which is verified in~\cite{feng2021rethinking} and our experiments. The optimization of \myref{equ:knn_training} is highly nontrivial as $B_{k,\theta,D_S}$ changes with model parameters $\theta$ in a non-differentiable manner. In this paper, we adopt an EM algorithm to solve it approximately. We only periodically update $B_{k,\theta,D_S}$ and keep them fixed for the remaining time, in which we can easily apply the standard optimization schemes in PyTorch or TensorFlow. In practice, one may also consider ignoring this step and directly using the pre-trained model with conventional methods (e.g., ERM~\cite{vapnik1998statistical}, IRM~\cite{arjovsky2020invariant}, and CORAL~\cite{sun2016deep}). In the experimental section, we show that we don't even need to fine-tune the pre-trained model in the source domain, and \knn can still achieve good generalization performance.

\textbf{Making prediction by non-parametric classifier.} Given an unseen sample $x_u$, we denote the prediction from \knn as follows:
\begin{align}
    \eta_k(x_u)=\mathrm{softmax}(\sum_{j\in B_{k,\theta,\mathcal{M}}(x_u)} w_{uj} \mathbf{1}\{y_j\}),\label{eq:xue_oct_31:1}
\end{align}
 where $ \mathbf{1}\{y_j\}$ is the one-hot class label of $x_j$ for $j\ge 1$, $w_{uj}$ is the cosine similarity between $h_{\theta}(x_u)$ and $h_{\theta}(x_j)$, and $\mathcal{M}$ is the memory bank and will be specified later. Intuitively, the predictor from \myref{eq:xue_oct_31:1} can be viewed as a voting procedure among the memorized samples similar to $x_u$. We first collect the $k$ closest samples to $x_u$ in the memory bank. The cosine similarity $w_{uk}$ is then computed between $k$ selected samples and $x_u$. The final decision is the label with the largest summed weights. In this predictor, we explicitly use the samples in the memory bank. It gives us better interpretability than conventional approaches.

\textbf{Non-parametric test-time adaptation by memory augmentation.}  At the beginning of the evaluation procedure, all source domain images are embedded to make up the \textbf{Memory bank} $\mathcal{M}=\{(h_{\theta}(x_i),y_i)\}_{i=1}^{n_s}$. During inference, assume each $x_u$ arrives one by one online and we must report the classification result once $x_u$ arrives. After classification, if the prediction confidence of $\eta_k(x_u)$ is greater than a predefined margin, then the memory set will be augmented by $\mathcal{M}=\mathcal{M}+(h_{\theta}(x_u),\eta_k(x_u))$. The overall pipeline of \knn is shown in~\figurename~\ref{fig:teaser}.

\textbf{Some useful tricks.} \textbf{1. BN retraining.} The performance of a non-parametric classifier highly depends on the model representation, to attain more powerful representations and maintain the simplicity of \knn, we optionally add \textit{one} BN layer before the classifier. Then during the evaluation, only the BN layer parameter will be re-trained by minimizing the prediction entropy of $\eta_k(x_u)$. \textbf{2. Efficient memory bank construction $\mathcal{M}$.} Although there are many advanced techniques for memory construction, such as momentum update in MoCo~\cite{he2020momentum} or Faiss~\cite{johnson2019billion}, for presentation simplicity consideration, we only adopt the simplest method via computation all pair-wise distance here and hold the usage of advance KNN searching algorithm for the future work. 
Moreover, to facilitate the training efficiency on the large source dataset, 
we consider to construct $\mathcal{M}$ with a relatively small size for faster training, where $|\mathcal{M}|\ll n_s$. $\mathcal{M}$ during training will be updated by the first-in-first-out (FIFO) strategy by each minibatch representation.



\section{Theoretical Analysis}
In this section, we theoretically verify that using KNN as our classifier can explicitly reduce domain divergence. Besides, incorporating target instances, namely the non-parametric Test-Time adaptation, will further reduce the unseen target error. Before we discuss the major results, we first state some necessary assumptions and notions. Refer to Appendix~\ref{app:proof} for missing proofs and detailed explanations.

\begin{assump}
\textit{(Strong Density Condition)} Given parameter $\mu_-,\mu_+,c_\mu,c_\mu^*,r_\mu>0$, we assume $\mathbbm{D}_S$, $\mathbbm{D}_U$ are absolutely continuous with respect to the Lebesgue measure in $\mathbb{R}^d$, and $\mathcal{B}(x,r)=\{x':\parallel x'-x\parallel\leq r\}$ is the ball centered on $x$ with radius $r$. We assume that $\forall x_u\in \mathbbm{D}_U$ and $r\in(0,r_{\mu}]$ we have
\begin{gather}
\lambda[\mathbbm{D}_S\cap \mathcal{B}(x_u,r)] \geq c_\mu\lambda[\mathcal{B}(x_u,r)]\notag\\
\lambda[\mathbbm{D}_U\cap \mathcal{B}(x_u,r)] \geq c_\mu^* \lambda[\mathcal{B}(x_u,r)]\notag\\
\mu_-<\frac{d \mathbbm{D}_S}{d \lambda} < \mu_+; \mu_-<\frac{d \mathbbm{D}_U}{d \lambda} < \mu_+,\notag
\end{gather}
where $\lambda$ is the Lebesgue measure in Euclidean space.
\label{assump1}
\end{assump}

Strong Density Condition is a commonly assumed condition when analyzing KNN classifier (e.g., \cite{audibert2007fast,cai2021transfer}). 
Intuitively, Assumption \ref{assump1} requires the divergence between supports of $\mathbbm{D}_S$ and $\mathbbm{D}_U$ being bounded. When $c_\mu=1$, for each $x\in \mathbbm{D}_U$, its neighbor ball $\mathcal{B}(x,r)$ is completely within $\mathbbm{D}_S$. In contrast, when $c_\mu\approx 0$, $\mathcal{B}(x,r)$ and $\mathbbm{D}_S$ are nearly disjoint. We then consider two common assumptions that parameterize the behavior of $\eta_U(x)$.

\begin{assump}
    (Smoothness) Let $\eta_U$ be the classification function  and $C$ be a positive constant. For all feasible $x,x^{\prime}$ we have $|\eta_U(x)-\eta_U(x')|\leq C\parallel x-x' \parallel$.
\label{assump:xue:2:Nov_11_2022}
\end{assump}
Assumption \ref{assump:xue:2:Nov_11_2022} describes that $\eta_U$ is Lipschitz continuous. Our analysis is capable with weaker condition, such as $(\alpha, C_\alpha)$-H\"{o}lder condition~\cite{cai2021transfer} for some $\alpha\in(0,1]$ and $C_\alpha>0$. For notation simplicity, we hold it for future work.

\begin{assump}
    (Low Noise Condition). Let $\beta,C_\beta$ be positive constants and we assume $\mathbbm{D}_U$ satisfies $P_{x\sim \mathbbm{D}_U}\left(\left| \eta_U(x)-\frac{1}{2} \right|<t\right)\leq C_\beta t^\beta$ for all $t>0$.
    \label{define_noise}
\end{assump}

The low noise condition is first proposed in \cite{tsybakov2004optimal}, which is also named margin assumption~\cite{cai2021transfer}. The assumption places a constraint on $\eta_U$ around $\eta_U(x)\approx 1/2$. A larger $\beta$ pushes $\eta_U$ far from $1/2$ and then the classification task will be easier.


\subsection{KNN classifier reduces domain divergence}
We characterize domain divergence reduction of KNN classifier in the following proposition \ref{prop:1}. 

\begin{prop}
 Let $\mathbbm{D}_S$ and $\mathbbm{D}_U$ be the source and target domain respectively. Per Assumptions 1 and 2,  the risk of hypothesis $\hat{f}$ on the unseen target domain is bounded by
\begin{equation}
\begin{aligned}
\epsilon_U(\hat{f})\leq \kappa_s+\epsilon_s(\hat{f})+\mathcal{O}\left(\mathcal{W}(\mathbbm{D}_S,\mathbbm{D}_U)\right),\label{eq:xue:Nov_10}
\end{aligned}
\end{equation}
where $\mathcal{W}(\cdot,\cdot)$ is the Wasserstein $1$-distance , $\kappa_{S}=\min_{f}\epsilon_U(f)+\epsilon_{S}(f)$, and we use $\mathcal{O}(\cdot)$ to hide the constant dependence.

Furthermore, if we switch from $\mathbbm{D}_{S}$ to a sampled distribution $\Omega$ of $\mathbbm{D}_{S}$ around the neighborhood of $\mathbbm{D}_{U}$, i.e., $\Omega:=\bigcup_{x\in \mathbbm{D}_U} \mathcal{B}(x,r_x)$ with $r_x \le r_{\mu}$ such that each $\mathcal{B}(x,r_x)$ contains exact $k$ elements, and assume that the unseen distribution $\mathbbm{D}_U$ is finite with cordiality $n_{\mathbbm{D}_u}$. The risk of hypothesis $\hat{f}$ on $\mathbbm{D}_U$ is then improved to
\begin{equation}
\begin{aligned}
\epsilon_U(\hat{f})\leq \kappa_{\Omega}+\epsilon_\Omega(\hat{f})+\mathcal{O}\left(\left(\frac{2k}{c_\mu\mu_-\pi_dn_s}\right)^{1/d}\right),
\end{aligned}
\label{lemma:bound_omega}
\end{equation}
with probability at least $1 - \exp(-\frac{k}{4} + \log n_{\mathbbm{D}_u})$, where $\kappa_{\Omega}=\min_f\epsilon_U(f)+\epsilon_\Omega(f)$ and $d$ is the feature representation dimension.
\label{prop:1}
\end{prop}



Inequality \myref{eq:xue:Nov_10} is adopted from \cite{shen2018wasserstein}, which indicates the error of a given hypothesis $\hat{f}$  bounded by three terms, the minimized combined error $\kappa_S$, the error in the source domain $\epsilon_S(\hat{f})$ and a term on the order of a constant term $\mathcal{W}(\mathbbm{D}_S,\mathbbm{D}_U)$.
When source domain $\mathbbm{D}_S$ is far away from the target domain $\mathcal{D}_U$, the $\mathcal{O}(\mathcal{W}(\mathbbm{D}_S,\mathbbm{D}_U))$ becomes the dominating quantity and leads to loose upper bound.

When switching to the non-parametric classifier, in \myref{lemma:bound_omega} we replace $\mathcal{O}(\mathcal{W}(\mathbbm{D}_S,\mathbbm{D}_U))$ by a quantity explicitly decaying in $n_s$. Intuitively, by constructing $\Omega$, we only keep the samples in the source domain with enough similarity to the target domain and it would naturally short the distance from $\mathcal{W}(\mathbbm{D}_S,\mathbbm{D}_U)$ to $\mathcal{W}(\Omega,\mathbbm{D}_S)$. We also want to highlight that Proposition \ref{prop:1} should imply the non-parametric classifier may be able to take more benefits from the large pretraining source dataset (\figurename~\ref{fig:rmnist_ratio}). See the Appendix for a detailed discussion of the influence of $k$ and $c_\mu$.


\subsection{\knn further reduces the target risk}
 In this section, we develop the target excess error bounds under the covariate-shift and the posterior-shift settings, which further articulate all factors that affect the performance of our algorithm (Proposition~\ref{prop:knn_S}) and the benefits of using online target data (Proposition~\ref{prop:knn_SU}).

\begin{prop}
We consider $\eta_U$ and $\eta_S$ to be the KNN predictor in form of \myref{eq:xue_oct_31:1} with all $w_i$ fixed as $\frac{1}{k}$. Per assumptions 1-3, the following results hold with high-probability when choosing $k = \mathcal{O}(\log n_s)$.

\textbf{Under the covariate-shift setting}, we have $\eta_U=\eta_S=\eta$ for the source and target domains, and
\begin{equation}
\begin{aligned}
\mathcal{E}_U(\hat{f})&\leq \mathcal{O}\left(\left(\frac{1}{k}\right)^{1/4}+  \left(\frac{k}{c_\mu n_s}\right)^{1 /{d}} \right)^{{1+\beta}}\\
&=\mathcal{O}\left(\left(\frac{1}{\log n_s}\right)^{1/4} + \left(\frac{\log n_s}{c_\mu n_s}\right)^{1 /{d}} \right)^{{1+\beta}}.
\end{aligned}
\end{equation}
\textbf{Under the posterior-shift setting}, the regression functions $\eta_U$ and $\eta_S$ are different and
\begin{small}
\begin{equation}
\mathcal{E}_U(\hat{f})\leq \mathcal{O}\left(\left(\frac{1}{\log n_s}\right)^{1/4}+  \left(\frac{\log n_s}{c_\mu n_s}\right)^{1 /{d}} +C_{ada} \right)^{{1+\beta}},
\end{equation}
\end{small}
where $\sup_{x_u\in \mathbbm{D}_U} |\eta_S-\eta_U|\leq C_{ada}$.
\label{prop:knn_S}
\end{prop}

We make a few remarks on the excess risk upper bound.

1. The upper bound is affected by $k,c_\mu,n_s$, which is similar to the discussion of Proposition~\ref{prop:1}. Differently, when setting $k = \mathcal{O}(\log n_s)$, the excess error bound reduces to $0$ under the covariate-shift setting  when $n_s\rightarrow \infty$ in a high probability manner.

2. Proposition~\ref{prop:knn_S} shows a trade-off on the choice of $k$. Although a small $k$ reduces the domain divergence or representing similarity $\left({k}/{c_\mu n_s}\right)^{1 /{d}}$, it is well known that the model will become too specific and fails to generalize well. 

3. When regression functions are different, an additional term is introduced in the bound, namely the adaptivity gap $\sup_{x_u\in \mathbbm{D}_U} |\eta_S(x_u)-\eta_U(x_u)|$, which measure the difference of two regression functions. The gap can be estimated and reduced by existing methods~\cite{zhang2022domain}.

4. For presentation simplification, we use the equal weighted $w_{uj}$ instead of 
cosine similarity between feature representations. In fact by choosing proper $k$, all cosine similarity values can be safely assumed lower bounded on $\mathcal{O}(k^{-\delta})$ for some $\delta>0$. Therefore one may introduce extra assumption on the lower bound of $w_{uj}$ to obtain a extended version of Proposition~\ref{prop:knn_S} with adaptive $w_{uj}$.

As discussed in Section~\ref{sec:method}, the proposed \knn is a special kind of Test-Time adaptation method that can utilize the online target samples to improve prediction generalization. We next theoretically verify that, by incorporating the online target samples into the KNN memory bank, the excess error bound is further reduced.

\begin{prop}
Denote $n_u$ as the number of target instances in the KNN memory bank
 and the KNN classifier finds the $k_s$ nearest neighbors in ${D}_S$ and $k_u$ nearest neighbors in ${D}_U$ during inference with $k_s+k_u=k$. Per Assumptions 1-3, \textbf{under the posterior-shift setting}, 
\begin{equation}
\begin{aligned}
\mathcal{E}_U(\hat{f})\leq \mathcal{O}\left(\left(\frac{1}{k}\right)^{1/4}+\frac{k_s}{k} \left(\frac{k_s}{c_\mu n_s}\right)^{1 /{d}} +\right.\\
\left.\frac{k_u}{k} \left(\frac{k_u}{c_\mu^* n_u}\right)^{1 /{d}} + \frac{k_s}{k} C_{ada} \right)^{{1+\beta}}
\end{aligned}
\end{equation}
where $\sup_{x_u\in \mathbbm{D}_U} |\eta_S-\eta_U|\leq C_{ada}$ and similar results also hold under the covariate-shift setting.
\label{prop:knn_SU}
\end{prop}

We want to highlight that the above error bound is tighter than the case without updating the memory bank. The detailed discussion is deferred to section~\ref{sec:app_target_sample} in Appendix.

\section{Experiments}\label{sec:exp}

\begin{table*}[t]
\centering
\adjustbox{max width=0.8\textwidth}{%
\setlength{\tabcolsep}{7.25pt}
\centering
\begin{tabular}{lcccccc}
\toprule
Method             & \textbf{RMNIST}     & \textbf{VLCS}             & \textbf{PACS}              & \textbf{DomainNet} & \textbf{TerraIncognita}       & \textbf{Avg}              \\
\midrule
ERM~\cite{vapnik1998statistical}                                  & 97.8 $\pm$ 0.1            & 77.6 $\pm$ 0.3            & 86.7 $\pm$ 0.3            & 41.3 $\pm$ 0.1    & 53.0 $\pm$ 0.3         & 71.3                      \\
IRM~\cite{arjovsky2020invariant}                                  & 97.5 $\pm$ 0.2            & 76.9 $\pm$ 0.6            & 84.5 $\pm$ 1.1            & 28.0 $\pm$ 5.1  & 50.5 $\pm$ 0.7          & 67.5                      \\
GDRO~\cite{sagawa2020distributionally}                              & 97.9 $\pm$ 0.1            & 77.4 $\pm$ 0.5            & 87.1 $\pm$ 0.1            & 33.4 $\pm$ 0.3      & 52.4 $\pm$ 0.1      & 69.6                      \\
CORAL~\cite{sun2016deep}                               & 98.0 $\pm$ 0.0            & 77.7 $\pm$ 0.2            & 87.1 $\pm$ 0.5            &  41.8 $\pm$ 0.1   & 52.8 $\pm$ 0.2          & 71.5                      \\
DANN~\cite{ganin2016domain}                               & 97.9 $\pm$ 0.1            & 79.7 $\pm$ 0.5            & 85.2 $\pm$ 0.2            & 38.3 $\pm$ 0.1     & 50.6 $\pm$ 0.4        & 70.3                     \\
MTL~\cite{blanchard2021domain}                            & 97.9 $\pm$ 0.1            & 77.7 $\pm$ 0.5            & 86.7 $\pm$ 0.2            &  40.8 $\pm$ 0.1      & 52.2 $\pm$ 0.4      & 71.1                      \\
SagNet~\cite{nam2021reducing}                               & 97.9 $\pm$ 0.0            & 77.6 $\pm$ 0.1            & 86.4 $\pm$ 0.4            &  40.8 $\pm$ 0.2      & 52.5 $\pm$ 0.4       & 71.1                     \\
ARM~\cite{zhang2021adaptive}                                & 98.1 $\pm$ 0.1            & 77.8 $\pm$ 0.3            & 85.8 $\pm$ 0.2            &  36.0 $\pm$ 0.2      & 51.2 $\pm$ 0.5       & 69.8                      \\
VREx~\cite{krueger2021out}                                 & 97.9 $\pm$ 0.1            & 78.1 $\pm$ 0.2            & 87.2 $\pm$ 0.6            & 30.1 $\pm$ 3.7      & 51.4 $\pm$ 0.5      & 68.9                     \\
Fish~\cite{shi2022gradient}                                & 97.9 $\pm$ 0.1            & 77.8 $\pm$ 0.6            & 85.8 $\pm$ 0.6            &  \textbf{43.4 $\pm$ 0.3}    & 50.8 $\pm$ 0.4         & 71.1                     \\
Fishr~\cite{rame2022fishr}                                 & 97.8 $\pm$ 0.1            & 78.2 $\pm$ 0.2            & 86.9 $\pm$ 0.2            &  41.8 $\pm$ 0.2       & 53.6 $\pm$ 0.4      & 71.7                     \\
\rowcolor{Gray}
\knn               & \textbf{{98.5 $\pm$ 0.1}}    &  79.5 $\pm$ 2.4 & {88.8 $\pm$ 0.1}    &   { 42.9} $\pm$ 0.5   & \textbf{53.9 $\pm$ 0.3}  &   {72.7}    \\
\rowcolor{Gray}
\knn+BN                & {{98.4 $\pm$ 0.1}}    &  \textbf{80.2 $\pm$ 0.2} & \textbf{88.9 $\pm$ 0.1}    &   { 43.1} $\pm$ 0.8 &  \textbf{54.0 $\pm$ 0.1}   &   \textbf{{72.9}}    \\
\bottomrule
\end{tabular}}
 \vspace{-2mm}
\caption{Out-of-distribution generalization performance.}
\label{tab:ood}
\vspace{-0.2cm}
\end{table*}

In our experiments, BN retraining and KNN loss are not used by default for fair comparisons. The model name will be \knn+BN when BN retraining is used and the effect of the KNN loss is verified in the ablation study.

\subsection{Experimental settings}

\textbf{Domain generalization benchmarks and baselines.} We use five popular OOD generalization benchmark datasets: Rotated MNIST~\cite{ghifary2015domain}, PACS~\cite{li2017deeper}, VLCS~\cite{torralba2011unbiased}, TerraIncognita~\cite{beery2018recognition} and DomainNet~\cite{peng2019moment}. We compare our model with ERM \cite{vapnik1998statistical}, IRM \cite{arjovsky2020invariant}, Mixup \cite{yan2020improve}, CORAL \cite{sun2016deep}, DANN \cite{ganin2016domain}, CDANN \cite{li2018deep}, MTL~\cite{blanchard2021domain}, SagNet~\cite{nam2021reducing}, ARM~\cite{zhang2021adaptive}, VREx~\cite{krueger2021out}, RSC~\cite{huang2020self}, Fish~\cite{shi2022gradient},  Fishr~\cite{rame2022fishr}. All experimental settings and baselines follow the Domainbed codebase \cite{gulrajani2021in}. The comparison of \knn with other TTA methods is detailed in the appendix.

See Appendix \ref{sec:data_detail} for more information, including datasets information, model selection, licensing information, hyperparameter search, and the total amount of computing. See Appendix~\ref{sec:addexp} for more experimental results and analysis.

\subsection{Experimental Results}

\begin{figure}[t]
   \subfigure[]{
    \includegraphics[width=0.45\linewidth]{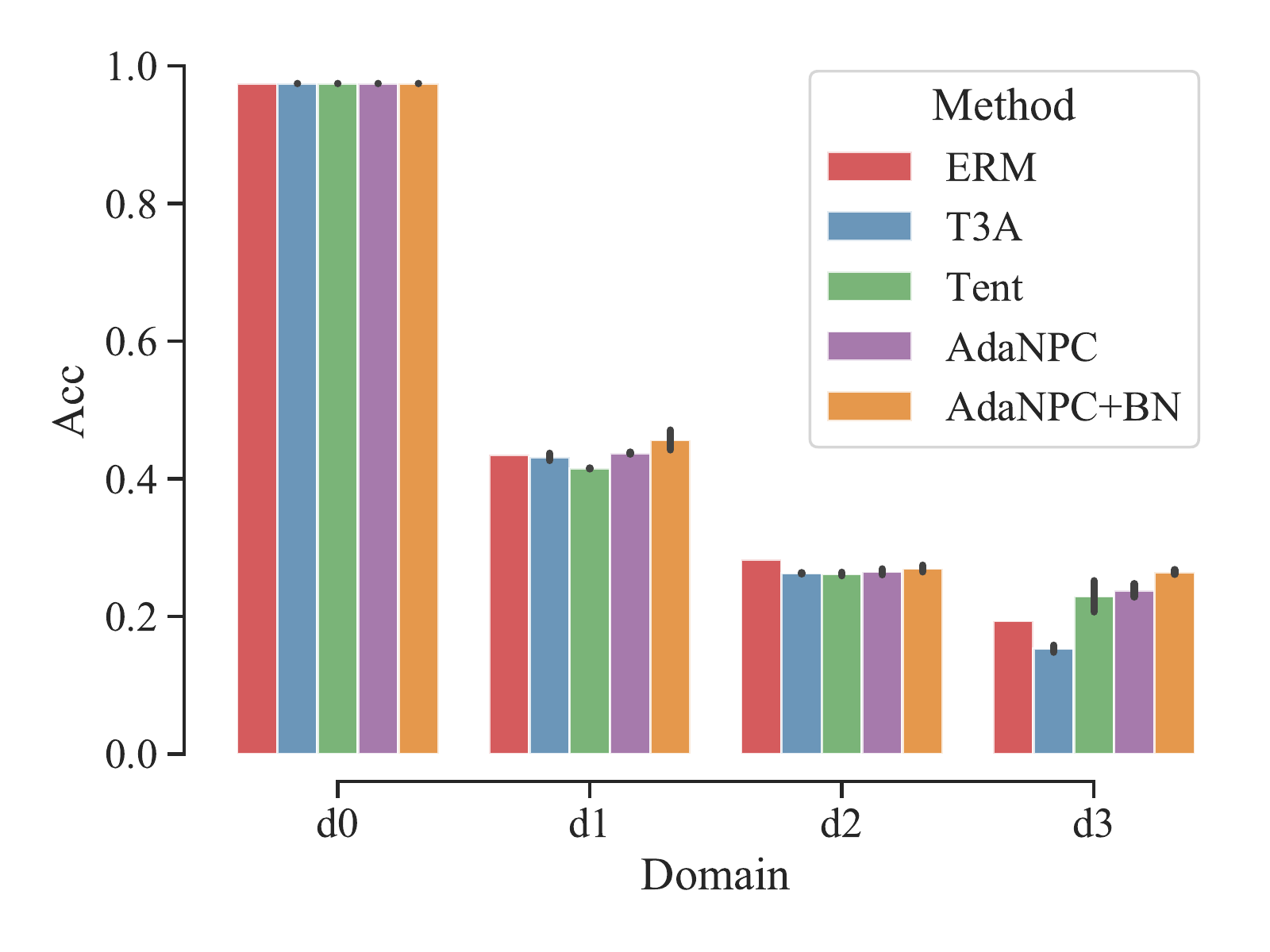}
    \vspace{-4mm}
    \label{fig:terr_coninue}
    }
   \subfigure[]{
    \includegraphics[width=0.45\linewidth]{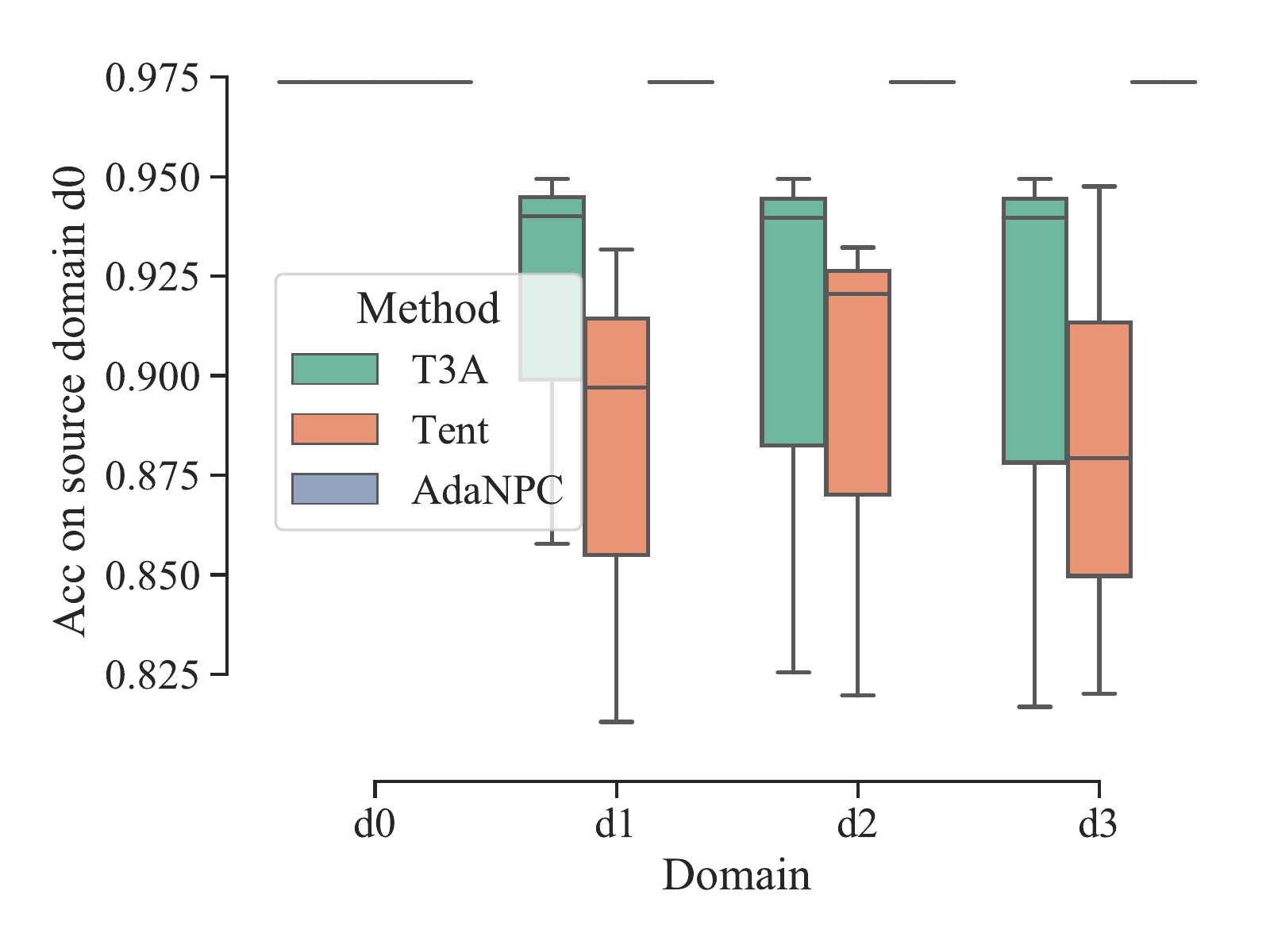}
    \vspace{-2mm}
    \label{fig:terr_forget}
    }
    \caption{\textbf{Successive adaptation results on the TerraIncognita dataset.} (a) Adaptation results on $d_1,d_2,d_3$. (b) Re-evaluation of the adapted model on the source domain $d_0$.} 
     \vspace{-2mm}
\end{figure}
\setlength{\columnsep}{8pt}

\begin{wrapfigure}{r}{0.242\textwidth}
  \begin{center}
  \advance\leftskip+1mm
    \vspace{-0.17in}  
    \includegraphics[width=0.242\textwidth]{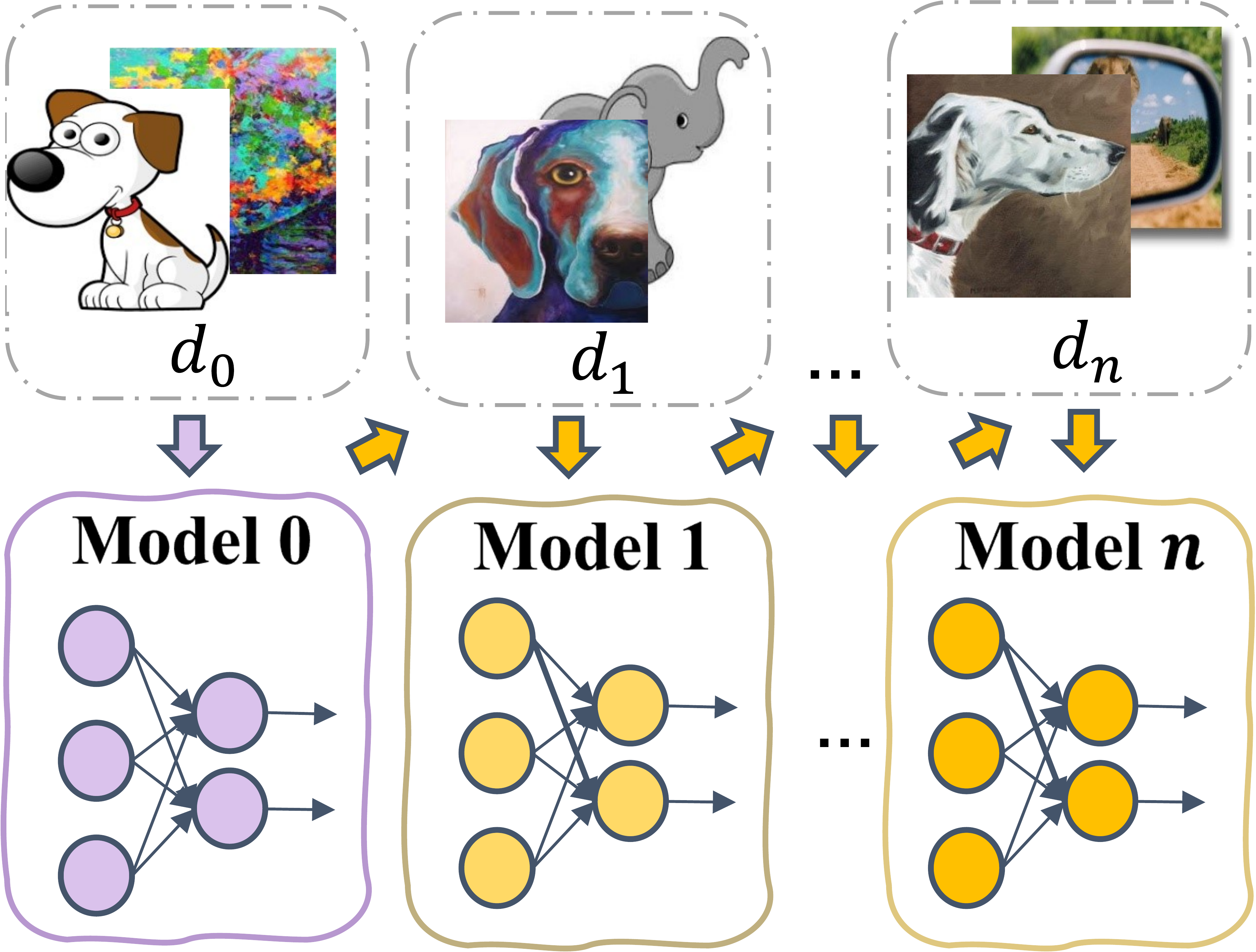}
    \vspace{-0.26in} 
    \caption{ An illustration of successive adaptation setting.}
        \label{fig:succ_dg}
    \vspace{-0.2in} 
  \end{center}
\end{wrapfigure}

\textbf{\knn has strong knowledge expandability.} For practical usage, a deployed model should be adapted to a series of domains, where domain partition is unknown and we should guarantee that the model performs well on both any unseen target sample and samples from the training domains. In this case, we propose a setting named \textit{Successive adaptation} that is more practical. As shown in~\figurename~\ref{fig:succ_dg}, a model (model $0$) that is trained on domain $d_0$ will be adapted to a series of domains. Specifically, model $i$ will be adapted and evaluated on a domain $d_{i+1}$. The results of successive adaptation results on the Rotated MNIST and TerraIncognita are shown in~\figurename~\ref{fig:rmnist_continue} and~\figurename~\ref{fig:terr_coninue}. Results show that the latest TTA method, namely T3A~\cite{iwasawa2021test} and Tent~\cite{wang2020tent} perform marginally above or even worse than the ERM baseline where no test-time adaptation is performed. 
In contrast, the proposed \knn is superior to these methods, especially in the Rotated MNIST dataset $d_5$, where the performance gains brought by \knn compared to ERM are greater than $50\%$. The reason will be that the domain indexes in the Rotated MNIST are continuously changed, namely, the rotation angles are changed from $0^\circ$ to $75^\circ$, which makes the knowledge transfer easier. The continuously changed domain index is also general in real world tasks~\cite{wang2020continuously}, for example, in medical applications, one needs to adapt disease diagnosis and prognosis across patients of different ages, where the age is continuously changed between domains, verifying the practical utility of \knn.

 \textbf{\knn overcomes catastrophic forgetting and reserves superior source performance that is even better than ERM.}~\figurename~\ref{fig:rmnist_forget} and~\figurename~\ref{fig:terr_forget} visualize the source accuracy of model $i$, where the x-coordinate $d_i$ means the model $i$ is evaluated on the source  $d_0$. Both T3A~\cite{iwasawa2021test} and Tent~\cite{wang2020tent} forget the knowledge of the source domains as they adapt to more target domains. That is, if these TTA methods are used for a deployed model after a series of adaptations, we cannot expect that the model will still give a correct prediction on the source domain instances. The tradeoff between adaptivity and source domain accuracy is nonexistent for the proposed \knn, which performs both tasks well. Dur to space limit, we leave the performance over other dataset in the appendix~\ref{app:forget}.

 \textbf{\knn achieves a new State-Of-The-Arts on domain generalization benchmarks.} The average OOD results on all benchmarks are shown in Table~\ref{tab:ood}. We observe consistent improvements achieved by \knn compared to existing algorithms and \textbf{BN retraining} can further boost the generalization capability with few parameter updating. Compared to the advanced DG method Fishr~\cite{rame2022fishr}, which achieves $0.4$ higher average accuracy than ERM, the proposed \knn attains a much larger margin ($1.4$). The results indicate the superiority of \knn in real-world diversity shift datasets.

\begin{table}[t]
\centering
\resizebox{0.8\columnwidth}{!}{%
\begin{tabular}{@{}lccccc@{}}
\toprule
\textbf{Method} & \textbf{A} & \textbf{C} & \textbf{P} & \textbf{S} & \textbf{Avg} \\\midrule
{\color{brown}ResNet18}  & 79.5 & 73.0 & 90.1 & 77.3 & 80.0 \\
\knn & 82.7 & 76.8 & 92.8 & 77.7 & 82.5 \\
\knn+BN & 83.6 & 77.7 & 93.1 & 77.9 & 83.1 \\ \hline
{\color{brown}ResNet50}  & 81.7 & 82.3 & 94.3 & 77.6 & 84.0 \\
\knn & 86.0 & 81.0 & 96.4 & 79.8 & 85.8 \\
\knn+BN & 85.6 & 80.8 & 96.5 & 79.7 & 85.7 \\ \hline
{\color{brown}ViT-B16} & 88.3 & 82.4 & 97.9 & 79.8 & 87.1 \\
\knn & 89.8 & 86.0 & 97.9 & 80.5 & 88.6 \\
\knn+BN & 89.8 & 86.1 & 98.3 & 80.5 & 88.7 \\ \bottomrule
\end{tabular}
}
\caption{OOD accuracy with different backbones on PACS.}\label{tab:backbons}
\vspace{-0.2cm}
\end{table}
\begin{figure*}[t]
\vspace{-0.2cm}
\centering
    \subfigure[]{
    \includegraphics[width=0.255\textwidth]{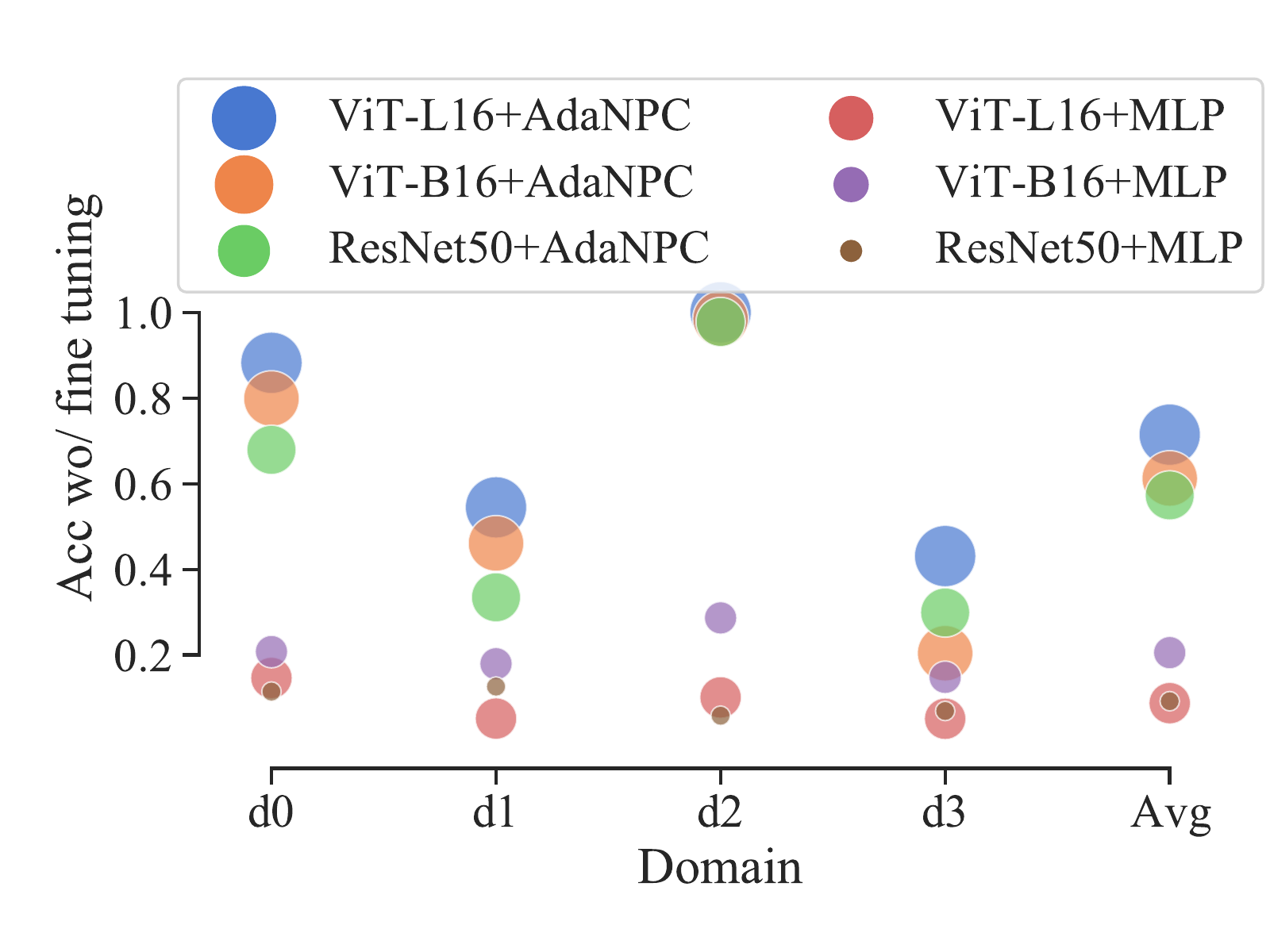}
    \label{fig:pacs_notrain}
    \vspace{-0.1cm}
    }
    \subfigure[]{
    \includegraphics[width=0.255\textwidth]{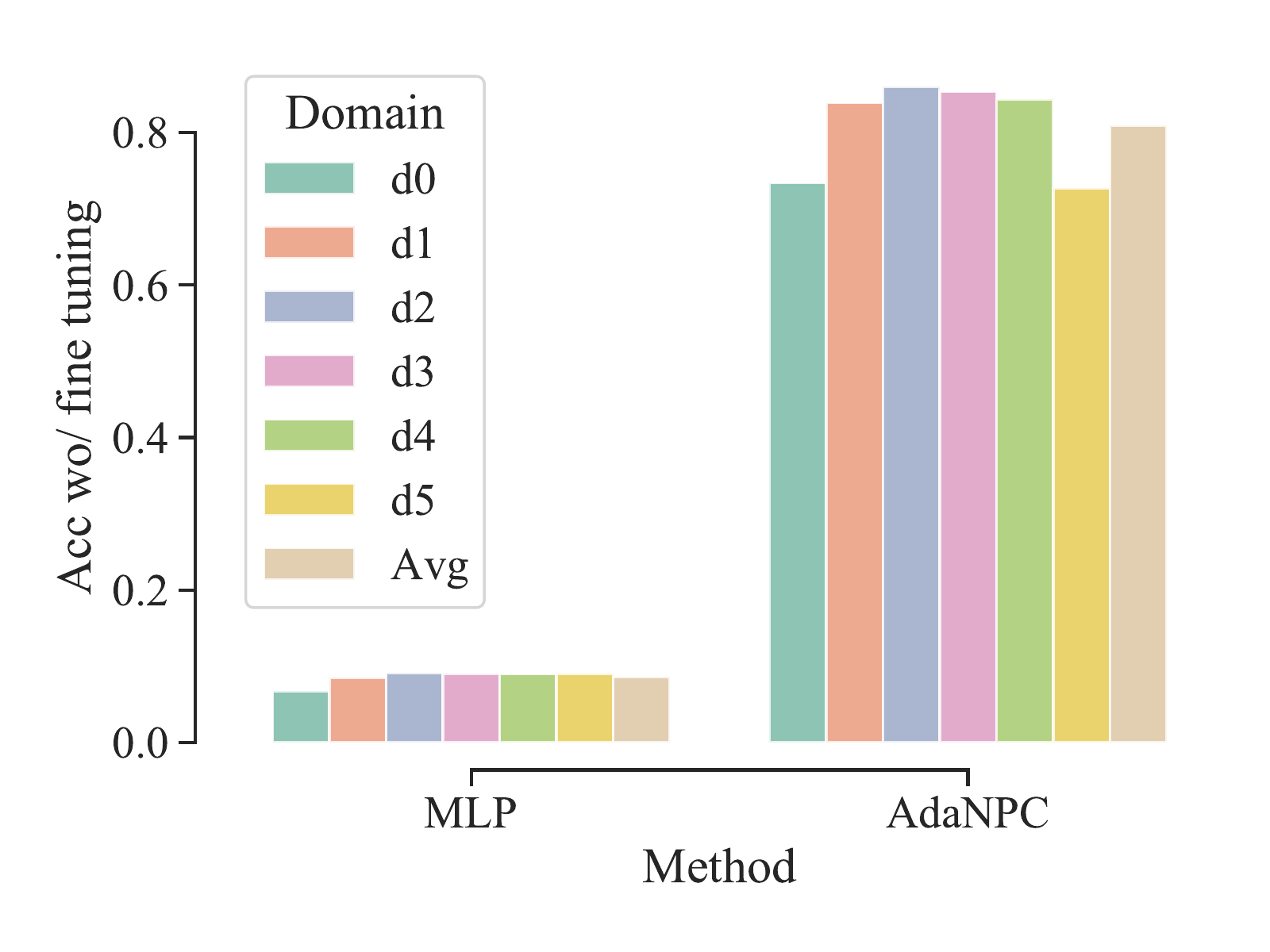}
    \vspace{-0.1cm}
    \label{fig:rmnist_notrain}
    }
    \subfigure[]{
    \includegraphics[width=0.155\textwidth]{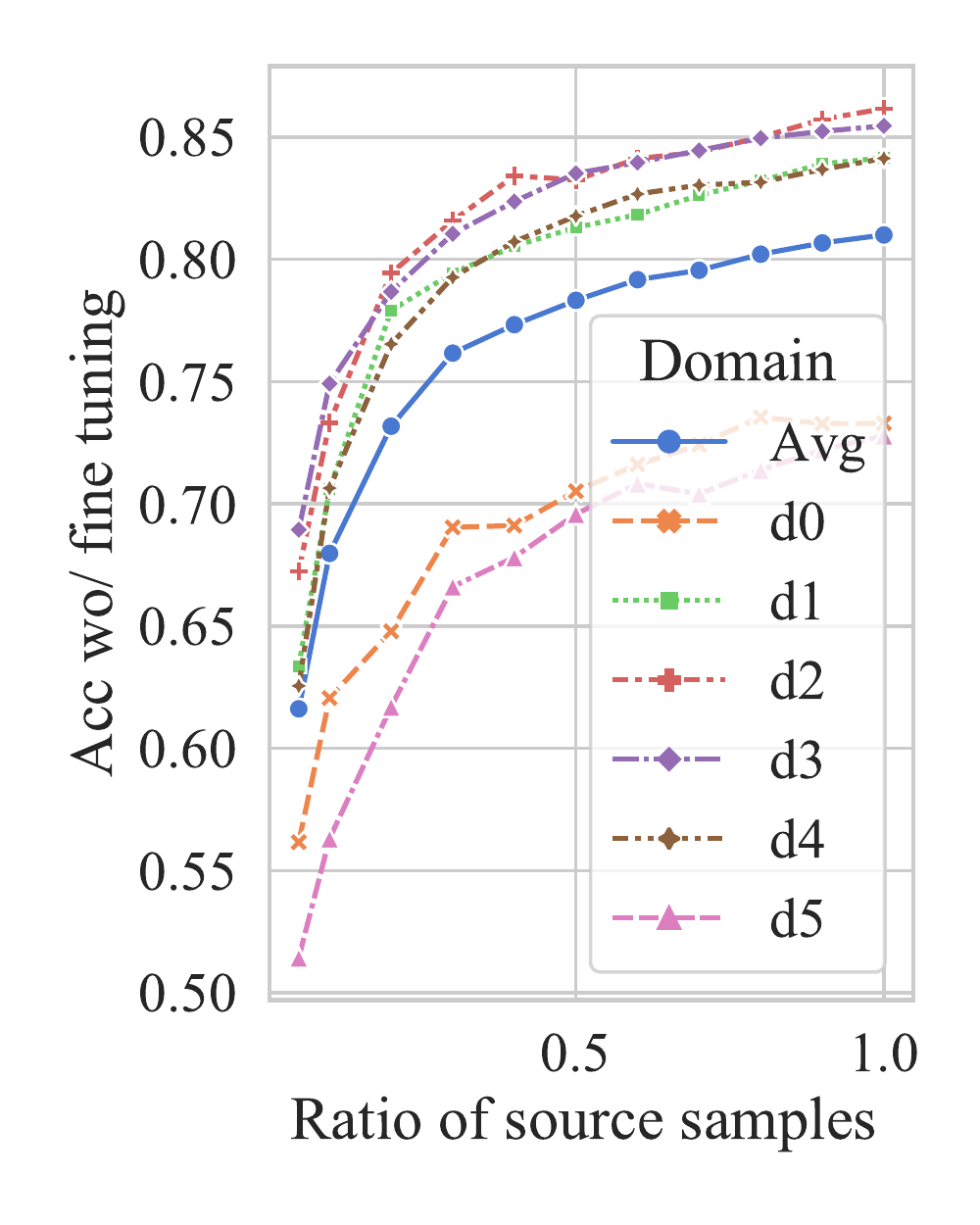}
    \vspace{-0.1cm}
    \label{fig:rmnist_ratio}
    }
    \subfigure[]{
    \includegraphics[width=0.255\textwidth]{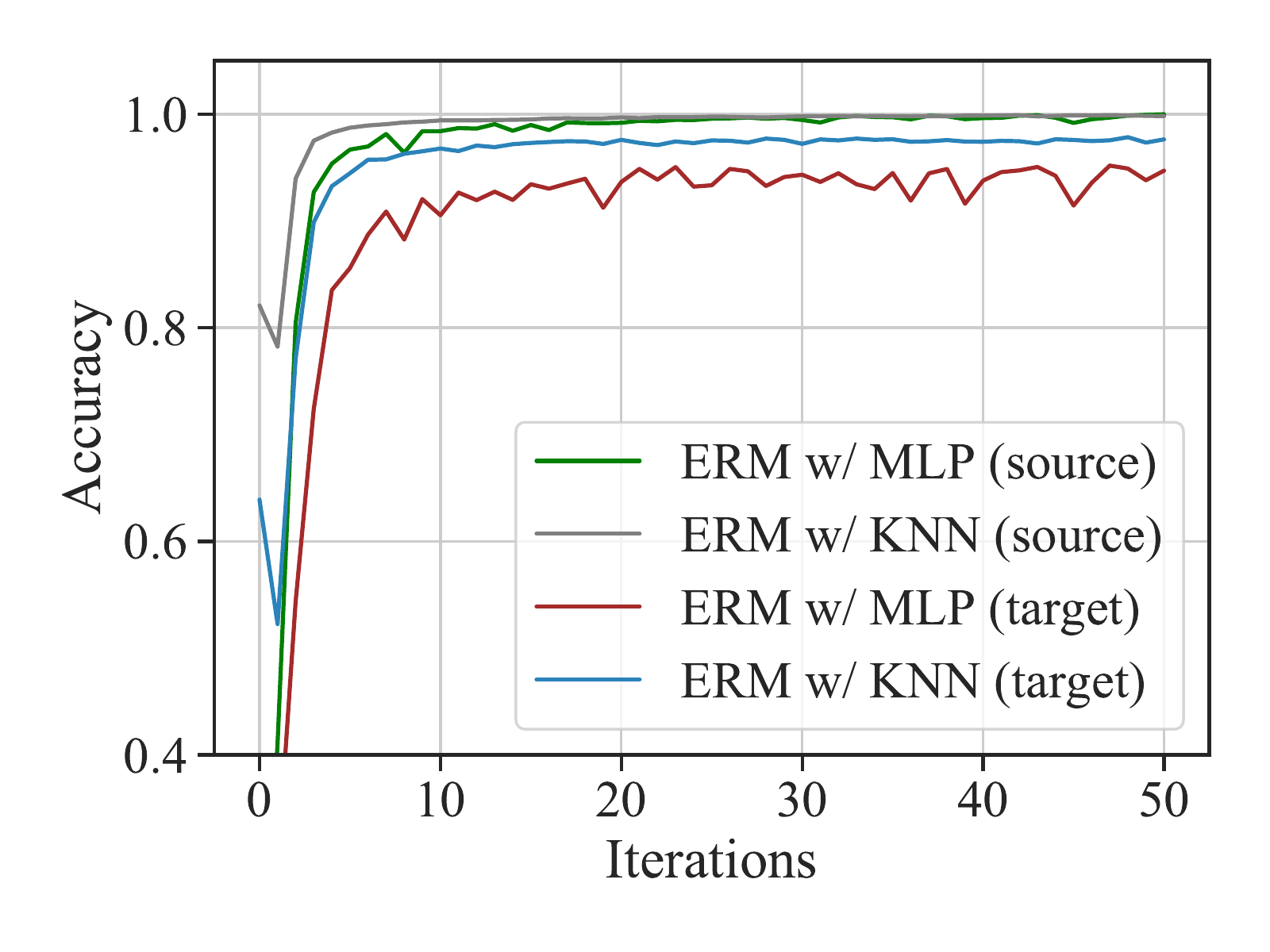}
   \vspace{-0.1cm}
     \label{fig:converge_exp}
    }
    \caption{\textbf{Generalization performance without fine-tuning} with various backbones on the (a) PACS and RotatedMNIST (b) dataset. (c) shows the results wo/ fine-tuning on Rotated MNIST with different ratios of source instances amounts. (d) Convergence comparison.} 
    \vspace{-0.2cm}
\end{figure*}
 \textbf{\knn improves different backbones by a large margin.} We conduct experiments with various backbones in Table~\ref{tab:backbons}, including ResNet50, ResNet18, and Vision Transformers (ViT-B16). \knn achieves consistent performance improvements compared to ERM. Specifically, \knn improves $2.5\%, 1.8\%$, $2.6\%$, and $1.5\%$ for ResNet-18, ResNet50, ResNet50-BN, and ViT-B16, respectively.

 \textbf{Without any model fine-tuning on source domains, \knn can perform well.} \figurename~\ref{fig:pacs_notrain} shows the results where the pre-trained model is directly evaluated on the target domain and without any fine-tuning on the source domains. The average generalization performance of using an MLP classifier is below $25\%$ even with a strong backbone (ViT-L16). On the contrary, the use of a KNN classifier achieves average generalization accuracy $71.4\%$. Nowadays, fine-tuning is usually computationally expensive due to the ever-growing size of pre-trained models. The requirement of \knn is not a gradient-based update but external high capacity storage to store knowledge for image classification, such as image feature maps, which provides a new promising direction to utilize pre-trained knowledge. Furthermore, as the number of source instances increases,~\figurename~\ref{fig:rmnist_ratio} shows that \knn achieves better performance, which validates our theoretical result.
%

\begin{figure}[t]
\vspace{-2mm}
    \centering
    \includegraphics[width=1.1\columnwidth]{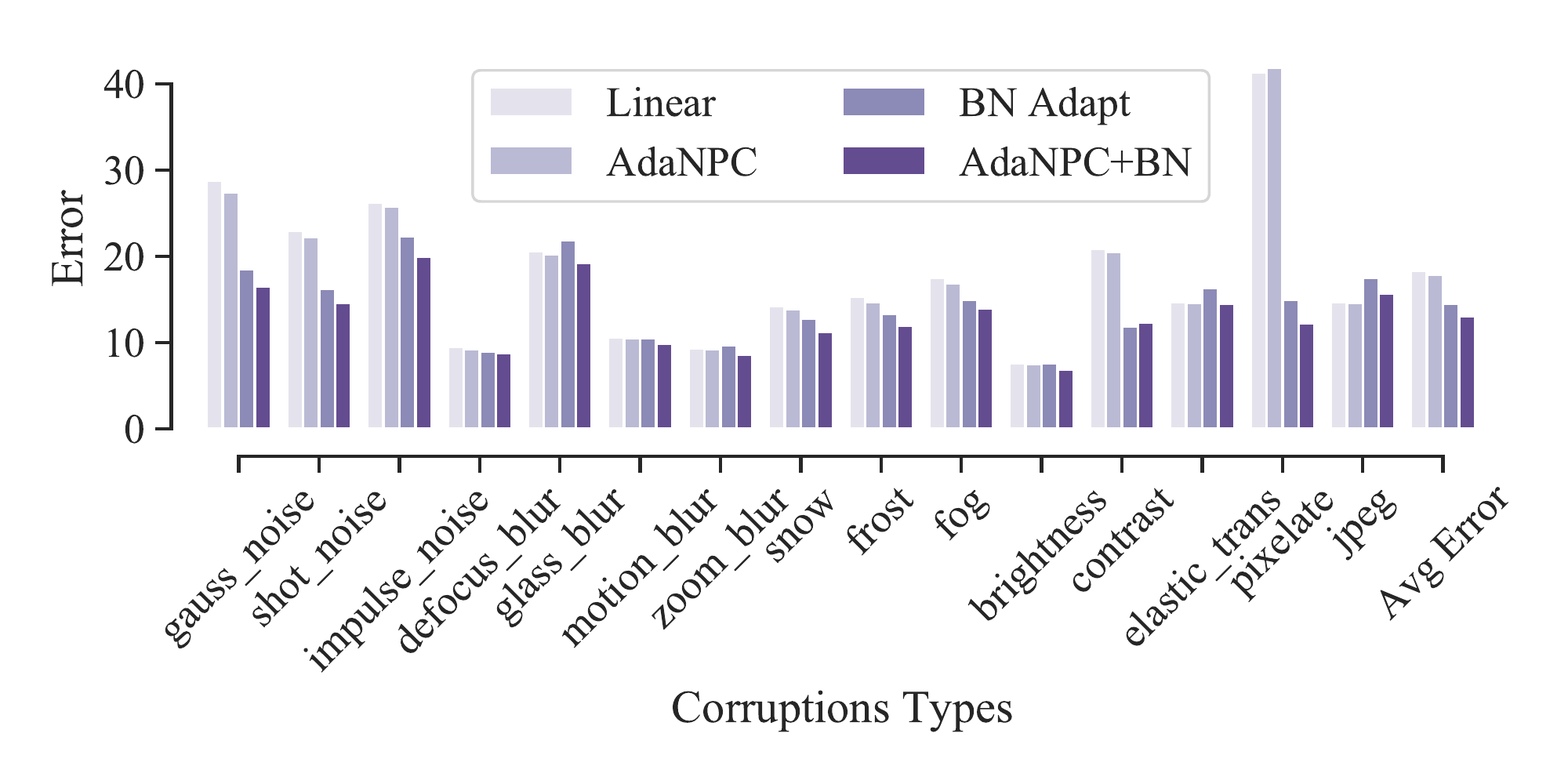}
    \vspace{-7mm}
    \caption{Corruption benchmark on CIFAR-10-C with severity 5.}
    \vspace{-2mm}
    \label{fig:cifar10-5}
\end{figure}

 \textbf{\knn reduces generalization error across various corruption types.} To verify robustness to corruptions, we evaluate \knn on the CIFAR-10-C benchmark~\cite{hendrycks2018benchmarking} with a 40-2 Wide ResNet backbone~\cite{zagoruyko2016wide} pre-trained on CIFAR-10. We implement two baselines, where \textbf{Linear} denotes that the trained classifier is used directly, and \textbf{BN Adapt} will update the batch normalization parameters of the backbone in the target data~\cite{schneider2020improving}. The results with the highest severity (five) are shown in~\figurename~\ref{fig:cifar10-5}, where \knn is shown to be more robust than Linear and updating the BN statistic of the backbone with \knn brings more benefits than that with a linear classifier. Results with the lowest severity (one), other backbones. \textbf{Considering large-scale corruption benchmarks}, we compared the method with Tent and EATA ~\cite{niu2022efficient} on the ImageNet-C dataset. Experimentally, we compared the effectiveness of AdaNPC and EATA separately in Table.~\ref{tab:corr}, where the experimental and baselines are all following EATA ~\cite{niu2022efficient}. As we can see, we only need to simply replace the Linear layer with KNN and remember the trustworthy samples during testing, and AdaNPC can achieve much better performance than ResNet50 with a linear head. In addition, AdaNPC can be combined with existing TTA methods such as Tent, ETA, etc., and the final performance surpasses these methods individually. See more analysis in the Appendix~\ref{sec:large}. 

\begin{table}[]
\centering
\begin{tabular}{lcccc}
\toprule
 & Defoc & Glass & Motion & Zoom \\ \midrule
R-50 (GN)+JT & 88.9 & 91.7 & 86.7 & 81.6 \\
R-50 (BN) & 82.1 & 90.2 & 85.2 & 77.5 \\\rowcolor{Gray}
AdaNPC & 83.1 & 83.0 & 72.3 & 60.6 \\ \midrule
TTT & 71.9 & 92.2 & 66.8 & 63.2 \\
TTA & 87.5 & 91.8 & 87.1 & 74.2 \\
BN adaptation & 80.0 & 80.0 & 71.5 & 60.0 \\
MEMO & 80.3 & 87.0 & 79.3 & 72.4 \\
Tent & 71.8 & 72.7 & 58.6 & 50.5 \\
Tent (episodic) & 85.5 & 85.4 & 74.6 & 62.2 \\\rowcolor{Gray}
AdaNPC+Tent & 71.1 & 72.0 & 58.2 & 49.2 \\ \midrule
ETA & 66.1 & 67.1 & 52.2 & 47.5 \\
EATA & 66.3 & 66.6 & 52.9 & 47.2 \\\rowcolor{Gray}
AdaNPC+ETA & 65.2 & 65.2 & 51.1 & 46.5 \\ \bottomrule
\end{tabular}%
\caption{\textbf{Comparison with state-of-the-art methods on ImageNet-C} with the highest severity level 5 regarding corruption Error.}\label{tab:corr}
\end{table}

\subsection{Comparison of \knn with other TTA methods.}\label{app:exp_tta}
\textbf{Comparison with test-time adaptation methods.}
For fair comparisons, following~\cite{iwasawa2021test}, the base models (ERM and \knn) are trained only on the default hyperparameters and without the fine-grained parametric search. Because~\cite{gulrajani2021in} omits the BN layer from pre-trained ResNet when fine-tuning on source domains, we cannot simply use BN-based methods on the ERM baseline.
For these methods, their baselines are additionally trained on ResNet-50 with BN.
Models with the highest IID accuracy are selected and all test-time adaptation methods are applied to improve the generalization performance.
The baselines include Tent~\cite{wang2020tent}, T3A~\cite{iwasawa2021test}, pseudo labeling (PL)~\cite{lee2013pseudo}, SHOT~\cite{liang2020we}, and SHOT-IM~\cite{liang2020we}.
For methods that use gradient backpropagation, we implement both update the prediction head (Clf) and full model (Full).
Results in Table~\ref{tab:adapt_comp} show that: Different from Tent~\cite{wang2020tent}, which is sensitive to batch size, the proposed \knn is not; (ii) The performance of \knn without BN retraining attains comparable results compared to existing methods. (iv)  Additionally, when the batch size is very small, updating the model parameters often has a negative impact, whereas the results of AdaNPC are not affected by the batch size.

\begin{table*}[]
\centering
\adjustbox{max width=0.8\linewidth}{%

\begin{tabular}{ccccccccc}
\toprule
\multirow{2}{*}{\textbf{Method}} & \multicolumn{4}{c}{\textbf{BSZ=32}} & \multicolumn{4}{c}{\textbf{BSZ=2}} \\\cmidrule(lr){2-5} \cmidrule(lr){6-9}
 & \textbf{RMNIST} & \textbf{PACS} & \textbf{VLCS} & \textbf{DomainNet} & \textbf{RMNIST} & \textbf{PACS} & \textbf{VLCS} & \textbf{DomainNet} \\
ResNet50 & 97.27 & 86.68 & 77.75 & 40.50 & 97.27 & 86.68 & 77.75 & 40.50 \\
PLClf & 98.13 & 87.73 & 80.75 & 40.80 & 69.78 & 87.73 & 80.50 & 40.67 \\
PLFull & 98.30 & 87.13 & 77.23 & 33.80 & 65.37 & 73.35 & 58.55 & 21.80 \\
SHOT & 98.30 & 88.16 & 67.83 & 41.21 & 92.12 & 83.55 & 58.45 & 36.76 \\
SHOTIM & 98.40 & 88.03 & 67.68 & 37.62 & 92.33 & 83.20 & 58.20 & 35.43 \\
T3A & 97.65 & 87.90 & 81.38 & 41.50 & 97.10 & 87.90 & 81.38 & 41.50 \\
Tent-clf & 96.82 & 87.05 & 77.78 & 40.97 & 96.18 & 87.03 & 77.50 & 40.16 \\\rowcolor{Gray}
\textbf{AdaNPC} & 98.85 & 88.93 & 82.45 & 42.60 & 98.85 & 88.93 & 82.45 & 42.60 \\ \bottomrule
\end{tabular}%
}
\caption{Comparison of our method and existing test-time adaptation methods on OOD benchmarks. The reported number is the average generalization performance over all domains.}\label{tab:adapt_comp}
\end{table*}

\vspace{-0.1cm}
\subsection{Analysis}
\vspace{-0.1cm}

 \textbf{\knn attains better performance with fewer iterations.} We investigate the accuracy dynamics of ERM, which is evaluated by using either a KNN classifier or a Multi-Layer Perception (MLP) classifier on the Rotated MNIST dataset, where the target domain is $d_0$. The learning curves in Fig.~\ref{fig:converge_exp} show that with the same training process and iteration, using a KNN classifier can attain superior performance on the source and target domains both.
\begin{table}[]
\centering
\adjustbox{max width=\columnwidth}{%
\centering
\begin{tabular}{@{}cccccccc@{}}
\toprule
\multicolumn{8}{c}{\textbf{PACS, ‘Test-domain’ validation}} \\ \midrule
\textbf{Training Loss} & $k$ & $|\mathcal{M}|$  & \textbf{A} & \textbf{C} & \textbf{P} & \textbf{S} & \textbf{Avg} \\ \midrule
\multirow{6}{*}{ERM} & Linear & / & 82.7 & 82.3 & 93.1 & 78.6 & 84.2 \\
 & 50 & / & 83.1 & 82.1 & 93.6 & 78.9 & 84.4 \\
 & 75 & / & 83.2 & 82.1 & 93.5 & 78.8 & 84.4 \\
 & 100 & / & 83.2 & 81.9 & 93.5 & 78.7 & 84.3 \\
 & 125 & / & 84.6 & 82.3 & 93.7 & 81.0 & 85.4 \\
 & 150 & / & 85.2 & 82.6 & 94.2 & 80.4 & 85.6 \\
 \midrule
\multirow{5}{*}{$\mathcal{L}_{KNN}$} & 10 & 500 & 84.0 & 82.6 & 93.5 & 80.8 & 85.2 \\
 & 50 & 500 & 85.2 & 81.7 & 92.6 & 80.8 & 85.1 \\
 & 50 & 1000 & \textbf{85.8} & 81.5 & 93.8 & \textbf{83.4} & \textbf{86.1} \\
 & 50 & 1500 & 82.9 & \textbf{83.0} & 91.4 & 80.5 & 84.5 \\
 & 100 & 1000 & 85.0 & 82.5 & \textbf{94.6} & 80.7 & 85.7 \\ \bottomrule
\end{tabular}
}
\vspace{-1mm}
\caption{Ablation studies of training loss on PACS.}
\label{tab:knn-training-pacs}
\end{table}
  

 \textbf{Ablation studies of training loss (\myref{equ:knn_training}) and the choice of $k$}. Results are shown in Table.~\ref{tab:knn-training}, where \textbf{ERM} means that the model will be trained by cross-entropy loss and only use KNN with parameter $k$ for inference and adaptation. The results show that with $\mathcal{L}_{KNN}$, the representation will be better and the generalization results will be improved. The choice of $|\mathcal{M}|$ depends on the dataset size, a large dataset with more classes generally needs a large memory bank. However, if $|\mathcal{M}|\geq batch\;size$, $\mathcal{M}$ contains many out-of-date features and leads to poor performance. The choice of $k$ depends both on the domain divergence and instances numbers of datasets. For example, the best $k$ is around $10$ for RMNIST and $50$ for PACS (PACS has a greater domain divergence), where PACS has $9,991$ images, which is less than $60,000$ images of RMNIST.

\begin{figure}[t]
    \centering
\includegraphics[width=0.45\textwidth]{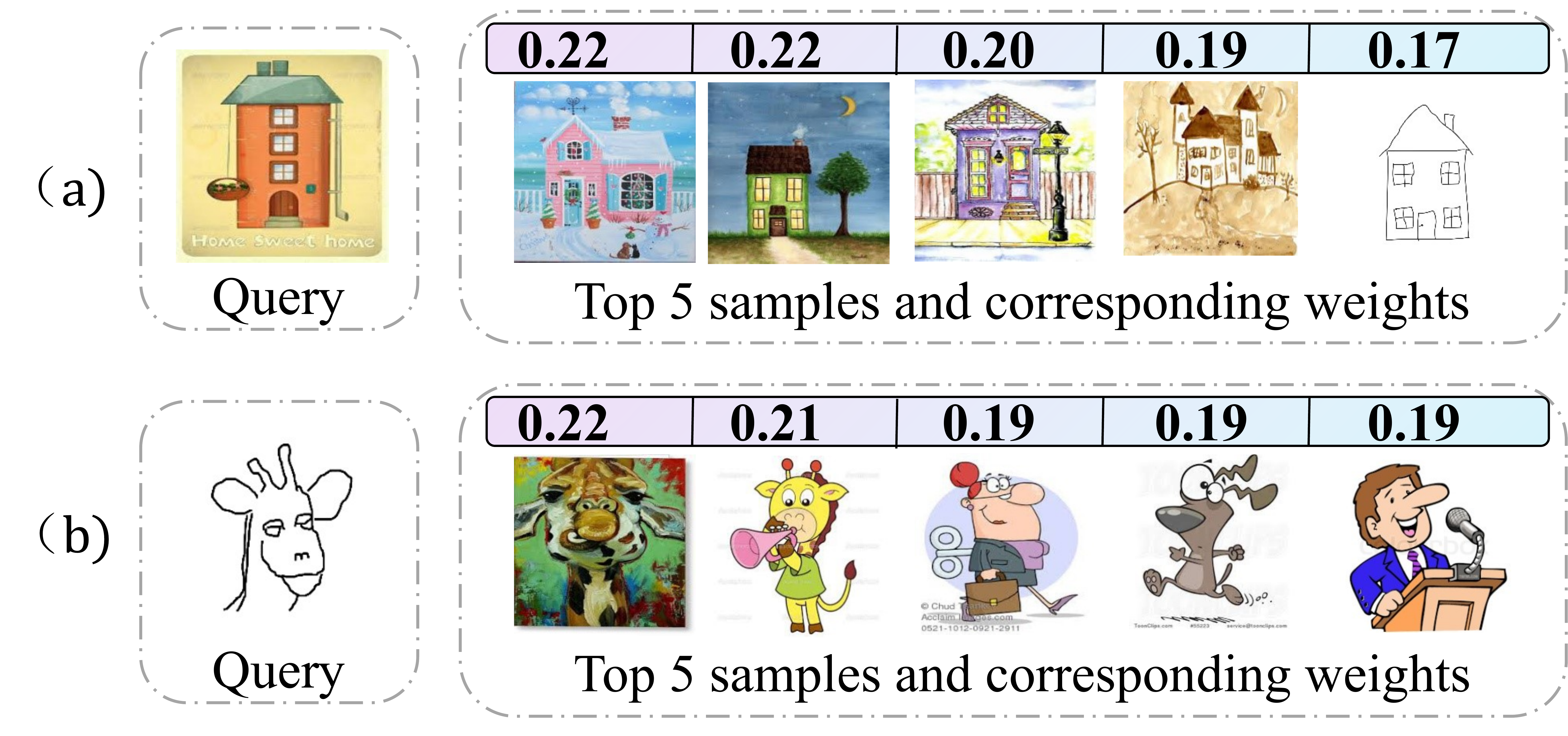}
    \caption{\textbf{Visualization of classified results attained by \knn.} (a) A successfully classified instance and (b) a failure case.}
    \label{fig:main_class}
    \vspace{-0.2cm}
\end{figure}

\textbf{Interpretability and human-model interaction.} \figurename~\ref{fig:main_class} shows how the source knowledge is used by the KNN classifier. The decision-making process will no longer be a black box. For example, the giraffe in \figurename~\ref{fig:main_class}(b) is classified with low confidence because its nearest neighbors are most persons or dogs that have similar poses. However, some important characteristics are ignored by the backbone representations, such as the shape of the face. However, these characteristics can be easily identified by humans; therefore, when we get a low confidence prediction, \knn allows us to manually remove some obvious wrong neighbors. In this case, our classification results will be more accurate and confident, which is promising for high-risk tasks to incorporate expert knowledge for better classification results.

\textbf{Inference time and Memory usage.}\label{app:sec_inf}
During testing, we store all samples from the source domain and do not delete any samples from the memory bank, thus preserving all information from the source domain. This may raise concerns about memory usage and inference time, but our experiments show that the K-nearest neighbor search is very fast. Furthermore, we provide a comparison of inference times for various TTA algorithms on different datasets in Table~\ref{tab:infernece_time}, measured as the average inference time per image in milliseconds.  For memory usage. Even for a dataset like ImageNet with 1281167 images, the additional memory cost is only 2.44GB (1281167*2048 B), which can be easily accommodated by current CPUs or memory. It is important to note that the memory bank does not have to be stored on the GPU, and all reported inference times in this paper are based on a CPU-stored memory bank.

\begin{table}[]
\centering
\adjustbox{max width=\columnwidth}{%
\begin{tabular}{lccccc}
\toprule
\multicolumn{6}{c}{\textbf{Inference time (ms)}} \\  \hline
\textbf{Method} & \multicolumn{1}{l}{\textbf{RMNIST}} & \multicolumn{1}{l}{\textbf{PACS}} & \multicolumn{1}{l}{\textbf{VLCS}} & \multicolumn{1}{l}{\textbf{DomainNet}} & \multicolumn{1}{l}{\textbf{ImageNet}} \\ \hline
PLClf & 0.49 & 5.38 & 5.57 & 6.17 & 5.55 \\
PLFull & 123.44 & 281.90 & 266.00 & 267.17 & 270.68 \\
SHOT & 0.66 & 7.34 & 7.12 & 6.96 & 7.45 \\
SHOTIM & 1.58 & 10.14 & 10.16 & 7.86 & 10.23 \\
T3A & 0.21 & 1.67 & 1.51 & 8.11 & 18.46 \\
Tent-clf & 11.24 & 41.63 & 27.60 & 92.37 & 32.34 \\  \rowcolor{Gray}
AdaNPC & 0.13 & 1.57 & 1.57 & 3.16 & 22.23 \\ \bottomrule
\end{tabular}%
}
\caption{\textbf{Inference time of different TTA methods}, where the \knn is implemented by Faiss~\cite{johnson2019billion}.}\label{tab:infernece_time}
\end{table}

\vspace{-0.2cm}
\section{Concluding Remarks}
\label{sec:conclusion}

The paper proposes a new Test-Time adaption method for domain generalization, \knn, which revisits a non-parametric classifier, namely KNN classifier, for prediction and adaptation. Unlike current domain generalization or Test-Time adaptive methods that need model updating and are easy to forget previous knowledge, the proposed method is parameter-free and can memorize all the knowledge, making \knn suitable for practical settings, especially for adapting to a series of domains.

We derive error bounds under both the covariate-shift and the posterior-shift settings, where \knn is theoretically shown able to reduce unseen target error. We empirically show that \knn reduces generalization error on both unseen target domains and corrupted data. Besides, \knn attains faster convergence, better interpretability, and strong knowledge expandability. More importantly, \knn achieves high generalization accuracy without any fine-tuning on source domains, which provides a promising direction for utilizing pretrained models with growing sizes.

\textbf{Discussion and Limitations.} One potential limitation of \knn will be the \textbf{computation time of dense vectors’ searching} (finding the $k$-nearest neighbors). However, with efficient nearest neighbor search technique~\cite{johnson2019billion}, even when the memory bank contains more than $1$ million samples, the inference time will around $10$ms~\cite{sun2022out}, which is significantly
faster than backward gradient updates.
In the future, we will consider updating the memory bank with advanced methods and try to further reduce the memory cost and inference time.




\bibliography{icml2023}
\bibliographystyle{icml2023}

\newpage
\onecolumn
\appendix
\begin{center}
{\LARGE \textbf{AdaNPC: Exploring Non-Parametric Classifier for \\ Test-Time Adaptation\\ $\;$ \\ ————Appendix————}}
\end{center}



The structure of Appendix is as follows

\begin{itemize}
    \item Appendix~\ref{sec:app_related} contains the extended related work.
    \item Appendix~\ref{app:proof} contains all missing proofs in the main manuscript.
    \item Appendix~\ref{app:alg} details the optimization algorithm of the proposed \knn.
    \item Appendix~\ref{sec:data_detail} details the dataset and implementation Details.
    \item Appendix~\ref{sec:addexp} contains additional experimental results.
\end{itemize}

\begin{table}[b]
\centering
\adjustbox{max width=\columnwidth}{%
\begin{tabular}{@{}cccccc@{}}
\toprule
                              \multicolumn{6}{c}{Test-Time Adaptative Methods}             \\ \midrule
                             & Target Batch & Source Training &Fine-tune & Extra Model & Adaptive \\
Native    DG~\cite{arjovsky2020invariant}                   &   $\times$    & $\times$    &     $\times$      &    $\times$          &  $\times$         \\
Test-Time Training~\cite{sun2020test,zhang2021test,zhang2021memo}  & $\times$         &   $\times$   & $\checkmark$          &   $\checkmark$          &    $\checkmark$      \\
Test-Time Adaptation~\cite{wang2020tent}    & $\checkmark$     &   $\times$   &  $\checkmark$         &       $\times$       &   $\checkmark$       \\
Domain-adaptative method~\cite{dubey2021adaptive,zhang2022domain}     & $\checkmark$     &   $\times$   &    $\times$        &     $\checkmark$        &    $\checkmark$      \\
Single sample generalization~\cite{xiao2022learning}& $\times$ &   $\checkmark$   &    $\times$        &      $\times$        &   $\checkmark$       \\
\textbf{Non-Parametric Adaptation}     & $\times$    &     $\times$  &      $\times$      &        $\times$      &     $\checkmark$        \\ \bottomrule
\end{tabular}
}
\caption{\textbf{Test-Time adaptive methods}. The target batch means that the methods need batches of target samples for adaptation. Compared to existing test-time adaptative methods, the proposed \knn imposes no additional parameter~\cite{sun2020test,dubey2021adaptive,zhang2022domain}, no extra tunning steps~\cite{sun2020test,wang2020tent,iwasawa2021test,zhang2021test}, and does not need to use source data to learn adaptive strategies~\cite{xiao2022learning}.}
\label{tab:test_methods}
\vspace{-0.3cm}
\end{table}

\section{Extended Related work}\label{sec:app_related}
\noindent\textbf{Transfer learning theory.} The first line of work that considers bounding the error on target domains by the source domain classification error and a divergence measure, such as $d_\mathcal{A}$ divergence~\cite{ben2006analysis,david2010impossibility} and $\mathcal{Y}$ divergence~\cite{mansour2009domain,mohri2012new}. However, the symmetric differences carry the wrong intuition and most of these bounds depend on the choice of hypothesis~\cite{kpotufe2018marginal,zhang2022domain}. There are also some studies consider the density ratio between the source and target domain~\cite{quinonero2008dataset,sugiyama2012density,zhang2022domain}, and transfer-exponent for non-parametric transfer learning~\cite{kpotufe2018marginal,cai2021transfer,hanneke2019value,reeve2021adaptive}. In this work, we conduct error bounds considering a setting that consists of two different components, namely, the non-parametric classifier and online arrival target samples.

\textbf{Domain generalization.} Previous DG methods mostly focus on representation learning, namely learning domain-invariant representations or only task-relevant representations. However, empirical risk minimization (ERM) has been shown to be able to beat most existing domain generalization methods in average performance~\cite{gulrajani2021in}. Recent work finds that ERM has learned a high-quality representation on datasets with spurious correlations, even when the model relies primarily on spurious features to make predictions~\cite{kirichenko2022last}. The current bottleneck to out-of-distribution generalization primarily lies in learning simple and reliable classifiers~\cite{rosenfeld2022domain}. However, most existing methods have an over-confidence hypothesis space~\cite{zhang2022domain}, namely, they assume that the hypothesis performs well on source domains can also perform well on the target domain~\cite{arjovsky2020invariant,krueger2021out,rame2022fishr,zhang2023free,zhang2022learning}, which cannot hold on any unseen target domains.  Our method is orthogonal to most existing DG methods since it replaces the linear classifier with a KNN classifier. \knn retains all information seen in training and has a complex hypothesis space controlled by the parameter $k$. During inference, \textit{the hypothesis space complicity can be flexibly controlled compared to existing methods that use a frozen classifier}. 

\section{Proof of Theoretical Statement}\label{app:proof}

\subsection{Non-parametric reduce target-source domain divergence (Proof of Proposition 1)}\label{sec:theo}
To complete the proofs, we begin by introducing some necessary definitions and assumptions. 

\begin{definition}
(\textbf{Wasserstein-distance and the dual form} ~\cite{arjovsky2017wasserstein}). The $\rho$-th Wasserstein distance between two distributions $\mathbbm{D}_S, \mathbbm{D}_U$ is defined as
\begin{equation}
\mathcal{W}_\rho(\mathbbm{D}_S,\mathbbm{D}_U)=\left(\inf_{\gamma\in \Pi[\mathbbm{D}_S,\mathbbm{D}_U]} \iint d({x_s},{x_u})^\rho  d\gamma({x_s},{x_u})\right)^{1/\rho}
\label{eq:wd}
\end{equation}
where $\Pi[\mathbbm{D}_S,\mathbbm{D}_U]$ is the set of all joint distribution on $\mathcal{X}\times\mathcal{X}$ with marginals $\mathbbm{D}_S$ and $\mathbbm{D}_U$ and $d(x_s,x_u)$ is a distance function for two instances $x_s,x_u$.

Wasserstein distance can get intuition from the optimal transport problem, where $d(x_s,x_u)^\rho$ is the unit cost for transporting a unit of material from $x_s\in \mathbbm{D}_S$ to $x_u\in \mathbbm{D}_U$ and $\gamma(x_s,x_u)$ is the transport policy which satisfies the marginal constraint. According to the Kantorovich-Rubinstein theorem, the dual representation of the first Wasserstein distance (Earth-Mover distance) can be written as
\begin{equation}
\mathcal{W}_1(\mathbbm{D}_S,\mathbbm{D}_U)=\sup_{\parallel f\parallel_L\leq 1} \mathbb{E}_{x_s\in \mathbbm{D}_S}[f(x_s)]-\mathbb{E}_{x_u\in \mathbbm{D}_U}[f(x_u)],
\end{equation}
where $\parallel f\parallel_L=\sup|f(x_s)-f(x_u)|/d(x_s,x_u)$ is the Lipschitz semi-norm. 
\label{define1}
\end{definition}

We first use the domain adaptation result, Theorem 1 in~\cite{shen2018wasserstein} that considers the Wasserstein distance. On this basis, we can clearly show the effect of \knn on the domain divergence. In this paper, we use $\mathcal{W}_1(\mathbbm{D}_S,\mathbbm{D}_U)$ as default and ignore subscript 1. For completeness, we present the Theorem 1 in~\cite{shen2018wasserstein} as follow:

\begin{prop}
(Theorem 1 in~\cite{shen2018wasserstein}) Given two domain distributions $\mathbbm{D}_S,\mathbbm{D}_U$, denote $f^*=\arg\min_{f\in\mathcal{H}}(\epsilon_U(f)+\epsilon_S(f))$ and $\kappa=\epsilon_U(f^*)+\epsilon_S(f^*)$. Assume all hypotheses $h$ are $L$-Lipschitz continuous, the risk of hypothesis $\hat{f}$ on the unseen target domain is then bounded by
\begin{equation}
    \epsilon_U(\hat{f})\leq \kappa+\epsilon_S(\hat{f})+2L\mathcal{W}(\mathbbm{D}_S,\mathbbm{D}_U).
    \label{lemma:bound}
\end{equation}
\label{prop:wass}
\end{prop}

Intuitively, by using the non-parametric classifier, during inference, a large number of samples in source domains that are not similar to the target samples are ignored, and thus the domain divergence will be reduced. That is, the source distribution $\mathbbm{D}_S$ is replaced by $\Omega:=\bigcup_{x\in \mathbbm{D}_U} \mathcal{B}(x,r)$, where $\mathcal{B}(x,r)=\{x':\parallel x'-x\parallel\leq r\}$ denotes a ball centered on $x$ with radius $r$, and With a small $r$, $\Omega$ is intuitively close to $\mathbbm{D}_U$ because these dissimilar data points are ignored and the selected source data are all close to the target data. Informally, according to \myref{eq:wd}, we have $\mathcal{W}(\Omega,\mathbbm{D}_U)=\inf_{\gamma\in \Pi[\Omega,\mathbbm{D}_U]} \iint \parallel {x_s}-{x_u}\parallel  d\gamma({x_s},{x_u})$, where for each $x_s\in\Omega$, we can find at least one $x_u\in \mathbbm{D}_U$ such that $\parallel x_s-x_u\parallel\leq r$, the overall distance will then be bounded by $r$. If $r$ is small enough, $\mathcal{W}(\mathbbm{D}_S,\mathbbm{D}_U)$ in Proposition~\ref{prop:wass} is largely reduced. Specifically, we can choose a density function $\gamma^*$ where $\gamma^*(x_s,x_u)>0$ only if $x_s\in \mathcal{B}(x_u,r)$ otherwise 0, then we have

\begin{equation}
\mathcal{W}(\Omega,\mathbbm{D}_U)=\inf_{\gamma\in \Pi[\Omega,\mathbbm{D}_U]} \iint \parallel {x_s}-{x_u}\parallel  d\gamma({x_s},{x_u})\leq  \iint \parallel {x_s}-{x_u}\parallel \gamma^*({x_s},{x_u})  d x_sx_u\leq r
\end{equation}

Although a small $r$ will reduce the generalization bound, there is no guarantee that each data $x_u\in \mathbbm{D}_U$ can find a neighbor $\mathcal{B}(x,r)$ with $|\mathcal{B}(x,r)|>0$. To this end, we theoretically discuss the choice of $r$ and show \textit{given a choice radius $r$, what probability that the set of neighbors $\mathcal{B}(x,r)$ of each $x\in \mathbbm{D}_U$ is not measuring zero}?

We denote $k$ is the number of neighbors that we prefer to choose, namely the parameter for the KNN classifier, $n_s$ is the total number of data in ${D}_S$. With the strong density assumption, for any $x_u\in \mathbbm{D}_U, r<r_\mu$, according to Assumption~\ref{assump1}, we have

\begin{equation}
\mathbbm{D}_S(x_s\in \mathcal{B}(x_u,r))=\int_{\mathcal{B}(x_u,r)\cap \mathbbm{D}_S} \frac{d \mathbbm{D}_S}{d\lambda}(x_s)dx_s\geq \mu_-\lambda(\mathcal{B}(x_u,r)\cap \mathbbm{D}_S)\geq c_\mu\mu_-\pi_dr^d,
\label{equ:a}
\end{equation}

where $\pi_d=\lambda(\mathcal{B}(0,1))$ is the volume of the $d$ dimension unit ball and $\lambda$ is the Lebesgue measure of a set in a Euclidean space. Set $r_0=(\frac{2k}{c_\mu\mu_-\pi_dn_s})^{1/d}$, with a additional assumption that $\frac{k}{n_s}<\frac{c_\mu\mu_-\pi_dr_\mu^d}{2}$\footnote{The assumption is rational because, $n_s\gg k$ in general.}, we have $r_0<r_\mu$. Then for any $x_u\in \mathbbm{D}_U$, according to \myref{equ:a}, we have
\begin{equation}
    \mathbbm{D}_S(x_s\in \mathcal{B}(x_u,r_0)) \geq  c_\mu\mu_-\pi_dr_0^d >\frac{2k}{n_s}
\end{equation}
Denote $\mathbb{I}$ an indicator function and then $\mathbb{I}(x_s\in \mathcal{B}(x_u,r_0))$  are independent and identically Bernuoli variables, which mean is $\mathbbm{D}_S(x_s\in \mathcal{B}(x_u,r_0))$. Let $S_n(x_u)=\sum_{i=1}^{n_s}\mathbb{I}(x_s\in \mathcal{B}(x_u,r_0))$ denote the number of data $x_s\in \mathbbm{D}_S$ that fall into $\mathcal{B}(x,r_0)$, then $S_n(x_u)$ follows the Binomial distribution. Let $W\sim Binomial(n_s,\frac{2k}{n_s})$, according to the Chernoff inequality~\cite{chernoff1981note,chung2006concentration},

\begin{equation}
    P(S_n(x_u)<k)\leq P(W<k)= P(W-\mathbb{E}[W]<-k)\leq \exp(-k^2/2\mathbb{E}[W])=\exp(-k/4),
\end{equation}

where the second inequality is because $S_n(x)$ has a larger mean than $W$. We can see the probability that $S_n(x)<k$ is small for any $x_u\in \mathbbm{D}_U$, especially when $k$ is large. Denoting $x_s^{(i)}$ the $i-$th nearest data to $x_u$ among $\mathcal{B}(x_u,r_0)$, we have for any $x_u\in \mathbbm{D}_U$

\begin{equation}
P(\parallel x_s^{(k)}-x_u \parallel\leq r_0)= P(S_n(x)\geq k)\geq 1-\exp(-k/4)
\label{bound:x_diff}
\end{equation}
Combine \myref{bound:x_diff} with the assumption that the distribution $\mathbbm{D}_U$ is finite with cardinality $n_{\mathbbm{D}_U}$ and the desired probability part is shown by union bound.  

\begin{align}
\bigcap_{x_u\in \mathbbm{D}_U} P(\parallel x_s^{(k)}-x_u \parallel\leq r_0)&=\bigcap_{x_u\in \mathbbm{D}_U} P(S_n(x)\geq k)\notag\\
&=1-\bigcup_{x_u\in \mathbbm{D}_U}P(S_n(x)< k)\notag\\
&\geq 1-n_{\mathbbm{D}_U}\exp\left(-\frac{k}{4}\right)\notag\\
&= 1-\exp\left(-\frac{k}{4}+\log n_{\mathbbm{D}_U}\right).
\end{align}

Finally, the following proposition is derived.

\begin{prop}
Given two domain distributions $\mathbbm{D}_S,\mathbbm{D}_U$, and $\Omega:=\bigcup_{x\in \mathbbm{D}_U} \mathcal{B}(x,r)$, where $\mathcal{B}(x,r)=\{x':\parallel x'-x\parallel\leq r\}$ denotes a ball centered on $x$ with radius $r$. Denote $f^*=\arg\min_{f\in\mathcal{H}}(\epsilon_U(f)+\epsilon_\Omega(f))$ and $\kappa=\epsilon_U(f^*)+\epsilon_\Omega(f^*)$. Assume all hypotheses $h$ are $L$-Lipschitz continuous, the risk of hypothesis $\hat{f}$ on the unseen target domain is then bounded by
\begin{equation}
\begin{aligned}
\epsilon_U(\hat{f})\leq \kappa+\epsilon_\Omega(\hat{f})+2L\left(\frac{2k}{c_\mu\mu_-\pi_dn_s}\right)^{1/d}.   
\end{aligned}
\end{equation}
with probability $1-\exp(-\frac{k}{4}+\log n_{\mathbbm{D}_U})$
\label{prop:wass_omega}
\end{prop}

\textbf{Remarks.} We make the following conclusions (i) with a larger number of source data, the error will be lower; (ii) a large $c_\mu$ will reduce the error bound, which is intuitive because $x_u\in \mathbbm{D}_U$ will not be so far from $\mathbbm{D}_S$ when $c_\mu$ is large and the adaptation will be easier; (iii) a smaller parameter $k$ will reduce the domain divergence $\left(\frac{2k}{c_\mu\mu_-\pi_dn_s}\right)^{1/d}$. For example, when $k=1$, we only choose the closest source data with respect to $x_u$ and the divergence will be the minimum.

Although Proposition~\ref{prop:wass_omega} provides a nice intuition for using a non-parametric classifier, it highly depends on the risk of the optimal hypothesis $\kappa$, that is, the hypothesis space should contain an optimal classifier that performs well on both the source and the target domains. This assumption cannot be guaranteed to hold true under all scenarios, making the bound conservative and loose. Furthermore, the KNN classifier in Proposition~\ref{prop:wass_omega} can only affect domain divergence, and how KNN affects the prediction results is unknown. To further aid the study of the proposed algorithm, we conduct the following bounds to fully explore the theoretical properties.

\subsection{Analysis of the excess error upper-bound under covariate-shift (Proof of Proposition 2)}
Per the statement of the Proposition 2, we assume $k$ being of order $\log n_s$. It is quite small number. For example, in the RotatedMNIST dataset, the optimal $k$ is around $10$ and the number of total instances from the source domain $n_s\approx50,000$.

\textbf{Under the covariate-shift setting}, we have $\eta_U=\eta_S=\eta$ for source and target domains. We denote the KNN classifier with $k$ nearest neighbors as $\hat{f}_k=\mathbb{I}\{\hat{\eta}_k\geq \frac{1}{2}\}$. Because we focus on the binary classification setting, then $\hat{f}_k(x_u)\neq f_U^*(x_u)$ implies that $\left| \hat{\eta}_k(x_u) -\eta(x_u)\right|\geq \left|\eta(x_u)-\frac{1}{2}\right|$. In this way, we can build the connection between the excess error and the regress error:
\begin{equation}
    \mathcal{E}_U(\hat{f})=2\mathbb{E}_{x_u\sim \mathbbm{D}_U}\left[\left|\eta(x_u)-\frac{1}{2}\right|\mathbb{I}\left\{\left| \hat{\eta}_k(x_u) -\eta(x_u)\right|\geq \left|\eta(x_u)-\frac{1}{2}\right|\right\}\right]
\end{equation}

Let $Z=\left|\eta(x_u)-\frac{1}{2}\right|$, if we can bound $\sup_{x_u} \left| \hat{\eta}_k(x_u) -\eta(x_u) \right|\leq t$, then by the marginal assumption in Assumption~\ref{define_noise} and the fact that 
\begin{equation}
    \mathbb{E}\left[Z\cdot \mathbb{I}\{Z\leq t\}\right]\leq tP(Z\leq t),
\label{equ:indcator}
\end{equation}
we have $ \mathcal{E}_U(\hat{f})\leq C_\beta t^{\beta+1}$. To bound $\left| \hat{\eta}_k(x_u) -\eta(x_u) \right|$, we denote $(x_s^{(i)},y_s^{(i)})$ as the $i-$th nearest data and the corresponding labels to $x_u$ in $\mathcal{B}(x_u,r_0)$. The KNN classification result will be $\hat{\eta}(x_u)=\sum_{i=1}^k w_i y_s^{(i)}$, where $w_i$ is the weight for the $i$ -th nearest neighbor and $\sum_{i=1}^k w_i=1$. In this work, we use the cosine similarity as the weight, where the distance-weighted KNN is shown able to reduce the misclassification error~\cite{dudani1976distance}. However, for brevity of the proof, we assume $w_i=\frac{1}{k},\forall i\in[1,...,k]$, namely all nearest data labels are uniformly mixed. Based on the assumptions and notions above, we have for any $x_u\in \mathbbm{D}_U$

\begin{equation}
\begin{aligned}
\left|\hat{\eta}_k(x_u) -\eta(x_u) \right |&=\left| \frac{1}{k}\sum_{i=1}^k y_s^{(i)}-\eta(x_u)  \right|\\
& \leq \left|\frac{1}{k}\sum_{i=1}^k y_s^{(i)}-\frac{1}{k}\sum_{i=1}^k\eta\left(x_s^{(i)}\right)\right|+\left| \frac{1}{k}\sum_{i=1}^k\eta\left(x_s^{(i)}\right)-\eta(x_u)  \right|\\
&\leq \underbrace{\frac{1}{k}\left| \sum_{i=1}^k y_s^{(i)}-\sum_{i=1}^k \eta\left(x_s^{(i)}\right)\right|}_{\rm \circled{\rm 1}}+\underbrace{\frac{1}{k}\sum_{i=1}^k\left| \eta\left(x_s^{(i)}\right)-\eta(x_u)  \right|}_{\rm \circled{\rm 2}},
\end{aligned}
\label{equ:bound_cova}
\end{equation}
where ${\rm \circled{\rm 2}}$ is easy to bound. According to the assumption that $\eta_U$ is $C$-Smoothness, we have
\begin{equation}
\sum_{i=1}^k\frac{1}{k}\left| \eta\left(x_s^{(i)}\right)-\eta(x_u)  \right|\leq  \sum_{i=1}^k\frac{1}{k} C \cdot \parallel x_s^{(i)}- x_u \parallel \leq C \cdot \parallel x_s^{(k)}- x_u \parallel
\label{equ:19}
\end{equation}
According to \myref{bound:x_diff}, with probability at least $1-\exp(-k/4)$, ${\rm \circled{\rm 2}}\leq C\left(\frac{2k}{c_\mu\mu_-\pi_dn_s}\right)^{1/d}$. Note that $E_{Y|X}[y_s^{(i)}]=\eta(x_s^{(i)})$, then we use the Hoeffding inequality to obtain the upper bound of ${\rm \circled{\rm 1}}$
\begin{equation}
P_{X,Y}\left(\frac{1}{k}\left|\sum_{i=1}^k y_s^{(i)}-\sum_{i=1}^k\eta\left(x_s^{(i)}\right)\right|>\epsilon\right)=\mathbb{E}_X\left[P_{Y|X}\left(\frac{1}{k}\left|\sum_{i=1}^k y_s^{(i)}-\sum_{i=1}^k\eta\left(x_s^{(i)}\right)\right|>\epsilon\right)\right]\leq 2\exp(-2k\epsilon^2)
\label{equ:21}
\end{equation}
Set $\epsilon=(1/k)^{1/4}$, we have, with probability, at least $1-2\exp(-2\sqrt{k})$, ${\rm \circled{\rm 1}}\leq(1/k)^{1/4}$, ${\rm \circled{\rm 2}}\leq C\left(\frac{2k}{c_\mu\mu_-\pi_dn_s}\right)^{1/d}$, and then $\left|\hat{\eta}_k(x_u) -\eta(x_u) \right |\leq (1/k)^{1/4}+C\left(\frac{2k}{c_\mu\mu_-\pi_dn_s}\right)^{1/d}$. According to \myref{bound:x_diff} and \myref{equ:indcator}, the excess error is bounded by

\begin{equation}
\mathcal{E}_U(\hat{f})\leq 2C_\beta\left( \left(\frac{1}{k}\right)^{1/4}+  C\left(\frac{2k}{c_\mu\mu_-\pi_dn_s}\right)^{1/{d}}\right)^{{1+\beta}}\approx \left(\left(\frac{1}{k}\right)^{1/4}+C_1  \left(\frac{k}{c_\mu n_s}\right)^{1/{d}} \right)^{{1+\beta}},
\label{bound:22}
\end{equation}
where $C_1$ is a newly introduced constant. There is a clear tradeoff between the upper bound of ${\rm \circled{\rm 1}}$ and ${\rm \circled{\rm 2}}$ with respect to the value of $k$. A small $k$ will reduce the representation difference in ${\rm \circled{\rm 2}}$, extremely when $k=1$, only the nearest sample to $x_u$ will be chosen. However, when $k$ is small, there is no guarantee that the nearest selected data will have a confident prediction. Specifically, a smaller ${\rm \circled{\rm 1}}$ indicates that the selected $k$ nearest data samples are representative enough and have confidence in the prediction results. Finally, using $k = \mathcal{O}(\log n_s)$,  we have 
\begin{equation}
\begin{aligned}
&\min\{1-2\exp(-2\sqrt{k}),1-\exp(-k/4)\}\\ 
&\geq 1-2\exp(-2\sqrt{k})-\exp(-k/4)\\
&\geq 1-3\exp(-2\sqrt{k})=1-3\exp(-2\sqrt{\mathcal{O}(\log n_s)})\\
&=1-3\exp(-\mathcal{O}(1)\sqrt{\log n_s})
\end{aligned}
\end{equation}
where the third line is because $k/4>2\sqrt{k}$ for large enough $k$. Namely, with probability at least $ 1-3\exp(-\sqrt{\log n_s})^{\mathcal{O}(1)}$, the following bound holds true.

\begin{equation}
\mathcal{E}_U(\hat{f})\leq  \mathcal{O}\left(\left(\frac{1}{\log n_s}\right)^{1/4}+\left(\frac{\log n_s}{c_\mu n_s}\right)^{1/{d}} \right)^{{1+\beta}},\label{eq:xue_2022_nov_11_1}
\end{equation}

\subsection{Analysis of the excess error upper-bound under posterior-shift settings}
\textbf{Under the posterior-shift setting}, the support of $\mathbbm{D}_S$ and $\mathbbm{D}_U$ are the same, i.e., $\text{Supp}(\mathbbm{D}_S)=\text{Supp}(\mathbbm{D}_U)=\Omega$. The regression functions $\eta_U$ and $\eta_S$ are different. Then we have the following.
\begin{equation}
\begin{aligned}
\left|\hat{\eta}_k(x_u) -\eta_U(x_u) \right |&=\left| \frac{1}{k}\sum_{i=1}^k y_s^{(i)}-\eta_U(x_u) \right|\\
&\leq \left|\frac{1}{k}\sum_{i=1}^k y_s^{(i)}-\frac{1}{k}\sum_{i=1}^k\eta_S\left(x_s^{(i)}\right)\right|+\left| \frac{1}{k}\sum_{i=1}^k\eta_S\left(x_s^{(i)}\right)-\eta_U(x_u)  \right|\\
&=\frac{1}{k}\left| \sum_{i=1}^k y_s^{(i)}-\sum_{i=1}^k\eta\left(x_s^{(i)}\right)\right|+\frac{1}{k}\sum_{i=1}^k\left| \eta_S\left(x_s^{(i)}\right)-\eta_U(x_u)  \right|\\
&=\frac{1}{k} \left| \sum_{i=1}^ky_s^{(i)}-\sum_{i=1}^k\eta\left(x_s^{(i)}\right)\right|+\frac{1}{k}\sum_{i=1}^k\left| \eta_S\left(x_s^{(i)}\right)-\eta_S(x_u)+\eta_S(x_u)-\eta_U(x_u)  \right|\\
&\leq\frac{1}{k} \left| \sum_{i=1}^ky_s^{(i)}-\sum_{i=1}^k\eta\left(x_s^{(i)}\right)\right|+\frac{1}{k}\sum_{i=1}^k\left| \eta_S\left(x_s^{(i)}\right)-\eta_S(x_u)\right|+ \underbrace{\left|\eta_S(x_u)-\eta_U(x_u)  \right|}_{\text{Adaptivity gap}}
\end{aligned}
\label{equ:bound_post}
\end{equation}
Compared to \myref{equ:bound_cova}, \myref{equ:bound_post} has an additional term $\left|\eta_S(x_u)-\eta_U(x_u)  \right|$ (the adaptivity gap~\cite{zhang2022domain}), which measure the difference of two regression functions directly. Although previous work has similar definition, for example, the regression functions difference defined in~\cite{zhao2019learning}: $\min\{{\mathbb{E}_{\mathbbm{D}_S}[|\eta_S-\eta_U|]},\mathbb{E}_{\mathbbm{D}_U}[|\eta_S-\eta_U|]\}$, which care about ``how $\eta_U$ performs on source data''. In comparison, our definition is more similar to~\cite{zhang2022domain}, which only focuses on the regression difference when evaluated on examples from the target domain and shown to be more practical and intuitive~\cite{kpotufe2018marginal,zhang2022domain}.

We assume that $|\eta_S-\eta_U|$ is upper bounded by some constant $C_{ada}$, namely $\sup_{x_u\in \mathbbm{D}_U} |\eta_S(x_u)-\eta_U(x_u)|\leq C_{ada}$, under the posterior-shift setting, we have
\begin{equation}
\mathcal{E}_U(\hat{f})\leq \left(\left(\frac{1}{k}\right)^{1/4}+C_1  \left(\frac{k}{c_\mu n_s}\right)^{1/{d}} +C_{ada} \right)^{{1+\beta}}
\end{equation}

Via the similar duration of inequality \myref{eq:xue_2022_nov_11_1},  with at least $ 1-3\exp(-\sqrt{\log n_s})^{\mathcal{O}(1)}$ probability we have, 
\begin{equation}
\mathcal{E}_U(\hat{f})\leq  \mathcal{O}\left(\left(\frac{1}{\log n_s}\right)^{1/4}+C_1  \left(\frac{\log n_s}{c_\mu n_s}\right)^{1/{d}} +C_{ada}\right)^{{1+\beta}}.
\end{equation}
\subsection{Effect of utilizing online target samples (Proof of Proposition 3)}\label{sec:app_target_sample}

Despite the assumptions and notions mentioned above, to study the effect of target data, we denote $\{x_s^{(i)},y_s^{(i)}\}_{i=1}^{k_s}+\{x_u^{(i)},y_u^{(i)}\}_{i=1}^{k_u}$ as the nearest data and the corresponding labels to $x_u$ in $\mathcal{B}(x_u,r_0)$, where $k_s+k_u=k$ and $y_u^{(i)}$ is the pseudo-label of $x_u^{(i)}$, that is, $y_u^{(i)}=\mathbb{I}\{\hat{\eta}_k(x_u^{(i)})\geq 1/2\}$. The KNN classification result will be $\hat{\eta}_k(x_u)=\frac{1}{k}\sum_{i=1}^{k_s} y_s^{(i)}+\frac{1}{k}\sum_{i=1}^{k_u} y_u^{(i)}$. We have the following.

\begin{equation}
\begin{aligned}
\left|\hat{\eta}_k(x_u) -\eta(x_u) \right |&=\left| \frac{1}{k}\sum_{i=1}^{k_s} y_s^{(i)}-\frac{1}{k}\sum_{i=1}^{k_s}\eta(x_u) +\frac{1}{k}\sum_{i=1}^{k_u} y_u^{(i)}-\frac{1}{k}\sum_{i=1}^{k_u}\eta(x_u)  \right|\\
& \leq \left|\frac{1}{k}\sum_{i=1}^{k_s} y_s^{(i)}-\frac{1}{k}\sum_{i=1}^{k_s} \eta\left(x_s^{(i)}\right)\right|+\left| \frac{1}{k}\sum_{i=1}^{k_s} \eta\left(x_s^{(i)}\right)-\frac{k_s}{k}\eta(x_u)  \right|\\
&\quad\quad\quad +\left|\frac{1}{k}\sum_{i=1}^{k_u} y_u^{(i)}-\frac{1}{k}\sum_{i=1}^{k_u} \eta\left(x_u^{(i)}\right)\right|+\left| \frac{1}{k}\sum_{i=1}^{k_u} \eta\left(x_u^{(i)}\right)-\frac{k_u}{k}\eta(x_u)  \right|\\
&\leq \underbrace{\frac{1}{k}\left|\sum_{i=1}^{k_s} y_s^{(i)}+\sum_{i=1}^{k_u}y_u^{(i)}-\sum_{i=1}^{k_s}\eta\left(x_s^{(i)}\right)-\sum_{i=1}^{k_u}\eta\left(x_u^{(i)}\right)\right|}_{\rm \circled{\rm 1}}+\underbrace{\frac{1}{k}\sum_{i=1}^{k_s}\left| \eta\left(x_s^{(i)}\right)-\eta(x_u)  \right|}_{\rm \circled{\rm 2}}\\
&\quad\quad\quad+\underbrace{\frac{1}{k}\sum_{i=1}^{k_u}\left| \eta\left(x_u^{(i)}\right)-\eta(x_u)  \right|}_{\rm \circled{\rm 3}}
\end{aligned}
\end{equation}
Although the true labels of target samples are unknown, we store the target sample into the KNN query set only when its prediction confidence is large enough. Therefore, it is natural to assume that $\mathbb{E}_{Y|X}[y_u^{(i)}]=\eta(x_u^{(i)})$. According to \myref{equ:21}, we have
\begin{equation}
\begin{aligned}
&P_{X,Y}\left(\frac{1}{k}\left|\sum_{i=1}^{k_s} y_s^{(i)}+\sum_{i=1}^{k_u}y_u^{(i)}-\sum_{i=1}^{k_s}\eta\left(x_s^{(i)}\right)-\sum_{i=1}^{k_u}\eta\left(x_u^{(i)}\right)\right|\right)\\
&=\mathbb{E}_X\left[P_{Y|X}\left(\frac{1}{k}\left|\sum_{i=1}^{k_s} y_s^{(i)}+\sum_{i=1}^{k_u}y_u^{(i)}-\sum_{i=1}^{k_s}\eta\left(x_s^{(i)}\right)-\sum_{i=1}^{k_u}\eta\left(x_u^{(i)}\right)\right|
\right)\right]\leq 2\exp({-2k\epsilon^2})
\end{aligned}
\label{equ:27}
\end{equation}
Set $\epsilon=(1/k)^{1/4}$, we have, with probability, at least $1-2\exp(-2\sqrt{k})$, ${\rm \circled{\rm 1}}\leq (1/k)^{1/4}$. Then, according to \myref{equ:19}, we have
\begin{equation}
{\rm \circled{\rm 2}}\leq \frac{k_s}{k} C\left(\frac{2k_s}{c_\mu\mu_-\pi_dn_s}\right)^{1/d}; {\rm \circled{\rm 3}}\leq \frac{k_u}{k} C\left(\frac{2k_u}{c_\mu^*\mu_-\pi_dn_u}\right)^{1/d}
\end{equation}
Finally, the excess error under the covariate shift setting can be bounded by
\begin{equation}
\begin{aligned}
\mathcal{E}_U(\hat{f})&\leq 2C_\beta\left( 
(1/k)^{1/4}
+ \frac{k_s}{k}C\left(\frac{2k_s}{c_\mu\mu_-\pi_dn_s}\right)^{1/d} + \frac{k_u}{k} C\left(\frac{2k_u}{c_\mu^*\mu_-\pi_dn_u}\right)^{1/d}
\right)^{{1+\beta}}\\
& \approx  \left(\left(\frac{1}{k}\right)^{1/4}+C_1 k_s  \left(\frac{k_s}{c_\mu n_s}\right)^{1/{d}} +C_1 k_u \left(\frac{k_u}{c_\mu^* n_u}\right)^{1/{d}}  \right)^{{1+\beta}}
\end{aligned}
\label{bound:28}
\end{equation}
Compared \myref{bound:28} to \myref{bound:22}, it is easy to verify that
\begin{equation}
\begin{aligned}
& \frac{k}{k}C\left(\frac{2k}{c_\mu\mu_-\pi_dn_s}\right)^{1/{d}}-\frac{k_s}{k}C\left(\frac{2k_s}{c_\mu\mu_-\pi_dn_s}\right)^{1/d} - \frac{k_u}{k} C\left(\frac{2k_u}{c_\mu^*\mu_-\pi_dn_u}\right)^{1/d}\\
&\geq \frac{k_u}{k}C\left(\frac{2k_s}{c_\mu\mu_-\pi_dn_s}\right)^{1/d}- \frac{k_u}{k} C\left(\frac{2k_u}{c_\mu^*\mu_-\pi_dn_u}\right)^{1/d}
\end{aligned}
\end{equation}
Because in general, we have $c_\mu^*>c_u$, the difference is then larger than $0$, namely incorporating target samples into the KNN memory bank, the excess error can be further reduced. When $\mathbbm{D}_S$ is very close to $\mathbbm{D}_U$, that is, $c_\mu^*\approx c_\mu$, the two bounds will be similar. 


Similar results can be derived under the posterior-shift setting. Under the assumption that $\mathbb{E}_{Y|X}[y_u^{(i)}]=\eta_U(x_u^{(i)})$ and $\sup_{x_u\in \mathbbm{D}_U} |\eta_S(x_u)-\eta_U(x_u)|\leq C_{ada}$, we have

\begin{equation}
\begin{aligned}
\left|\hat{\eta}_k(x_u) -\eta_U(x_u) \right |&=\left| \frac{1}{k}\sum_{i=1}^{k_s} y_s^{(i)}-\frac{1}{k}\sum_{i=1}^{k_s}\eta_U(x_u) +\frac{1}{k}\sum_{i=1}^{k_u} y_u^{(i)}-\frac{1}{k}\sum_{i=1}^{k_u}\eta_U(x_u)  \right|\\
& \leq \left|\frac{1}{k}\sum_{i=1}^{k_s} y_s^{(i)}-\frac{1}{k}\sum_{i=1}^{k_s} \eta_S\left(x_s^{(i)}\right)\right|+\left| \frac{1}{k}\sum_{i=1}^{k_s} \eta_S\left(x_s^{(i)}\right)-\frac{k_s}{k}\eta_U(x_u)  \right|\\
&\quad\quad\quad +\left|\frac{1}{k}\sum_{i=1}^{k_u} y_u^{(i)}-\frac{1}{k}\sum_{i=1}^{k_u} \eta_U\left(x_u^{(i)}\right)\right|+\left| \frac{1}{k}\sum_{i=1}^{k_u} \eta_U\left(x_u^{(i)}\right)-\frac{k_u}{k}\eta_U(x_u)  \right|\\
&\leq \frac{1}{k}\left|\sum_{i=1}^{k_s} y_s^{(i)}+\sum_{i=1}^{k_u}y_u^{(i)}-\sum_{i=1}^{k_s}\eta_S\left(x_s^{(i)}\right)-\sum_{i=1}^{k_u}\eta_U\left(x_u^{(i)}\right)\right|+\frac{1}{k}\sum_{i=1}^{k_s}\left| \eta_S\left(x_s^{(i)}\right)-\eta_S(x_u)  \right|\\
&\quad\quad\quad+\frac{1}{k}\sum_{i=1}^{k_u}\left| \eta_U\left(x_u^{(i)}\right)-\eta_U(x_u)  \right| + \frac{k_s}{k}\left|\eta_S(x_u)-\eta_U(x_u)  \right|,
\end{aligned}
\end{equation}
and the conclusion will be
\begin{equation}
\begin{aligned}
\mathcal{E}_U(\hat{f})&\leq 2C_\beta\left( 
(1/k)^{1/4}
+ \frac{k_s}{k}C\left(\frac{2k_s}{c_\mu\mu_-\pi_dn_s}\right)^{1/d} + \frac{k_u}{k} C\left(\frac{2k_u}{c_\mu^*\mu_-\pi_dn_u}\right)^{1/d} +
\frac{k_s}{k} C_{ada}
\right)^{{1+\beta}}
\\
\end{aligned}
\end{equation}


\section{Online Optimization Algorithm for Optimizing the KNN Loss Function $\mathcal{L}_{KNN}$}\label{app:alg}

\begin{algorithm}[h]
\begin{algorithmic}
\caption{Online optimization algorithm for optimizing the KNN loss function $\mathcal{L}_{KNN}$.}
 \STATE {\bfseries Input:}{training data ${D}_S$, batch size $N$, learning rate $\eta$, training iterations $T$, Adam hyperparameters $\beta_1$, $\beta_2$.}\\
\STATE \textbf{Initial}: model parameters $\theta$ and memory bank $\mathcal{M}=\{h_\theta(x_i),y_i\}_{i=1}^M$ with a predefined size $M$, where $(x_i,y_i)$ are randomly sampled from $D_S$.\\
\FOR{t = $1, \dots, T$}
\STATE $(x_i,y_i)_{i=1}^N\sim{D}_S\qquad\qquad\qquad\qquad\qquad\qquad~\;\;// \textit{Data sampling}$ \\
 \STATE $\mathcal{L}_{KNN}=-\frac{1}{N}\sum_{i} \log \frac{\sum_{j\in B_{k,\theta,\mathcal{M}}(x_i)} \exp\left(w_{ij}/\tau \right)\mathbb{I}\{y_i=y_j\}}{\sum_{j\in B_{k,\theta,\mathcal{M}}(x_i)} \exp\left(w_{ij}/\tau \right)},  // \textit{Calculate the loss}$\\
 $\mathcal{M}\leftarrow \mathcal{M}\bigcup \{h_{\theta}(x_i),y_i\}_{i=1}^N \;\;// \textit{Update the memory bank by the first-in-first-out (FIFO) strategy}$ \\
 $\theta \leftarrow \operatorname{Adam}\left(\mathcal{L}_{KNN}, \theta, \eta, \beta_1, \beta_2 \right) 
 \qquad\qquad\qquad\qquad\;\;\;\;\;//\textit{Update model parameters}$
\ENDFOR
\label{algo:main}
\end{algorithmic}
\end{algorithm}
\section{Dataset and Implementation Details}\label{sec:data_detail}
\subsection{Dataset details}

\textbf{Rotated MNIST} \cite{ghifary2015domain} consists of 10,000 digits in MNIST with different rotated angles where the domain is determined by the degrees $d \in \{0, 15, 30, 45,
60, 75\}$.

\textbf{PACS} \cite{li2017deeper} includes 9, 991 images with 7 classes $y \in  \{$ dog, elephant, giraffe, guitar, horse, house, person $\}$ from 4 domains $d \in$ $\{$art, cartoons, photos, sketches$\}$. 

\textbf{VLCS} \cite{torralba2011unbiased} is composed of 10,729 images, 5 classes $y \in \{$ bird, car, chair, dog, person $\}$ from domains $d \in \{$Caltech101, LabelMe, SUN09, VOC2007$\}$. 

\textbf{TerraIncognita}~\cite{beery2018recognition} contains photographs of wild animals taken by camera traps at locations $d\in \{L100,L38, L43, L46\}$, with $24,788$ examples of dimension $(3,224,224)$ and $10$ classes.

\textbf{DomainNet}~\cite{peng2019moment} has six domains $d \in$ $\{$clipart, infograph, painting, quickdraw, real, sketch$\}$. This dataset contains $586,575$ examples of sizes $(3,224,224)$ and $345$ classes.

\subsection{Implementation and hyper-parameter details}

\textbf{Hyperparameter search.} Following the experimental settings in \cite{gulrajani2021in}, we conduct a random search of 20 trials over the hyperparameter distribution for each algorithm and test domain. Specifically, we split the data from each domain into $80\%$ and $20\%$ proportions, where the larger split is used for training and evaluation, and the smaller ones are used for select hyperparameters. We repeat the entire experiment twice using three different seeds to reduce the randomness. Finally, we report the mean over these repetitions as well as their estimated standard error. 

\textbf{Model selection.} The model selection in domain generalization is intrinsically a learning problem, and we use both the test-domain validation and training domain validation, two of the three selection methods in \cite{gulrajani2021in}. Test-domain validation chooses the model maximizing the accuracy on a validation set that follows the distribution of the test domain. Training domain validation chooses the model with the highest average source domain accuracy. In the main paper, test domain validation results are presented by default.


\textbf{Model architectures.}
Following \cite{gulrajani2021in}, we use ConvNet (Table.\ref{tab:convnet}) as the encoder for RotatedMNIST (detailed in Appendix D.1 in \cite{gulrajani2021in}) with MIT License. For other datasets, torch-vision for ResNet18 and ResNet50 (Apache-2.0), \textit{timm} for Vision Transformer (Apache-2.0), and the official repository of T3A (MIT License) are used.
\begin{table}
\centering
\begin{tabular}{@{}ll@{}}
\toprule
\# & Layer                                                     \\ \midrule
1  & \cellcolor[HTML]{FFFFFF}Conv2D (in=d, out=64)             \\
2  & \cellcolor[HTML]{FFFFFF}ReLU                              \\
3  & \cellcolor[HTML]{FFFFFF}GroupNorm (groups=8)              \\
4  & \cellcolor[HTML]{FFFFFF}Conv2D (in=64, out=128, stride=2) \\
5  & \cellcolor[HTML]{FFFFFF}ReLU                              \\
6  & \cellcolor[HTML]{FFFFFF}GroupNorm (8 groups)              \\
7  & \cellcolor[HTML]{FFFFFF}Conv2D (in=128, out=128)          \\
8  & \cellcolor[HTML]{FFFFFF}ReLU                              \\
9  & \cellcolor[HTML]{FFFFFF}GroupNorm (8 groups)              \\
10 & \cellcolor[HTML]{FFFFFF}Conv2D (in=128, out=128)          \\
11 & \cellcolor[HTML]{FFFFFF}ReLU                              \\
12 & \cellcolor[HTML]{FFFFFF}GroupNorm (8 groups)              \\
13 & \cellcolor[HTML]{FFFFFF}Global average-pooling            \\ \bottomrule
\end{tabular}
\caption{Details of our MNIST ConvNet architecture. All convolutions use 3×3 kernels and ``same'' padding}\label{tab:convnet}
\end{table}
We run our experiments mainly on Tesla-V100 (32G)x4 instances.
\section{Additional Experimental Results}\label{sec:addexp}

\subsection{Detailed generalization results}

Tables~\ref{tab:rotatedmnist},~\ref{tab:vlcs},~\ref{tab:pacs},~\ref{tab:domainnet} contain detailed results for each dataset with ’Test-domain’ and ’Training-domain’ model selection methods.


\subsection{AdaNPC can mitigate the issue of domain forgetting}\label{app:forget}
At first, recall our experimental setting is: we first trained our model on $d_0$ of the dataset as shown in Figure~\ref{fig:succ_dg}. Then, we utilized the TTA algorithm to adapt the model to $d_1,...,d_n$ one by one. To illustrate this problem more vividly, we conducted experiments on several datasets, and their results are presented in Table~\ref{reval-1} and Table~\ref{reval-2}. These results demonstrate that as the TTA progresses, the model's performance on the source domain declines, indicating that TTA causes the model to forget the knowledge learned from the source domain. In practical scenarios, the model is required to perform for an extended period and may encounter numerous novel data. Under such circumstances, the model's performance on the source domain may suffer greatly, leading to inaccurate predictions on the source domain samples. As can be seen, for most existing baselines, the model forgets the knowledge in the source domain with the progress of TTA on all datasets, while AdaNPC can avoid this situation.

\begin{table}[]
\centering
\begin{tabular}{ccccccccccc}
\toprule
\multirow{2}{*}{Method} & \multicolumn{4}{c}{\textbf{PACS}} &  & \multicolumn{5}{c}{\textbf{Terr}} \\ \cmidrule(lr){2-6} \cmidrule(lr){7-11}
 & A & C & P & S & Avg & L100 & L38 & L43 & L46 & Avg \\
T3A & 94.5 & 94.2 & 93.7 & 91 & 93.4 & 89.6 & 86.7 & 85.5 & 83.8 & 86.4 \\
Tent & 94.5 & 94.2 & 93.1 & 92.2 & 93.5 & 89.6 & 89.1 & 88.6 & 82.3 & 87.4 \\
PLFull & 94.5 & 94.1 & 93.3 & 90.2 & 93.0 & 89.6 & 81.22 & 65.8 & 55.1 & 72.9 \\ \rowcolor{Gray}
AdaNPC & 94.5 & 94.5 & 94.5 & 94.5 & 94.5 & 89.6 & 89.6 & 89.6 & 89.6 & 89.6 \\ \bottomrule
\end{tabular}%
\caption{\textbf{Successive adaptation results on the PACS and TerraIncognita datasets.} The metric is the re-evaluation accuracy of the adapted model on the source domain $d_0$.}\label{reval-1}
\end{table}

\begin{table}[]
\centering
\begin{tabular}{lccccccc}
\toprule
\textbf{DomainNet} & \textbf{clip} & \textbf{info} & \textbf{paint} & \textbf{quick} & \textbf{real} & \textbf{sketch} & \textbf{Avg} \\
T3A & 63.4 & 37.7 & 45.7 & 45.7 & 50.0 & 51.1 & 48.9 \\
Tent & 63.4 & 50.4 & 21.2 & 8.2 & 6.9 & 5.3 & 25.9 \\
PLFull & 63.4 & 61.3 & 55.5 & 16.5 & 10.0 & 7.0 & 35.6 \\ \rowcolor{Gray}
AdaNPC & 63.4 & 63.4 & 63.4 & 63.4 & 63.4 & 63.4 & 63.4 \\ \hline
\textbf{RMNIST} & \textbf{0.0} & \textbf{15.0} & \textbf{30.0} & \textbf{45.0} & \textbf{60.0} & \textbf{75.0} & \textbf{Avg} \\
T3A & 100.0 & 96.1 & 96.1 & 93.2 & 90.3 & 89.9 & 94.3 \\
Tent & 100.0 & 99.5 & 96.5 & 92.5 & 89.3 & 89.4 & 94.5 \\
PLFull & 100.0 & 98.3 & 87.5 & 79.2 & 67.9 & 63.5 & 82.7 \\  \rowcolor{Gray}
AdaNPC & 100.0 & 100.0 & 100.0 & 100.0 & 100.0 & 100.0 & 100.0 \\ \bottomrule
\end{tabular}%
\caption{\textbf{Successive adaptation results on the DomainNet and RMNIST datasets.} The metric is the re-evaluation accuracy of the adapted model on the source domain $d_0$.}\label{reval-2}
\end{table}

\subsection{Extended experiments on Cifar-10-C and ImageNet-C}\label{sec:large}

\begin{figure*}
    \centering
    \includegraphics[width=\textwidth]{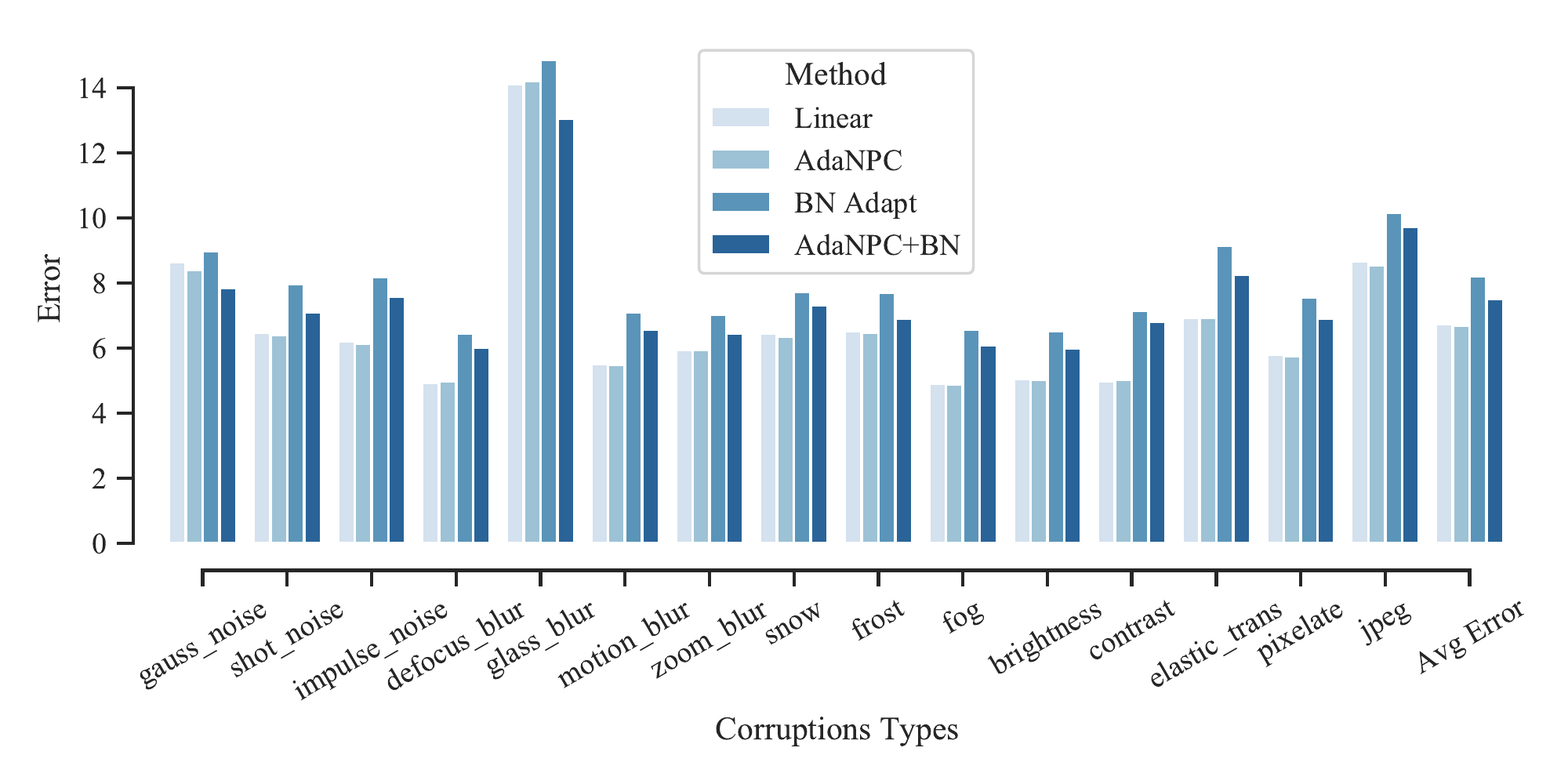}
    \caption{\textbf{Corruption benchmark on CIFAR-10-C with the lowest severity (one)}. \knn+BN means that the KNN classifier and BN retraining are both used.}
    \label{fig:cifar10-1}
\end{figure*}

\begin{figure*}
    \centering
    \includegraphics[width=\textwidth]{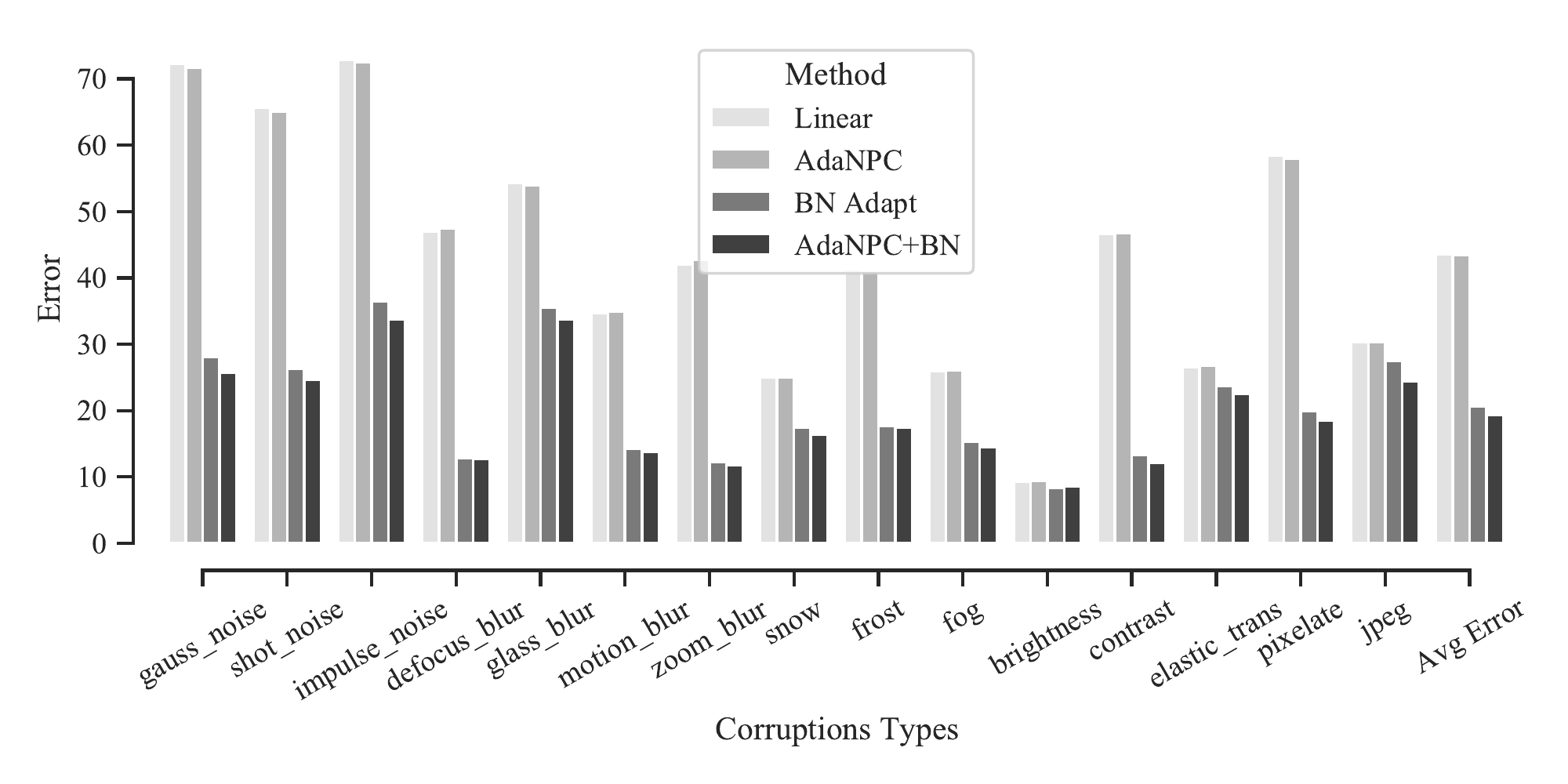}
    \caption{\textbf{Corruption benchmark on CIFAR-10-C with the highest severity (five) and a 40-2 Wide ResNet backbone~\cite{zagoruyko2016wide} pre-trained on CIFAR-10.} \knn+BN means that the KNN classifier and BN retraining are both used.}
    \label{fig:cifar10-3}
\end{figure*}

\begin{figure*}
    \centering
    \includegraphics[width=\textwidth]{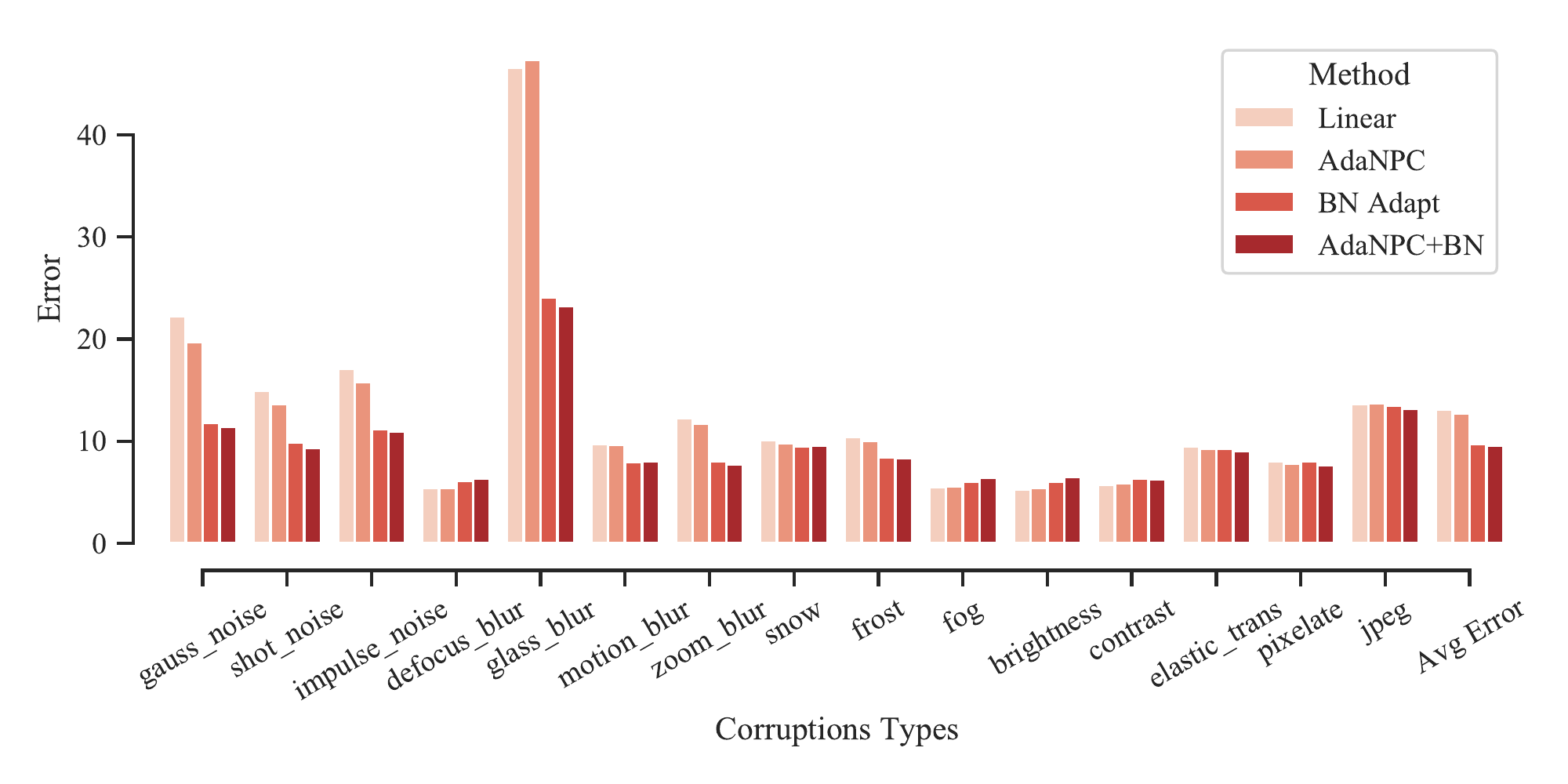}
    \caption{\textbf{Corruption benchmark on CIFAR-10-C with the lowest severity (one) and a 40-2 Wide ResNet backbone~\cite{zagoruyko2016wide} pre-trained on CIFAR-10}. \knn+BN means that the KNN classifier and BN retraining are both used.}
    \label{fig:cifar10-4}
\end{figure*}

This paper focus on domain generalization (DG), where TTA is considered as one of the DG methods. Therefore, we emphasized the comparison with existing DG methods in the main text. However, we also conduct additional experiments on corruption datasets to verify the effectiveness of the proposed \knn. ~\figurename~\ref{fig:cifar10-1} shows the results with the lowest severity (one), where \knn performs the best and retraining the BN statistic will not be beneficial. For WRN-28-10 Wide ResNet Backbone, the results are shown in ~\figurename~\ref{fig:cifar10-3} and ~\figurename~\ref{fig:cifar10-4}, where the same pattern as results with 40-2 Wide ResNet Backbone are observed.

Considering large-scale corruption benchmarks, we compared the method with Tent and EATA ~\cite{niu2022efficient} on the ImageNet-C dataset. Experimentally, we compared the effectiveness of AdaNPC and EATA separately in Table.~\ref{tab:corr}, where the experimental and baselines are all following EATA ~\cite{niu2022efficient}. As we can see, we only need to simply replace the Linear layer with KNN and remember the trustworthy samples during testing, and AdaNPC can achieve much better performance than ResNet50 with a linear head. In addition, AdaNPC can be combined with existing TTA methods such as Tent, ETA, etc., and the final performance surpasses these methods individually. 

\textbf{Existing baselines will be highly affected by the batch size}. As can be seen in Table~\ref{tab:corr_bsz1} (left), existing methods tend to have very poor performance when the batch size is set to $1$, as the gradient noise for individual samples is very high, which is detrimental to model optimization. However, it should be emphasized that batch data does not align with the setting of online learning, where inference is required on-demand instead of waiting for an incoming batch or when inference is happening on an edge device (such as a mobile phone) where there is no opportunity for batching. Therefore, AdaNPC, a TTA method that is insensitive to batch size, is valuable for the current research field. In addition, the three baselines in the above experiments all suffered from severe collapse due to the influence of the BN layer, which may not be a fair comparison to AdaNPC. We additionally considered a setting where all BN layer parameters were frozen during testing, and the final linear layer was updated using the objective functions of each algorithm (Clf Retraining). The final results are shown in Table~\ref{tab:corr_bsz1}(right). It can be seen that even if the influence of the BN layer is excluded, high-noisy gradients can still make the models perform very poorly. Although ETA and EATA, as strong baselines, perform much better than using Tent alone, they are still heavily affected by the batch size.

\begin{table}[]
\centering
\begin{tabular}{lcccccccc}
\toprule
\multirow{2}{*}{Method} & \multicolumn{4}{c}{BN Retraining} & \multicolumn{4}{c}{Clf Retraining} \\ \cmidrule(lr){2-5} \cmidrule(lr){6-9}
 & Defoc & Glass & Motion & Zoom & Defoc & Glass & Motion & Zoom \\ \midrule
R-50 (BN) & 82.10 & 90.20 & 85.20 & 77.50 & 82.10 & 90.20 & 85.20 & 77.50 \\
Tent & 99.10 & 99.10 & 99.11 & 99.10 & 0.90 & 0.90 & 0.89 & 0.90 \\
ETA & 99.14 & 99.14 & 99.14 & 99.16 & 0.86 & 0.86 & 0.86 & 0.84 \\
EATA & 99.14 & 99.14 & 99.14 & 99.16 & 0.86 & 0.86 & 0.86 & 0.84 \\ \rowcolor{Gray}
AdaNPC & 83.10 & 83.00 & 72.30 & 60.57 & 83.10 & 83.00 & 72.30 & 60.57 \\ \bottomrule
\end{tabular}%
\caption{Comparison with state-of-the-art methods on ImageNet-C with the highest severity level 5 regarding corruption Error, where the TTA batch size is set to 1.}\label{tab:corr_bsz1}
\end{table}


\subsection{Extended ablation studies of training algorithms}
 Results in Table.~\ref{tab:knn-training} shows that on the Rotated MNIST dataset, with $\mathcal{L}_{KNN}$, the representation will be better and the generalization results will be improved.
\begin{table*}[]
\centering
\begin{tabular}{@{}lccccccc@{}}
\toprule
 & $d_0$ & $d_1$ & $d_2$ & $d_3$ & $d_4$ & $d_5$ & Avg \\ \midrule
ERM & 94.0 $\pm$ 0.2 & 98.1 $\pm$ 0.1 & 98.8 $\pm$ 0.4 & 99.1 $\pm$ 0.1 & \textbf{98.9 $\pm$ 0.1} & 96.4 $\pm$ 0.1 & 97.6 \\
ERM+\knn & 96.6 $\pm$ 0.4 & 98.7 $\pm$ 0.1 & 98.9 $\pm$ 0.3 & 99.1 $\pm$ 0.1 & 98.7 $\pm$ 0.1 & 96.7 $\pm$ 0.3 & 98.1 \\
$\mathcal{L}_{KNN}$ & 96.1 $\pm$ 0.9 & 98.6 $\pm$ 0.3 & 98.9 $\pm$ 0.0 & 99.1 $\pm$ 0.2 & 98.8 $\pm$ 0.2 & 97.0 $\pm$ 0.5 & 98.1 \\
$\mathcal{L}_{KNN}$ + \knn & \textbf{96.9} $\pm$ 0.3 & \textbf{98.8 $\pm$ 0.1} & \textbf{99.1 $\pm$ 0.2} & \textbf{99.2 $\pm$ 0.1} & \textbf{98.9 $\pm$ 0.1} & \textbf{98.0 $\pm$ 0.1} & \textbf{98.5} \\ \bottomrule
\end{tabular}%

\caption{Ablation studies of training loss on Rotated MNIST.}
\label{tab:knn-training}
\end{table*}

\subsection{Extended visualization of classification results.}
\begin{figure}
    \centering
   \includegraphics[width=\textwidth]{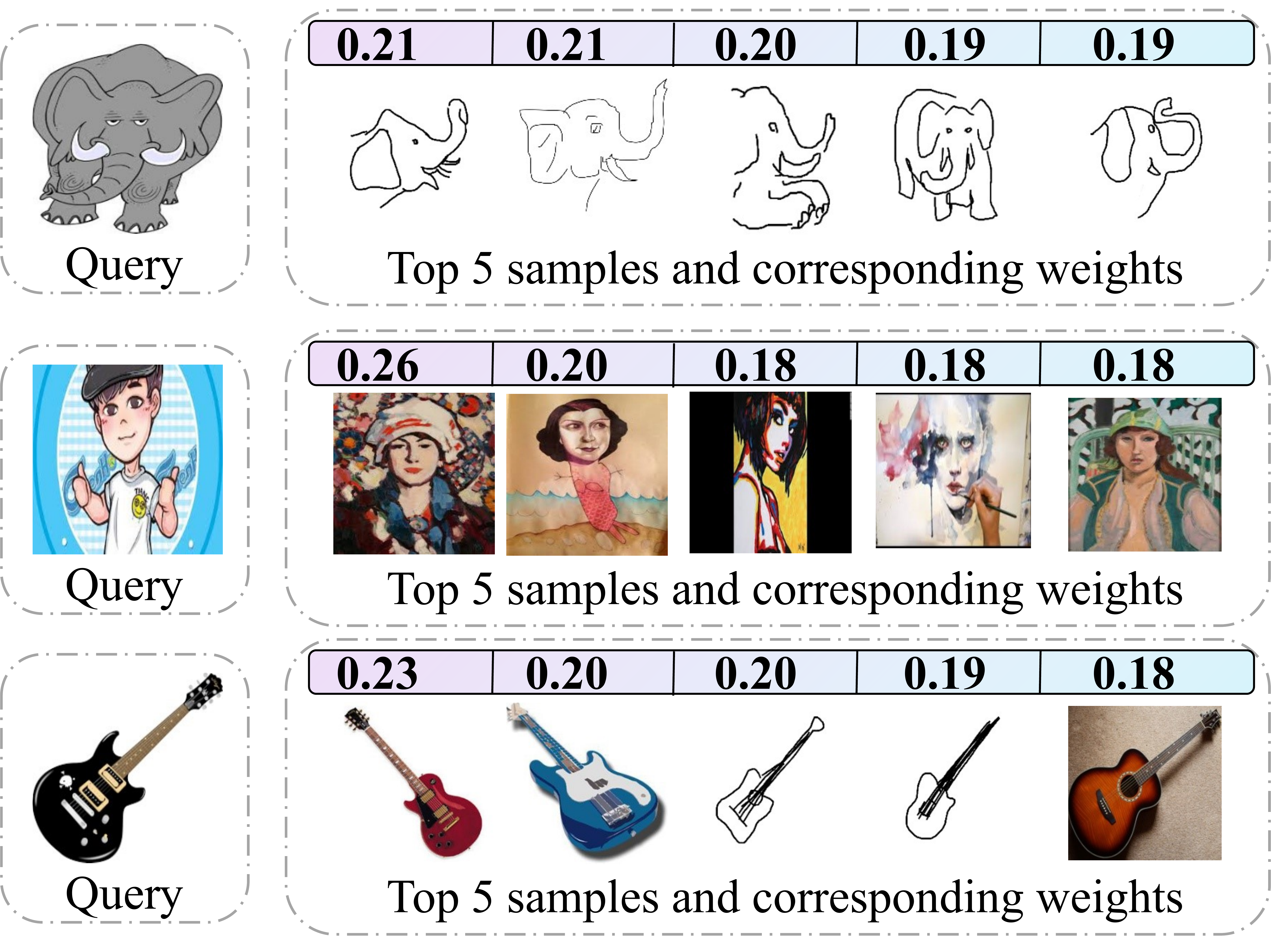}
    \caption{Visualization of successfully classified results attained by \knn.}
    \label{fig:succ_app}
\end{figure}

\begin{figure}
    \centering
    \includegraphics[width=\textwidth]{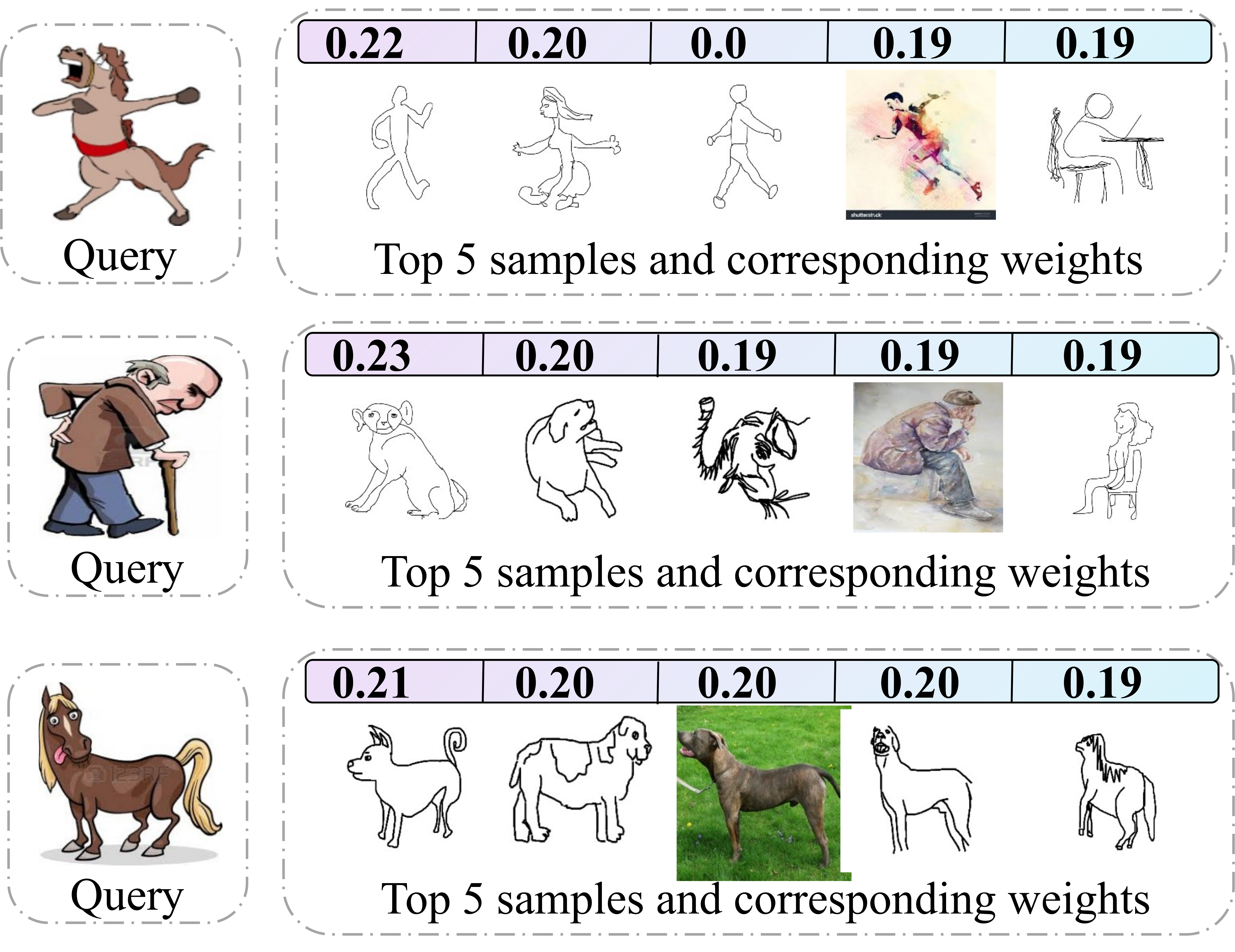}
    \caption{Visualization of misclassified results attained by \knn.}
    \label{fig:fail_app}
\end{figure}

\figurename~\ref{fig:succ_app} provide more evaluation instances that be prediction correctly and \figurename~\ref{fig:fail_app} supplies more failure cases.

\begin{table}[t]
\begin{center}
\adjustbox{max width=\textwidth}{%
\begin{tabular}{lccccccc}
\toprule
\multicolumn{8}{c}{\textbf{Rotated MNIST, Model selection: ‘Test-domain’ validation set}}\\
\textbf{Algorithm}   & \textbf{0}           & \textbf{15}          & \textbf{30}          & \textbf{45}          & \textbf{60}          & \textbf{75}          & \textbf{Avg}         \\
\midrule
ERM~\cite{vapnik1998statistical}                  & 95.3 $\pm$ 0.2       & 98.7 $\pm$ 0.1       & 98.9 $\pm$ 0.1       & 98.7 $\pm$ 0.2       & 98.9 $\pm$ 0.0       & 96.2 $\pm$ 0.2       & 97.8                 \\
IRM~\cite{arjovsky2020invariant}                  & 94.9 $\pm$ 0.6       & 98.7 $\pm$ 0.2       & 98.6 $\pm$ 0.1       & 98.6 $\pm$ 0.2       & 98.7 $\pm$ 0.1       & 95.2 $\pm$ 0.3       & 97.5                 \\
GDRO~\cite{sagawa2020distributionally}             & 95.9 $\pm$ 0.1       & 99.0 $\pm$ 0.1       & 98.9 $\pm$ 0.1       & 98.8 $\pm$ 0.1       & 98.6 $\pm$ 0.1       & 96.3 $\pm$ 0.4       & 97.9                 \\
Mixup~\cite{yan2020improve}                & 95.8 $\pm$ 0.3       & 98.7 $\pm$ 0.0       & 99.0 $\pm$ 0.1       & 98.8 $\pm$ 0.1       & 98.8 $\pm$ 0.1       & 96.6 $\pm$ 0.2       & 98.0                 \\
MLDG~\cite{li2018learning}                 & 95.7 $\pm$ 0.2       & 98.9 $\pm$ 0.1       & 98.8 $\pm$ 0.1       & 98.9 $\pm$ 0.1       & 98.6 $\pm$ 0.1       & 95.8 $\pm$ 0.4       & 97.8                 \\
CORAL~\cite{sun2016deep}                & 96.2 $\pm$ 0.2       & 98.8 $\pm$ 0.1       & 98.8 $\pm$ 0.1       & 98.8 $\pm$ 0.1       & 98.9 $\pm$ 0.1       & 96.4 $\pm$ 0.2       & 98.0                 \\
MMD~\cite{li2018domain}                  & 96.1 $\pm$ 0.2       & 98.9 $\pm$ 0.0       & 99.0 $\pm$ 0.0       & 98.8 $\pm$ 0.0       & 98.9 $\pm$ 0.0       & 96.4 $\pm$ 0.2       & 98.0                 \\
DANN~\cite{ganin2016domain}                 & 95.9 $\pm$ 0.1       & 98.9 $\pm$ 0.1       & 98.6 $\pm$ 0.2       & 98.7 $\pm$ 0.1       & 98.9 $\pm$ 0.0       & 96.3 $\pm$ 0.3       & 97.9                 \\
CDANN~\cite{li2018deep}               & 95.9 $\pm$ 0.2       & 98.8 $\pm$ 0.0       & 98.7 $\pm$ 0.1       & 98.9 $\pm$ 0.1       & 98.8 $\pm$ 0.1       & 96.1 $\pm$ 0.3       & 97.9                 \\
MTL~\cite{blanchard2021domain}                  & 96.1 $\pm$ 0.2       & 98.9 $\pm$ 0.0       & 99.0 $\pm$ 0.0       & 98.7 $\pm$ 0.1       & 99.0 $\pm$ 0.0       & 95.8 $\pm$ 0.3       & 97.9                 \\
SagNet~\cite{nam2021reducing}               & 95.9 $\pm$ 0.1       & 99.0 $\pm$ 0.1       & 98.9 $\pm$ 0.1       & 98.6 $\pm$ 0.1       & 98.8 $\pm$ 0.1       & 96.3 $\pm$ 0.1       & 97.9                 \\
ARM~\cite{zhang2021adaptive}                  & 95.9 $\pm$ 0.4       & 99.0 $\pm$ 0.1       & 98.8 $\pm$ 0.1       & 98.9 $\pm$ 0.1       & 99.1 $\pm$ 0.1       & 96.7 $\pm$ 0.2       & 98.1                 \\
VREx~\cite{krueger2021out}                 & 95.5 $\pm$ 0.2       & 99.0 $\pm$ 0.0       & 98.7 $\pm$ 0.2       & 98.8 $\pm$ 0.1       & 98.8 $\pm$ 0.0       & 96.4 $\pm$ 0.0       & 97.9                 \\
RSC~\cite{huang2020self}                  & 95.4 $\pm$ 0.1       & 98.6 $\pm$ 0.1       & 98.6 $\pm$ 0.1       & 98.9 $\pm$ 0.0       & 98.8 $\pm$ 0.1       & 95.4 $\pm$ 0.3       & 97.6                 \\
Fish~\cite{shi2022gradient}&        &       &        &        &       &        & 97.9                \\
Fisher~\cite{rame2022fishr} & 95.8 $\pm$ 0.1       & 98.3 $\pm$ 0.1       & 98.8 $\pm$ 0.1       & 98.6 $\pm$ 0.3       & 98.7 $\pm$ 0.1       & 96.5 $\pm$ 0.1       & 97.8              \\
\rowcolor{Gray}
\knn                   & {97.2 $\pm$ 0.1}       & \textbf{99.2 $\pm$ 0.0}       & \textbf{99.1 $\pm$ 0.0}       & \textbf{99.0 $\pm$ 0.1}       & \textbf{99.2 $\pm$ 0.0}       & \textbf{97.6 $\pm$ 0.1}       & \textbf{98.5}                 \\
\rowcolor{Gray}
\knn+NP                   & \textbf{97.4 $\pm$ 0.1}       & {99.0 $\pm$ 0.0}       & {98.8 $\pm$ 0.0}       & {98.8 $\pm$ 0.1}       & {99.0 $\pm$ 0.1}       & {97.3 $\pm$ 0.1}       & {98.4}                 \\

\bottomrule
\multicolumn{8}{c}{\textbf{Rotated MNIST, Model selection: ‘Training-domain’ validation set}}\\
\textbf{Algorithm}   & \textbf{0}           & \textbf{15}          & \textbf{30}          & \textbf{45}          & \textbf{60}          & \textbf{75}          & \textbf{Avg}         \\
\midrule
ERM~\cite{vapnik1998statistical}                  & 95.9 $\pm$ 0.1       & 98.9 $\pm$ 0.0       & 98.8 $\pm$ 0.0       & 98.9 $\pm$ 0.0       & 98.9 $\pm$ 0.0       & 96.4 $\pm$ 0.0       & 98.0                 \\
IRM~\cite{arjovsky2020invariant}                   & 95.5 $\pm$ 0.1       & 98.8 $\pm$ 0.2       & 98.7 $\pm$ 0.1       & 98.6 $\pm$ 0.1       & 98.7 $\pm$ 0.0       & 95.9 $\pm$ 0.2       & 97.7                 \\
GDRO~\cite{sagawa2020distributionally}             & 95.6 $\pm$ 0.1       & 98.9 $\pm$ 0.1       & 98.9 $\pm$ 0.1       & 99.0 $\pm$ 0.0       & 98.9 $\pm$ 0.0       & 96.5 $\pm$ 0.2       & 98.0                 \\
Mixup~\cite{yan2020improve}                & 95.8 $\pm$ 0.3       & 98.9 $\pm$ 0.0       & 98.9 $\pm$ 0.0       & 98.9 $\pm$ 0.0       & 98.8 $\pm$ 0.1       & 96.5 $\pm$ 0.3       & 98.0                 \\
MLDG~\cite{li2018learning}                 & 95.8 $\pm$ 0.1       & 98.9 $\pm$ 0.1       & 99.0 $\pm$ 0.0       & 98.9 $\pm$ 0.1       & 99.0 $\pm$ 0.0       & 95.8 $\pm$ 0.3       & 97.9                 \\
CORAL~\cite{sun2016deep}                & 95.8 $\pm$ 0.3       & 98.8 $\pm$ 0.0       & 98.9 $\pm$ 0.0       & 99.0 $\pm$ 0.0       & 98.9 $\pm$ 0.1       & 96.4 $\pm$ 0.2       & 98.0                 \\
MMD~\cite{li2018domain}                  & 95.6 $\pm$ 0.1       & 98.9 $\pm$ 0.1       & 99.0 $\pm$ 0.0       & 99.0 $\pm$ 0.0       & 98.9 $\pm$ 0.0       & 96.0 $\pm$ 0.2       & 97.9                 \\
DANN~\cite{ganin2016domain}                 & 95.0 $\pm$ 0.5       & 98.9 $\pm$ 0.1       & 99.0 $\pm$ 0.0       & 99.0 $\pm$ 0.1       & 98.9 $\pm$ 0.0       & 96.3 $\pm$ 0.2       & 97.8                 \\
CDANN~\cite{li2018deep}                & 95.7 $\pm$ 0.2       & 98.8 $\pm$ 0.0       & 98.9 $\pm$ 0.1       & 98.9 $\pm$ 0.1       & 98.9 $\pm$ 0.1       & 96.1 $\pm$ 0.3       & 97.9                 \\
MTL~\cite{blanchard2021domain}                  & 95.6 $\pm$ 0.1       & 99.0 $\pm$ 0.1       & 99.0 $\pm$ 0.0       & 98.9 $\pm$ 0.1       & 99.0 $\pm$ 0.1       & 95.8 $\pm$ 0.2       & 97.9                 \\
SagNet~\cite{nam2021reducing}                & 95.9 $\pm$ 0.3       & 98.9 $\pm$ 0.1       & 99.0 $\pm$ 0.1       & \textbf{99.1 $\pm$ 0.0}       & 99.0 $\pm$ 0.1       & 96.3 $\pm$ 0.1       & 98.0                 \\
ARM~\cite{zhang2021adaptive}                  & 96.7 $\pm$ 0.2       & 99.1 $\pm$ 0.0       & 99.0 $\pm$ 0.0       & 99.0 $\pm$ 0.1       & {99.1 $\pm$ 0.1}       & 96.5 $\pm$ 0.4       & 98.2                 \\
VREx~\cite{krueger2021out}                 & 95.9 $\pm$ 0.2       & 99.0 $\pm$ 0.1       & 98.9 $\pm$ 0.1       & 98.9 $\pm$ 0.1       & 98.7 $\pm$ 0.1       & 96.2 $\pm$ 0.2       & 97.9                 \\
RSC~\cite{huang2020self}                  & 94.8 $\pm$ 0.5       & 98.7 $\pm$ 0.1       & 98.8 $\pm$ 0.1       & 98.8 $\pm$ 0.0       & 98.9 $\pm$ 0.1       & 95.9 $\pm$ 0.2       & 97.6                 \\
Fish~\cite{shi2022gradient}&        &       &        &        &       &        & 98.0                \\
Fisher~\cite{rame2022fishr}  & 95.0 $\pm$ 0.3       & 98.5 $\pm$ 0.0       & \textbf{99.2 $\pm$ 0.1}       & 98.9 $\pm$ 0.0       & 98.9 $\pm$ 0.1       & 96.5 $\pm$ 0.1       & 97.8                 \\
\rowcolor{Gray}
\knn                   & {97.7 $\pm$ 0.4}       & \textbf{99.1 $\pm$ 0.0}       & 99.1 $\pm$ 0.1       & \textbf{99.1 $\pm$ 0.1 }      & \textbf{99.2 $\pm$ 0.0 }      & {97.5 $\pm$ 0.2}       & {98.6 }                \\
\rowcolor{Gray}
\knn+BN                   & \textbf{97.9 $\pm$ 0.3}       & \textbf{99.1 $\pm$ 0.1}       & \textbf{99.2 $\pm$ 0.0}       & \textbf{99.1 $\pm$ 0.1}       & \textbf{99.2 $\pm$ 0.0}       & \textbf{98.0 $\pm$ 0.4}       & \textbf{98.8}                 \\
\bottomrule
\end{tabular}}
\end{center}
\caption{ Domain generalization accuracy (\%) on Rotated MNIST.}
\label{tab:rotatedmnist}
\end{table}

\begin{table}[b]
\begin{center}
\adjustbox{max width=\textwidth}{%
\setlength{\tabcolsep}{7.25pt}
\begin{tabular}{lccccc}
\toprule
\multicolumn{6}{c}{\textbf{VLCS, Model selection: ‘Test-domain’ validation set}}\\
\textbf{Algorithm}   & \textbf{C}           & \textbf{L}           & \textbf{S}           & \textbf{V}           & \textbf{Avg}         \\
\midrule
ERM~\cite{vapnik1998statistical}                  & 97.6 $\pm$ 0.3       & 67.9 $\pm$ 0.7       & 70.9 $\pm$ 0.2       & 74.0 $\pm$ 0.6       & 77.6                 \\
IRM~\cite{arjovsky2020invariant}                  & 97.3 $\pm$ 0.2       & 66.7 $\pm$ 0.1       & 71.0 $\pm$ 2.3       & 72.8 $\pm$ 0.4       & 76.9                 \\
GDRO~\cite{sagawa2020distributionally}             & 97.7 $\pm$ 0.2       & 65.9 $\pm$ 0.2       & 72.8 $\pm$ 0.8       & 73.4 $\pm$ 1.3       & 77.4                 \\
Mixup~\cite{yan2020improve}                & 97.8 $\pm$ 0.4       & 67.2 $\pm$ 0.4       & 71.5 $\pm$ 0.2       & 75.7 $\pm$ 0.6       & 78.1                 \\
MLDG~\cite{li2018learning}                 & 97.1 $\pm$ 0.5       & 66.6 $\pm$ 0.5       & 71.5 $\pm$ 0.1       & 75.0 $\pm$ 0.9       & 77.5                 \\
CORAL~\cite{sun2016deep}                & 97.3 $\pm$ 0.2       & 67.5 $\pm$ 0.6       & 71.6 $\pm$ 0.6       & 74.5 $\pm$ 0.0       & 77.7                 \\
MMD~\cite{li2018domain}                  & 98.8 $\pm$ 0.0       & 66.4 $\pm$ 0.4       & 70.8 $\pm$ 0.5       & 75.6 $\pm$ 0.4       & 77.9                 \\
DANN~\cite{ganin2016domain}                 & 99.0 $\pm$ 0.2       & 66.3 $\pm$ 1.2       & 73.4 $\pm$ 1.4       & 80.1 $\pm$ 0.5       & 79.7                 \\
CDANN~\cite{li2018deep}               & 98.2 $\pm$ 0.1       & 68.8 $\pm$ 0.5       & 74.3 $\pm$ 0.6       & 78.1 $\pm$ 0.5       & 79.9                 \\
MTL~\cite{blanchard2021domain}                  & 97.9 $\pm$ 0.7       & 66.1 $\pm$ 0.7       & 72.0 $\pm$ 0.4       & 74.9 $\pm$ 1.1       & 77.7                 \\
SagNet~\cite{nam2021reducing}               & 97.4 $\pm$ 0.3       & 66.4 $\pm$ 0.4       & 71.6 $\pm$ 0.1       & 75.0 $\pm$ 0.8       & 77.6                 \\
ARM~\cite{zhang2021adaptive}                  & 97.6 $\pm$ 0.6       & 66.5 $\pm$ 0.3       & 72.7 $\pm$ 0.6       & 74.4 $\pm$ 0.7       & 77.8                 \\
VREx~\cite{krueger2021out}                 & 98.4 $\pm$ 0.2       & 66.4 $\pm$ 0.7       & 72.8 $\pm$ 0.1       & 75.0 $\pm$ 1.4       & 78.1                 \\
RSC~\cite{huang2020self}                  & 98.0 $\pm$ 0.4       & 67.2 $\pm$ 0.3       & 70.3 $\pm$ 1.3       & 75.6 $\pm$ 0.4       & 77.8                 \\
Fish~\cite{shi2022gradient} &       &       &  &  &         77.8      \\
Fisher~\cite{rame2022fishr} & 97.6 $\pm$ 0.7       & 67.3 $\pm$ 0.5       & 72.2 $\pm$ 0.9       & 75.7 $\pm$ 0.3       & 78.2    \\
\rowcolor{Gray}
\knn                & \textbf{98.7 $\pm$ 0.2}       & 66.6 $\pm$ 0.2      & 74.6 $\pm$ 0.3       & 79.6 $\pm$ 0.5    &79.9            \\
\rowcolor{Gray}
+BN retraining                & \textbf{98.7 $\pm$ 0.2}       & \textbf{67.4 $\pm$ 0.3}      &\textbf{ 74.9 $\pm$ 0.5}       & \textbf{79.7 $\pm$ 0.5}    &\textbf{80.2 }           \\\midrule
\multicolumn{6}{c}{\textbf{VLCS, Model selection: ‘Training-domain’ validation set}}\\
\textbf{Algorithm}   & \textbf{C}           & \textbf{L}           & \textbf{S}           & \textbf{V}           & \textbf{Avg}         \\
\midrule
ERM~\cite{vapnik1998statistical}                 & 97.7 $\pm$ 0.4       & 64.3 $\pm$ 0.9       & 73.4 $\pm$ 0.5       & 74.6 $\pm$ 1.3       & 77.5                 \\
IRM~\cite{arjovsky2020invariant}              & 98.6 $\pm$ 0.1       & 64.9 $\pm$ 0.9       & 73.4 $\pm$ 0.6       & 77.3 $\pm$ 0.9       & 78.5                 \\
GDRO~\cite{sagawa2020distributionally}          & 97.3 $\pm$ 0.3       & 63.4 $\pm$ 0.9       & 69.5 $\pm$ 0.8       & 76.7 $\pm$ 0.7       & 76.7                 \\
Mixup~\cite{yan2020improve}            & 98.3 $\pm$ 0.6       & 64.8 $\pm$ 1.0       & 72.1 $\pm$ 0.5       & 74.3 $\pm$ 0.8       & 77.4                 \\
MLDG~\cite{li2018learning}             & 97.4 $\pm$ 0.2       & 65.2 $\pm$ 0.7       & 71.0 $\pm$ 1.4       & 75.3 $\pm$ 1.0       & 77.2                 \\
CORAL~\cite{sun2016deep}            & 98.3 $\pm$ 0.1       & 66.1 $\pm$ 1.2       & 73.4 $\pm$ 0.3       & 77.5 $\pm$ 1.2       & 78.8                 \\
MMD~\cite{li2018domain}              & 97.7 $\pm$ 0.1       & 64.0 $\pm$ 1.1       & 72.8 $\pm$ 0.2       & 75.3 $\pm$ 3.3       & 77.5                 \\
DANN~\cite{ganin2016domain}              & \textbf{99.0 $\pm$ 0.3}       & 65.1 $\pm$ 1.4       & 73.1 $\pm$ 0.3       & 77.2 $\pm$ 0.6       & 78.6                 \\
CDANN~\cite{li2018deep}         & 97.1 $\pm$ 0.3       & 65.1 $\pm$ 1.2       & 70.7 $\pm$ 0.8       & 77.1 $\pm$ 1.5       & 77.5                 \\
MTL~\cite{blanchard2021domain}              & 97.8 $\pm$ 0.4       & 64.3 $\pm$ 0.3       & 71.5 $\pm$ 0.7       & 75.3 $\pm$ 1.7       & 77.2                 \\
SagNet~\cite{nam2021reducing}           & 97.9 $\pm$ 0.4       & 64.5 $\pm$ 0.5       & 71.4 $\pm$ 1.3       & 77.5 $\pm$ 0.5       & 77.8                 \\
ARM~\cite{zhang2021adaptive}              & 98.7 $\pm$ 0.2       & 63.6 $\pm$ 0.7       & 71.3 $\pm$ 1.2       & 76.7 $\pm$ 0.6       & 77.6                 \\
VREx~\cite{krueger2021out}             & 98.4 $\pm$ 0.3       & 64.4 $\pm$ 1.4       & 74.1 $\pm$ 0.4       & 76.2 $\pm$ 1.3       & 78.3                 \\
RSC~\cite{huang2020self}               & 97.9 $\pm$ 0.1       & 62.5 $\pm$ 0.7       & 72.3 $\pm$ 1.2       & 75.6 $\pm$ 0.8       & 77.1                 \\     
Fish~\cite{shi2022gradient} &       &       &  &  &         77.8      \\
Fisher~\cite{rame2022fishr} & 98.9 $\pm$ 0.3       & 64.0 $\pm$ 0.5       & 71.5 $\pm$ 0.2       & 76.8 $\pm$ 0.7       & 77.8    \\
\rowcolor{Gray}
\knn                & 98.9 $\pm$ 0.3       & 64.5 $\pm$ 1.0      & 73.5 $\pm$ 0.7       & 75.6 $\pm$ 0.8    &78.1            \\
\rowcolor{Gray}
+BN retraining                & 98.4 $\pm$ 0.6       & \textbf{65.2 $\pm$ 1.2}      &\textbf{ 74.4 $\pm$ 0.3}       & \textbf{77.4 $\pm$ 1.1}    &\textbf{78.9 }           \\
\bottomrule
\end{tabular}}
\end{center}
\caption{ Domain generalization accuracy (\%) on VLCS.}
\label{tab:vlcs}
\end{table}

\begin{table}[b]
\begin{center}
\adjustbox{max width=\textwidth}{%
\setlength{\tabcolsep}{7.25pt}
\begin{tabular}{lccccc}
\toprule
\multicolumn{6}{c}{\textbf{PACS, Model selection: ‘Test-domain’ validation set}}\\
\textbf{Algorithm}   & \textbf{A}           & \textbf{C}           & \textbf{P}           & \textbf{S}           & \textbf{Avg}         \\
\midrule
ERM~\cite{vapnik1998statistical}                 & 86.5 $\pm$ 1.0       & 81.3 $\pm$ 0.6       & 96.2 $\pm$ 0.3       & 82.7 $\pm$ 1.1       & 86.7                 \\
IRM~\cite{arjovsky2020invariant}              & 84.2 $\pm$ 0.9       & 79.7 $\pm$ 1.5       & 95.9 $\pm$ 0.4       & 78.3 $\pm$ 2.1       & 84.5                 \\
GDRO~\cite{sagawa2020distributionally}          & 87.5 $\pm$ 0.5       & 82.9 $\pm$ 0.6       & 97.1 $\pm$ 0.3       & 81.1 $\pm$ 1.2       & 87.1                 \\
Mixup~\cite{yan2020improve}            & 87.5 $\pm$ 0.4       & 81.6 $\pm$ 0.7       & 97.4 $\pm$ 0.2       & 80.8 $\pm$ 0.9       & 86.8                 \\
MLDG~\cite{li2018learning}             & 87.0 $\pm$ 1.2       & 82.5 $\pm$ 0.9       & 96.7 $\pm$ 0.3       & 81.2 $\pm$ 0.6       & 86.8                 \\
CORAL~\cite{sun2016deep}            & 86.6 $\pm$ 0.8       & 81.8 $\pm$ 0.9       & 97.1 $\pm$ 0.5       & 82.7 $\pm$ 0.6       & 87.1                 \\
MMD~\cite{li2018domain}              & 88.1 $\pm$ 0.8       & 82.6 $\pm$ 0.7       & 97.1 $\pm$ 0.5       & 81.2 $\pm$ 1.2       & 87.2                 \\
DANN~\cite{ganin2016domain}              & 87.0 $\pm$ 0.4       & 80.3 $\pm$ 0.6       & 96.8 $\pm$ 0.3       & 76.9 $\pm$ 1.1       & 85.2                 \\
CDANN~\cite{li2018deep}         & 87.7 $\pm$ 0.6       & 80.7 $\pm$ 1.2       & 97.3 $\pm$ 0.4       & 77.6 $\pm$ 1.5       & 85.8                 \\
MTL~\cite{blanchard2021domain}              & 87.0 $\pm$ 0.2       & 82.7 $\pm$ 0.8       & 96.5 $\pm$ 0.7       & 80.5 $\pm$ 0.8       & 86.7                 \\
SagNet~\cite{nam2021reducing}           & 87.4 $\pm$ 0.5       & 81.2 $\pm$ 1.2       & 96.3 $\pm$ 0.8       & 80.7 $\pm$ 1.1       & 86.4                 \\
ARM~\cite{zhang2021adaptive}              & 85.0 $\pm$ 1.2       & 81.4 $\pm$ 0.2       & 95.9 $\pm$ 0.3       & 80.9 $\pm$ 0.5       & 85.8                 \\
VREx~\cite{krueger2021out}             & 87.8 $\pm$ 1.2       & 81.8 $\pm$ 0.7       & 97.4 $\pm$ 0.2       & 82.1 $\pm$ 0.7       & 87.2                 \\
RSC~\cite{huang2020self}               & 86.0 $\pm$ 0.7       & 81.8 $\pm$ 0.9       & 96.8 $\pm$ 0.7       & 80.4 $\pm$ 0.5       & 86.2                 \\
Fish~\cite{shi2022gradient} &       &       &  &  &         85.8      \\
Fisher~\cite{rame2022fishr} & 87.9 $\pm$ 0.6       & 80.8 $\pm$ 0.5       & 97.9 $\pm$ 0.4       & 81.1 $\pm$ 0.8       & 86.9    \\
\rowcolor{Gray}
\knn                & 89.1 $\pm$ 0.3       & \textbf{84.3 $\pm$ 0.1}      & \textbf{98.1 $\pm$ 0.4}       & 83.7 $\pm$ 0.5    &\textbf{88.8}           \\
\rowcolor{Gray}
\knn+NP                & \textbf{89.2 $\pm$ 0.3}       & \textbf{84.3 $\pm$ 0.1}      & 98.0 $\pm$ 0.4       & \textbf{83.8 $\pm$ 0.4}    &\textbf{88.9}           \\
\midrule
\multicolumn{6}{c}{\textbf{PACS, Model selection: ‘Training-domain’ validation set}}\\
\textbf{Algorithm}   & \textbf{A}           & \textbf{C}           & \textbf{P}           & \textbf{S}           & \textbf{Avg}         \\
\midrule
ERM~\cite{vapnik1998statistical}                 & 84.7 $\pm$ 0.4       & 80.8 $\pm$ 0.6       & 97.2 $\pm$ 0.3       & 79.3 $\pm$ 1.0       & 85.5                 \\
IRM~\cite{arjovsky2020invariant}              & 84.8 $\pm$ 1.3       & 76.4 $\pm$ 1.1       & 96.7 $\pm$ 0.6       & 76.1 $\pm$ 1.0       & 83.5                 \\
GDRO~\cite{sagawa2020distributionally}          & 83.5 $\pm$ 0.9       & 79.1 $\pm$ 0.6       & 96.7 $\pm$ 0.3       & 78.3 $\pm$ 2.0       & 84.4                 \\
Mixup~\cite{yan2020improve}            & 86.1 $\pm$ 0.5       & 78.9 $\pm$ 0.8       & 97.6 $\pm$ 0.1       & 75.8 $\pm$ 1.8       & 84.6                 \\
MLDG~\cite{li2018learning}             & 85.5 $\pm$ 1.4       & 80.1 $\pm$ 1.7       & 97.4 $\pm$ 0.3       & 76.6 $\pm$ 1.1       & 84.9                 \\
CORAL~\cite{sun2016deep}            & 88.3 $\pm$ 0.2       & 80.0 $\pm$ 0.5       & 97.5 $\pm$ 0.3       & 78.8 $\pm$ 1.3       & 86.2                 \\
MMD~\cite{li2018domain}              & 86.1 $\pm$ 1.4       & 79.4 $\pm$ 0.9       & 96.6 $\pm$ 0.2       & 76.5 $\pm$ 0.5       & 84.6                 \\
DANN~\cite{ganin2016domain}              & 86.4 $\pm$ 0.8       & 77.4 $\pm$ 0.8       & 97.3 $\pm$ 0.4       & 73.5 $\pm$ 2.3       & 83.6                 \\
CDANN~\cite{li2018deep}         & 84.6 $\pm$ 1.8       & 75.5 $\pm$ 0.9       & 96.8 $\pm$ 0.3       & 73.5 $\pm$ 0.6       & 82.6                 \\
MTL~\cite{blanchard2021domain}              & 87.5 $\pm$ 0.8       & 77.1 $\pm$ 0.5       & 96.4 $\pm$ 0.8       & 77.3 $\pm$ 1.8       & 84.6                 \\
SagNet~\cite{nam2021reducing}           & 87.4 $\pm$ 1.0       & 80.7 $\pm$ 0.6       & 97.1 $\pm$ 0.1       & 80.0 $\pm$ 0.4       & 86.3                 \\
ARM~\cite{zhang2021adaptive}              & 86.8 $\pm$ 0.6       & 76.8 $\pm$ 0.5       & 97.4 $\pm$ 0.3       & 79.3 $\pm$ 1.2       & 85.1                 \\
VREx~\cite{krueger2021out}             & 86.0 $\pm$ 1.6       & 79.1 $\pm$ 0.6       & 96.9 $\pm$ 0.5       & 77.7 $\pm$ 1.7       & 84.9                 \\
RSC~\cite{huang2020self}               & 85.4 $\pm$ 0.8       & 79.7 $\pm$ 1.8       & 97.6 $\pm$ 0.3       & 78.2 $\pm$ 1.2       & 85.2                 \\
Fish~\cite{shi2022gradient} &       &       &  &  &          85.5      \\
Fisher~\cite{rame2022fishr} & \textbf{88.4 $\pm$ 0.2}       & 78.7 $\pm$ 0.7       & 97.0 $\pm$ 0.1       & 77.8 $\pm$ 2.0       & 85.5   \\
\rowcolor{Gray}
\knn                & 87.1 $\pm$ 1.3       &\textbf{ 82.2 $\pm$ 0.6}      & 97.5 $\pm$ 0.4       & \textbf{81.5 $\pm$ 0.8}    &\textbf{87.1}            \\
\rowcolor{Gray}
\knn+NP                & 86.2 $\pm$ 1.2       & \textbf{82.2 $\pm$ 0.6 }     & \textbf{98.1 $\pm$ 0.1}       & 80.2 $\pm$ 1.0    &{86.7}            \\
\bottomrule
\end{tabular}}
\end{center}
\caption{ Domain generalization accuracy (\%) on PACS.}
\label{tab:pacs}
\end{table}

\begin{table}[t]
\begin{center}
\adjustbox{max width=\textwidth}{%
\setlength{\tabcolsep}{7.25pt}
\begin{tabular}{lccccccc}
\specialrule{0em}{8pt}{0pt}
\toprule
\multicolumn{8}{c}{\textbf{DomainNet, Model selection: ‘Test-domain’ validation set}}\\
\textbf{Algorithm}   & \textbf{clip}        & \textbf{info}        & \textbf{paint}       & \textbf{quick}       & \textbf{real}        & \textbf{sketch}      & \textbf{Avg}         \\
\midrule
ERM~\cite{vapnik1998statistical}                 & 58.1 $\pm$ 0.3       & 18.8 $\pm$ 0.3       & 46.7 $\pm$ 0.3       & 12.2 $\pm$ 0.4       & 59.6 $\pm$ 0.1       & 49.8 $\pm$ 0.4       & 40.9                 \\
IRM~\cite{arjovsky2020invariant}              & 48.5 $\pm$ 2.8       & 15.0 $\pm$ 1.5       & 38.3 $\pm$ 4.3       & 10.9 $\pm$ 0.5       & 48.2 $\pm$ 5.2       & 42.3 $\pm$ 3.1       & 33.9                 \\
GDRO~\cite{sagawa2020distributionally}          & 47.2 $\pm$ 0.5       & 17.5 $\pm$ 0.4       & 33.8 $\pm$ 0.5       & 9.3 $\pm$ 0.3        & 51.6 $\pm$ 0.4       & 40.1 $\pm$ 0.6       & 33.3                 \\
Mixup~\cite{yan2020improve}            & 55.7 $\pm$ 0.3       & 18.5 $\pm$ 0.5       & 44.3 $\pm$ 0.5       & 12.5 $\pm$ 0.4       & 55.8 $\pm$ 0.3       & 48.2 $\pm$ 0.5       & 39.2                 \\
MLDG~\cite{li2018learning}             & 59.1 $\pm$ 0.2       & 19.1 $\pm$ 0.3       & 45.8 $\pm$ 0.7       & 13.4 $\pm$ 0.3       & 59.6 $\pm$ 0.2       & 50.2 $\pm$ 0.4       & 41.2                 \\
CORAL~\cite{sun2016deep}            & 59.2 $\pm$ 0.1       & 19.7 $\pm$ 0.2       & 46.6 $\pm$ 0.3       & 13.4 $\pm$ 0.4       & 59.8 $\pm$ 0.2       & 50.1 $\pm$ 0.6       & 41.5                 \\
MMD~\cite{li2018domain}              & 32.1 $\pm$ 13.3      & 11.0 $\pm$ 4.6       & 26.8 $\pm$ 11.3      & 8.7 $\pm$ 2.1        & 32.7 $\pm$ 13.8      & 28.9 $\pm$ 11.9      & 23.4                 \\
DANN~\cite{ganin2016domain}              & 53.1 $\pm$ 0.2       & 18.3 $\pm$ 0.1       & 44.2 $\pm$ 0.7       & 11.8 $\pm$ 0.1       & 55.5 $\pm$ 0.4       & 46.8 $\pm$ 0.6       & 38.3                 \\
CDANN~\cite{li2018deep}         & 54.6 $\pm$ 0.4       & 17.3 $\pm$ 0.1       & 43.7 $\pm$ 0.9       & 12.1 $\pm$ 0.7       & 56.2 $\pm$ 0.4       & 45.9 $\pm$ 0.5       & 38.3                 \\
MTL~\cite{blanchard2021domain}              & 57.9 $\pm$ 0.5       & 18.5 $\pm$ 0.4       & 46.0 $\pm$ 0.1       & 12.5 $\pm$ 0.1       & 59.5 $\pm$ 0.3       & 49.2 $\pm$ 0.1       & 40.6                 \\
SagNet~\cite{nam2021reducing}           & 57.7 $\pm$ 0.3       & 19.0 $\pm$ 0.2       & 45.3 $\pm$ 0.3       & 12.7 $\pm$ 0.5       & 58.1 $\pm$ 0.5       & 48.8 $\pm$ 0.2       & 40.3                 \\
ARM~\cite{zhang2021adaptive}              & 49.7 $\pm$ 0.3       & 16.3 $\pm$ 0.5       & 40.9 $\pm$ 1.1       & 9.4 $\pm$ 0.1        & 53.4 $\pm$ 0.4       & 43.5 $\pm$ 0.4       & 35.5                 \\
VREx~\cite{krueger2021out}             & 47.3 $\pm$ 3.5       & 16.0 $\pm$ 1.5       & 35.8 $\pm$ 4.6       & 10.9 $\pm$ 0.3       & 49.6 $\pm$ 4.9       & 42.0 $\pm$ 3.0       & 33.6                 \\
RSC~\cite{huang2020self}               & 55.0 $\pm$ 1.2       & 18.3 $\pm$ 0.5       & 44.4 $\pm$ 0.6       & 12.2 $\pm$ 0.2       & 55.7 $\pm$ 0.7       & 47.8 $\pm$ 0.9       & 38.9  \\
Fish~\cite{shi2022gradient}                  &        &        &        &       &       &       & 43.4      \\
Fisher~\cite{rame2022fishr} & 58.3 $\pm$ 0.5       & 20.2 $\pm$ 0.2      & 47.9 $\pm$ 0.2       & 13.6 $\pm$ 0.3       & 60.5 $\pm$ 0.3       & 50.5 $\pm$ 0.3       & 41.8                 \\
\rowcolor{Gray}
\knn      & \textbf{59.5 $\pm$ 0.1}       & \textbf{22.2 $\pm$ 0.9}       & \textbf{48.2 $\pm$ 0.9}       & \textbf{15.3 $\pm$ 0.2}       & \textbf{61.2 $\pm$ 0.0}       & \textbf{51.1 $\pm$ 0.1}       &     42.9            \\
\bottomrule
\multicolumn{8}{c}{\textbf{DomainNet, Model selection: ‘Training-domain’ validation set}}\\
\textbf{Algorithm}   & \textbf{clip}        & \textbf{info}        & \textbf{paint}       & \textbf{quick}       & \textbf{real}        & \textbf{sketch}      & \textbf{Avg}         \\
\midrule
ERM~\cite{vapnik1998statistical}                  & 58.6 $\pm$ 0.3       & 19.2 $\pm$ 0.2       & 47.0 $\pm$ 0.3       & 13.2 $\pm$ 0.2       & 59.9 $\pm$ 0.3       & 49.8 $\pm$ 0.4       & 41.3                 \\
IRM~\cite{arjovsky2020invariant}                  & 40.4 $\pm$ 6.6       & 12.1 $\pm$ 2.7       & 31.4 $\pm$ 5.7       & 9.8 $\pm$ 1.2        & 37.7 $\pm$ 9.0       & 36.7 $\pm$ 5.3       & 28.0                 \\
GDRO~\cite{sagawa2020distributionally}             & 47.2 $\pm$ 0.5       & 17.5 $\pm$ 0.4       & 34.2 $\pm$ 0.3       & 9.2 $\pm$ 0.4        & 51.9 $\pm$ 0.5       & 40.1 $\pm$ 0.6       & 33.4                 \\
Mixup~\cite{yan2020improve}                & 55.6 $\pm$ 0.1       & 18.7 $\pm$ 0.4       & 45.1 $\pm$ 0.5       & 12.8 $\pm$ 0.3       & 57.6 $\pm$ 0.5       & 48.2 $\pm$ 0.4       & 39.6                 \\
MLDG~\cite{li2018learning}                 & 59.3 $\pm$ 0.1       & 19.6 $\pm$ 0.2       & 46.8 $\pm$ 0.2       & 13.4 $\pm$ 0.2       & 60.1 $\pm$ 0.4       & 50.4 $\pm$ 0.3       & 41.6                 \\
CORAL~\cite{sun2016deep}                & 59.2 $\pm$ 0.1       & 19.9 $\pm$ 0.2       & 47.4 $\pm$ 0.2       & 14.0 $\pm$ 0.4       & 59.8 $\pm$ 0.2       & 50.4 $\pm$ 0.4       & 41.8                 \\
MMD~\cite{li2018domain}                  & 32.2 $\pm$ 13.3      & 11.2 $\pm$ 4.5       & 26.8 $\pm$ 11.3      & 8.8 $\pm$ 2.2        & 32.7 $\pm$ 13.8      & 29.0 $\pm$ 11.8      & 23.5                 \\
DANN~\cite{ganin2016domain}                 & 53.1 $\pm$ 0.2       & 18.3 $\pm$ 0.1       & 44.2 $\pm$ 0.7       & 11.9 $\pm$ 0.1       & 55.5 $\pm$ 0.4       & 46.8 $\pm$ 0.6       & 38.3                 \\
CDANN~\cite{li2018deep}                & 54.6 $\pm$ 0.4       & 17.3 $\pm$ 0.1       & 44.2 $\pm$ 0.7       & 12.8 $\pm$ 0.2       & 56.2 $\pm$ 0.4       & 45.9 $\pm$ 0.5       & 38.5                 \\
MTL~\cite{blanchard2021domain}                  & 58.0 $\pm$ 0.4       & 19.2 $\pm$ 0.2       & 46.2 $\pm$ 0.1       & 12.7 $\pm$ 0.2       & 59.9 $\pm$ 0.1       & 49.0 $\pm$ 0.0       & 40.8                 \\
SagNet~\cite{nam2021reducing}               & 57.7 $\pm$ 0.3       & 19.1 $\pm$ 0.1       & 46.3 $\pm$ 0.5       & 13.5 $\pm$ 0.4       & 58.9 $\pm$ 0.4       & 49.5 $\pm$ 0.2       & 40.8                 \\
ARM~\cite{zhang2021adaptive}                  & 49.6 $\pm$ 0.4       & 16.5 $\pm$ 0.3       & 41.5 $\pm$ 0.8       & 10.8 $\pm$ 0.1       & 53.5 $\pm$ 0.3       & 43.9 $\pm$ 0.4       & 36.0                 \\
VREx~\cite{krueger2021out}                 & 43.3 $\pm$ 4.5       & 14.1 $\pm$ 1.8       & 32.5 $\pm$ 5.0       & 9.8 $\pm$ 1.1        & 43.5 $\pm$ 5.6       & 37.7 $\pm$ 4.5       & 30.1                 \\
RSC~\cite{huang2020self}                  & 55.0 $\pm$ 1.2       & 18.3 $\pm$ 0.5       & 44.4 $\pm$ 0.6       & 12.5 $\pm$ 0.1       & 55.7 $\pm$ 0.7       & 47.8 $\pm$ 0.9       & 38.9       \\
Fish~\cite{shi2022gradient}                  &        &        &        &       &       &       & 42.7      \\
Fisher~\cite{rame2022fishr} & 58.2 $\pm$ 0.5       & 20.2 $\pm$ 0.2      & 47.7 $\pm$ 0.3       & 12.7 $\pm$ 0.2       & 60.3 $\pm$ 0.2       & 50.8 $\pm$ 0.1       & 41.7                 \\
\rowcolor{Gray}
\knn      & \textbf{59.3 $\pm$ 0.0}       & \textbf{22.2 $\pm$ 0.9}       & \textbf{48.3 $\pm$ 0.0}       & \textbf{14.3 $\pm$ 0.0}       & \textbf{61.0 $\pm$ 0.1}       & \textbf{51.4 $\pm$ 0.0}       &     \textbf{42.8}            \\
\bottomrule
\end{tabular}}
\end{center}
\caption{ Domain generalization accuracy (\%) on DomainNet.}
\label{tab:domainnet}
\end{table}

\end{document}